\ifdefined\XeTeXrevision\else\pdfoutput=1\fi
\def\withsupplement{}
\def\wideformat{}
\documentclass[pdflatex,sn-basic]{sn-jnl}
\usepackage{graphicx}
\usepackage{booktabs}
\usepackage{array}
\usepackage{amsmath,amssymb}
\usepackage{pifont}
\usepackage[title]{appendix}
\usepackage{xcolor}
\usepackage[T1]{fontenc}
\usepackage{microtype}
\usepackage{tikz}
\usetikzlibrary{arrows.meta,calc}
\ifdefined\wideformat\fi %

\newcommand{\NTestCases}{42}
\newcommand{\NClauses}{14}
\newcommand{\NStates}{6}

\newcommand{\NCleanC}{42}

\newcommand{\NRegistryFlagged}{11}

\newcommand{\NCorpusDocs}{111}
\newcommand{\NCorpusJurisdictions}{8}
\newcommand{\NPanelModels}{4}
\newcommand{\NCapable}{2}

\newcommand{\NDamageSeedsCapable}{5}
\newcommand{\NDamageSeeds}{5}
\newcommand{\DamageDropTwentyFive}{25}
\newcommand{\DamageDropFifty}{50}
\newcommand{\DamageSpuriousTen}{10}
\newcommand{\Temperature}{0}
\newcommand{\RunSeed}{42}
\newcommand{\OutputCap}{1024}
\newcommand{\CILevel}{95}
\newcommand{\Alpha}{0.05}
\newcommand{\MinClauses}{10}
\newcommand{\SparseItems}{5}
\newcommand{\SparseClauses}{3}
\newcommand{\InterpPoints}{10}
\newcommand{\NPermutations}{10{,}000}
\newcommand{\ScorerCheckN}{50}
\newcommand{\ScorerCheckPct}{95}

\newcommand{\NLinksAnnotated}{52}
\newcommand{\NLinksUsed}{18}
\newcommand{\LinkSamplePct}{20}
\newcommand{\NControlsAnnotated}{40}

\newcommand{\NSharedInstrument}{3}

\newcommand{\NFaultsSBoth}{42}
\newcommand{\NFaultsSSwap}{38}

\newcommand{\NControlsSBoth}{42}
\newcommand{\NControlsSSwap}{7}
\newcommand{\NClausesSBoth}{14}
\newcommand{\NClausesSSwap}{13}

\newcommand{\NClausesStateDL}{4}
\newcommand{\NClausesStateGJ}{4}
\newcommand{\NClausesStateIN}{1}
\newcommand{\NClausesStateKA}{2}
\newcommand{\NClausesStateTN}{1}
\newcommand{\NClausesStateUP}{2}
\newcommand{\NSharedLinks}{15}
\newcommand{\NAgreeItemsLinks}{15}
\newcommand{\AgreeLinksRelation}{87\%}
\newcommand{\KappaLinksRelation}{0.70}
\newcommand{\PabakLinksRelation}{0.82}
\newcommand{\AgreeLinksEffectiveDate}{80\%}
\newcommand{\KappaLinksEffectiveDate}{0.74}
\newcommand{\PabakLinksEffectiveDate}{0.78}
\newcommand{\AgreeLinksWording}{93\%}
\newcommand{\KappaLinksWording}{0.84}
\newcommand{\PabakLinksWording}{0.90}
\newcommand{\AgreeLinksNewClauseId}{7\%}
\newcommand{\KappaLinksNewClauseId}{0.06}
\newcommand{\PabakLinksNewClauseId}{0.03}
\newcommand{\NSharedCases}{20}
\newcommand{\NAgreeItemsCases}{20}
\newcommand{\AgreeCasesDecision}{80\%}
\newcommand{\KappaCasesDecision}{0.60}
\newcommand{\PabakCasesDecision}{0.70}
\newcommand{\AgreeCasesBlockingStage}{85\%}
\newcommand{\KappaCasesBlockingStage}{0.79}
\newcommand{\PabakCasesBlockingStage}{0.82}
\newcommand{\AgreeCasesDecidingClauses}{15\%}
\newcommand{\KappaCasesDecidingClauses}{0.14}
\newcommand{\PabakCasesDecidingClauses}{0.11}
\newcommand{\NCandidateCases}{60}
\newcommand{\NDroppedDecision}{18}
\newcommand{\NClauseDisputed}{36}
\newcommand{\NStrictKept}{6}
\newcommand{\NControlsChecked}{40}

\newcommand{\PctControlsCorrect}{57\%}
\newcommand{\NRepeatItems}{378}
\newcommand{\NRepeatSameFindings}{378}
\newcommand{\NDroppedDecisionOnly}{13}
\newcommand{\NDroppedBlockingStage}{5}
\newcommand{\NControlsCorrect}{23}
\newcommand{\NLinksRegistry}{186}
\newcommand{\NLinksSample}{34}
\newcommand{\NLinksSampleConfirmed}{25}
\newcommand{\NLinksSampleRelChanged}{4}
\newcommand{\NLinksSampleNotALink}{5}
\newcommand{\CINotALinkSample}{[6, 30]\%}
\newcommand{\NLinksTestPairNotALink}{1}
\newcommand{\NLinksTestPairRelChanged}{3}
\newcommand{\ScorerCheckAgreeN}{35 of 50}
\newcommand{\ScorerCheckPctObserved}{70\%}
\newcommand{\NScorerDisagree}{15}
\newcommand{\NScorerClauseReason}{10}
\newcommand{\NScorerCatchReason}{3}
\newcommand{\ControlsGatePct}{95}
\newcommand{\NPrimaryTestsVFive}{10}

\newcommand{\CatchLlamaRZero}{0\%}

\newcommand{\FAMLlamaRZero}{80\%}

\newcommand{\FACLlamaRZero}{51\%}

\newcommand{\CatchLlamaROne}{0\%}

\newcommand{\FAMLlamaROne}{80\%}

\newcommand{\FACLlamaROne}{49\%}

\newcommand{\CatchLlamaRTwo}{0\%}

\newcommand{\FAMLlamaRTwo}{49\%}

\newcommand{\FACLlamaRTwo}{49\%}

\newcommand{\CatchLlamaRThree}{0\%}

\newcommand{\FAMLlamaRThree}{49\%}

\newcommand{\FACLlamaRThree}{49\%}

\newcommand{\CatchLlamaRFour}{11\%}

\newcommand{\FAMLlamaRFour}{49\%}

\newcommand{\FACLlamaRFour}{51\%}

\newcommand{\CatchLlamaRFourP}{6\%}

\newcommand{\FAMLlamaRFourP}{80\%}

\newcommand{\FACLlamaRFourP}{51\%}

\newcommand{\CatchLlamaRFive}{79\%}

\newcommand{\FAMLlamaRFive}{13\%}

\newcommand{\FACLlamaRFive}{12\%}

\newcommand{\DiffRFourMinusRTwoLlama}{+11}
\newcommand{\CIRFourMinusRTwoLlama}{[+0, +29]}

\newcommand{\CatchQwenSevenRZero}{4\%}

\newcommand{\FAMQwenSevenRZero}{80\%}

\newcommand{\FACQwenSevenRZero}{20\%}

\newcommand{\CatchQwenSevenROne}{4\%}

\newcommand{\FAMQwenSevenROne}{60\%}

\newcommand{\FACQwenSevenROne}{12\%}

\newcommand{\CatchQwenSevenRTwo}{18\%}

\newcommand{\FAMQwenSevenRTwo}{18\%}

\newcommand{\FACQwenSevenRTwo}{12\%}

\newcommand{\CatchQwenSevenRThree}{18\%}

\newcommand{\FAMQwenSevenRThree}{20\%}

\newcommand{\FACQwenSevenRThree}{19\%}

\newcommand{\CatchQwenSevenRFour}{7\%}

\newcommand{\FAMQwenSevenRFour}{7\%}

\newcommand{\FACQwenSevenRFour}{5\%}

\newcommand{\CatchQwenSevenRFourP}{6\%}

\newcommand{\FAMQwenSevenRFourP}{40\%}

\newcommand{\FACQwenSevenRFourP}{5\%}

\newcommand{\CatchQwenSevenRFive}{79\%}

\newcommand{\FAMQwenSevenRFive}{13\%}

\newcommand{\FACQwenSevenRFive}{12\%}

\newcommand{\CatchQwenFourteenRZero}{8\%}

\newcommand{\FAMQwenFourteenRZero}{40\%}

\newcommand{\FACQwenFourteenRZero}{26\%}

\newcommand{\CatchQwenFourteenROne}{8\%}

\newcommand{\FAMQwenFourteenROne}{50\%}

\newcommand{\FACQwenFourteenROne}{27\%}

\newcommand{\CatchQwenFourteenRTwo}{33\%}

\newcommand{\FAMQwenFourteenRTwo}{40\%}

\newcommand{\FACQwenFourteenRTwo}{27\%}

\newcommand{\MisnamedKNQwenFourteenRTwo}{13/26}

\newcommand{\CatchQwenFourteenRThree}{30\%}

\newcommand{\FAMQwenFourteenRThree}{36\%}

\newcommand{\FACQwenFourteenRThree}{27\%}

\newcommand{\CatchQwenFourteenRFour}{79\%}

\newcommand{\FAMQwenFourteenRFour}{23\%}

\newcommand{\FACQwenFourteenRFour}{26\%}

\newcommand{\CatchQwenFourteenRFourP}{81\%}

\newcommand{\FAMQwenFourteenRFourP}{0\%}

\newcommand{\FACQwenFourteenRFourP}{26\%}

\newcommand{\CatchQwenFourteenRFive}{79\%}

\newcommand{\FAMQwenFourteenRFive}{13\%}

\newcommand{\FACQwenFourteenRFive}{12\%}

\newcommand{\CantConfirmKNQwenFourteenRTwoCC}{0/42}

\newcommand{\FAMQwenFourteenRFourDropFifty}{25\%}

\newcommand{\DiffRFourMinusRTwoQwenFourteen}{+45}
\newcommand{\CIRFourMinusRTwoQwenFourteen}{[+24, +65]}

\newcommand{\DiffRTwoMinusRZeroQwenFourteen}{+28}

\newcommand{\DiffRThreeMinusRTwoQwenFourteen}{$-$4}

\newcommand{\DiscRFourIntactQwenFourteen}{+55}
\newcommand{\DiscRFiveIntactQwenFourteen}{+67}
\newcommand{\DiffNetIntactQwenFourteen}{$-$11}

\newcommand{\PHolmNetIntactQwenFourteen}{1.000}
\newcommand{\DiscRFourDropTwentyFiveQwenFourteen}{+39}
\newcommand{\DiscRFiveDropTwentyFiveQwenFourteen}{+51}
\newcommand{\DiffNetDropTwentyFiveQwenFourteen}{$-$12}

\newcommand{\DiscRFourDropFiftyQwenFourteen}{+21}
\newcommand{\DiscRFiveDropFiftyQwenFourteen}{+42}
\newcommand{\DiffNetDropFiftyQwenFourteen}{$-$21}
\newcommand{\CINetDropFiftyQwenFourteen}{[$-$38, $-$6]}
\newcommand{\PHolmNetDropFiftyQwenFourteen}{0.206}
\newcommand{\DiscRFourSpuriousTenQwenFourteen}{+53}
\newcommand{\DiscRFiveSpuriousTenQwenFourteen}{+62}
\newcommand{\DiffNetSpuriousTenQwenFourteen}{$-$10}

\newcommand{\PHolmNetSpuriousTenQwenFourteen}{1.000}
\newcommand{\DiffDeferQwenFourteen}{+34}

\newcommand{\DiffNoRecQwenFourteen}{+9}

\newcommand{\RecGainedDropFiftyQwenFourteen}{13}
\newcommand{\RecLostDropFiftyQwenFourteen}{13}

\newcommand{\DiffNetVIntactQwenFourteen}{$-$3}

\newcommand{\DiffNetVDropFiftyQwenFourteen}{$-$13}
\newcommand{\CINetVDropFiftyQwenFourteen}{[$-$28, 0]}

\newcommand{\CatchDroppedRFourQwenFourteen}{7\%}
\newcommand{\CatchDroppedRFournQwenFourteen}{16\%}
\newcommand{\NNoRecClausesQwenFourteen}{9}

\newcommand{\DiscRTwoQwenFourteen}{$-$7}
\newcommand{\FAMQwenFourteenRFourN}{30\%}
\newcommand{\FAMQwenFourteenRFourNDropFifty}{26\%}

\newcommand{\CatchDroppedQwenFourteen}{7\%}
\newcommand{\OwnCatchDroppedQwenFourteen}{41\%}
\newcommand{\NDroppedClausesQwenFourteen}{9}
\newcommand{\NSpuriousClausesQwenFourteen}{3}
\newcommand{\KNSpuriousMarkedQwenFourteen}{9}
\newcommand{\FASpuriousMarkedQwenFourteen}{9/9}
\newcommand{\OwnFASpuriousMarkedQwenFourteen}{6/9}
\newcommand{\ScAnyCatchQwenFourteenRZero}{40\%}
\newcommand{\ScClauseCatchQwenFourteenRZero}{40\%}
\newcommand{\ScAnyFAQwenFourteenRZero}{40\%}
\newcommand{\ScReasonFAQwenFourteenRZero}{40\%}
\newcommand{\ScAnyCatchQwenFourteenRTwo}{62\%}
\newcommand{\ScClauseCatchQwenFourteenRTwo}{62\%}
\newcommand{\ScAnyFAQwenFourteenRTwo}{40\%}
\newcommand{\ScReasonFAQwenFourteenRTwo}{23\%}
\newcommand{\ScAnyCatchQwenFourteenRFour}{79\%}
\newcommand{\ScClauseCatchQwenFourteenRFour}{79\%}
\newcommand{\ScAnyFAQwenFourteenRFour}{23\%}
\newcommand{\ScReasonFAQwenFourteenRFour}{15\%}
\newcommand{\CatchGemmaRZero}{0\%}

\newcommand{\FAMGemmaRZero}{40\%}

\newcommand{\FACGemmaRZero}{34\%}

\newcommand{\CatchGemmaROne}{0\%}

\newcommand{\FAMGemmaROne}{20\%}

\newcommand{\FACGemmaROne}{36\%}

\newcommand{\CatchGemmaRTwo}{24\%}

\newcommand{\FAMGemmaRTwo}{45\%}

\newcommand{\FACGemmaRTwo}{36\%}

\newcommand{\MisnamedKNGemmaRTwo}{23/32}

\newcommand{\CatchGemmaRThree}{33\%}

\newcommand{\FAMGemmaRThree}{45\%}

\newcommand{\FACGemmaRThree}{36\%}

\newcommand{\CatchGemmaRFour}{64\%}

\newcommand{\FAMGemmaRFour}{43\%}

\newcommand{\FACGemmaRFour}{39\%}

\newcommand{\CatchGemmaRFourP}{62\%}

\newcommand{\FAMGemmaRFourP}{40\%}

\newcommand{\FACGemmaRFourP}{39\%}

\newcommand{\CatchGemmaRFive}{79\%}

\newcommand{\FAMGemmaRFive}{13\%}

\newcommand{\FACGemmaRFive}{12\%}

\newcommand{\CantConfirmKNGemmaRTwoCC}{0/42}

\newcommand{\FAMGemmaRFourDropFifty}{43\%}

\newcommand{\InvalidKNGemmaRFourDropFifty}{5/630}

\newcommand{\DiffRFourMinusRTwoGemma}{+40}
\newcommand{\CIRFourMinusRTwoGemma}{[+20, +58]}

\newcommand{\DiffRTwoMinusRZeroGemma}{+26}
\newcommand{\CIRTwoMinusRZeroGemma}{[+8, +48]}

\newcommand{\ClausesRTwoMinusRZeroGemma}{13}
\newcommand{\DiffRThreeMinusRTwoGemma}{+9}

\newcommand{\DiscRFourIntactGemma}{+21}
\newcommand{\DiscRFiveIntactGemma}{+67}
\newcommand{\DiffNetIntactGemma}{$-$45}
\newcommand{\CINetIntactGemma}{[$-$69, $-$24]}
\newcommand{\PHolmNetIntactGemma}{0.043}
\newcommand{\DiscRFourDropTwentyFiveGemma}{+14}
\newcommand{\DiscRFiveDropTwentyFiveGemma}{+51}
\newcommand{\DiffNetDropTwentyFiveGemma}{$-$37}

\newcommand{\DiscRFourDropFiftyGemma}{+4}
\newcommand{\DiscRFiveDropFiftyGemma}{+42}
\newcommand{\DiffNetDropFiftyGemma}{$-$38}

\newcommand{\PHolmNetDropFiftyGemma}{0.178}
\newcommand{\DiscRFourSpuriousTenGemma}{+20}
\newcommand{\DiscRFiveSpuriousTenGemma}{+62}
\newcommand{\DiffNetSpuriousTenGemma}{$-$42}
\newcommand{\CINetSpuriousTenGemma}{[$-$66, $-$22]}
\newcommand{\PHolmNetSpuriousTenGemma}{0.043}
\newcommand{\DiffDeferGemma}{+1}

\newcommand{\DiffNoRecGemma}{+9}

\newcommand{\RecGainedDropFiftyGemma}{25}
\newcommand{\RecLostDropFiftyGemma}{24}

\newcommand{\InstrPIntactGemma}{0.031}
\newcommand{\InstrNIntactGemma}{8}

\newcommand{\InstrPSpuriousTenGemma}{0.031}

\newcommand{\DiffNetVIntactGemma}{$-$6}
\newcommand{\CINetVIntactGemma}{[$-$20, +8]}

\newcommand{\DiffNetVDropFiftyGemma}{+1}

\newcommand{\CatchDroppedRFourGemma}{31\%}
\newcommand{\CatchDroppedRFournGemma}{41\%}

\newcommand{\DiscRTwoGemma}{$-$20}
\newcommand{\FAMGemmaRFourN}{48\%}
\newcommand{\FAMGemmaRFourNDropFifty}{48\%}

\newcommand{\CatchDroppedGemma}{31\%}
\newcommand{\OwnCatchDroppedGemma}{32\%}

\newcommand{\FASpuriousMarkedGemma}{8/9}
\newcommand{\OwnFASpuriousMarkedGemma}{4/9}
\newcommand{\ScAnyCatchGemmaRZero}{50\%}
\newcommand{\ScClauseCatchGemmaRZero}{48\%}
\newcommand{\ScAnyFAGemmaRZero}{40\%}
\newcommand{\ScReasonFAGemmaRZero}{20\%}
\newcommand{\ScAnyCatchGemmaRTwo}{76\%}
\newcommand{\ScClauseCatchGemmaRTwo}{76\%}
\newcommand{\ScAnyFAGemmaRTwo}{45\%}
\newcommand{\ScReasonFAGemmaRTwo}{0\%}
\newcommand{\ScAnyCatchGemmaRFour}{79\%}
\newcommand{\ScClauseCatchGemmaRFour}{79\%}
\newcommand{\ScAnyFAGemmaRFour}{43\%}
\newcommand{\ScReasonFAGemmaRFour}{4\%}

\newcommand{\CatchFlashRTwo}{46\%}

\newcommand{\CatchFlashRFour}{82\%}

\newcommand{\FAMFlashRFour}{44\%}

\newcommand{\RobFlaggedClaudeOpusRFour}{37/42}
\newcommand{\RobNonValidityClaudeOpus}{44/55}

\newcommand{\SOneDiffNetIntactGemma}{$-$47}

\newcommand{\SOnePHolmNetIntactGemma}{0.083}

\newcommand{\SOneDiffNetSpuriousTenGemma}{$-$45}

\newcommand{\SOnePHolmNetSpuriousTenGemma}{0.083}

\newcommand{\STwoDiffNetIntactQwenFourteen}{+17}

\newcommand{\STwoDiffNetIntactGemma}{$-$50}

\providecommand{\DiscRFourIntactQwenFourteen}{\textbf{[TBD]}}
\providecommand{\DiscRFiveIntactQwenFourteen}{\textbf{[TBD]}}
\providecommand{\DiffNetIntactQwenFourteen}{\textbf{[TBD]}}

\providecommand{\PHolmNetIntactQwenFourteen}{\textbf{[TBD]}}
\providecommand{\DiscRFourDropTwentyFiveQwenFourteen}{\textbf{[TBD]}}
\providecommand{\DiscRFiveDropTwentyFiveQwenFourteen}{\textbf{[TBD]}}
\providecommand{\DiffNetDropTwentyFiveQwenFourteen}{\textbf{[TBD]}}

\providecommand{\DiscRFourDropFiftyQwenFourteen}{\textbf{[TBD]}}
\providecommand{\DiscRFiveDropFiftyQwenFourteen}{\textbf{[TBD]}}
\providecommand{\DiffNetDropFiftyQwenFourteen}{\textbf{[TBD]}}
\providecommand{\CINetDropFiftyQwenFourteen}{\textbf{[TBD]}}
\providecommand{\PHolmNetDropFiftyQwenFourteen}{\textbf{[TBD]}}
\providecommand{\DiscRFourSpuriousTenQwenFourteen}{\textbf{[TBD]}}
\providecommand{\DiscRFiveSpuriousTenQwenFourteen}{\textbf{[TBD]}}
\providecommand{\DiffNetSpuriousTenQwenFourteen}{\textbf{[TBD]}}

\providecommand{\PHolmNetSpuriousTenQwenFourteen}{\textbf{[TBD]}}
\providecommand{\CatchDroppedQwenFourteen}{\textbf{[TBD]}}
\providecommand{\OwnCatchDroppedQwenFourteen}{\textbf{[TBD]}}
\providecommand{\DiffDeferQwenFourteen}{\textbf{[TBD]}}

\providecommand{\NDroppedClausesQwenFourteen}{\textbf{[TBD]}}
\providecommand{\FASpuriousMarkedQwenFourteen}{\textbf{[TBD]}}
\providecommand{\OwnFASpuriousMarkedQwenFourteen}{\textbf{[TBD]}}
\providecommand{\KNSpuriousMarkedQwenFourteen}{\textbf{[TBD]}}
\providecommand{\ScAnyCatchQwenFourteenRZero}{\textbf{[TBD]}}
\providecommand{\ScClauseCatchQwenFourteenRZero}{\textbf{[TBD]}}
\providecommand{\ScAnyFAQwenFourteenRZero}{\textbf{[TBD]}}
\providecommand{\ScReasonFAQwenFourteenRZero}{\textbf{[TBD]}}
\providecommand{\ScAnyCatchQwenFourteenRTwo}{\textbf{[TBD]}}
\providecommand{\ScClauseCatchQwenFourteenRTwo}{\textbf{[TBD]}}
\providecommand{\ScAnyFAQwenFourteenRTwo}{\textbf{[TBD]}}
\providecommand{\ScReasonFAQwenFourteenRTwo}{\textbf{[TBD]}}
\providecommand{\ScAnyCatchQwenFourteenRFour}{\textbf{[TBD]}}
\providecommand{\ScClauseCatchQwenFourteenRFour}{\textbf{[TBD]}}
\providecommand{\ScAnyFAQwenFourteenRFour}{\textbf{[TBD]}}
\providecommand{\ScReasonFAQwenFourteenRFour}{\textbf{[TBD]}}
\providecommand{\DiscRFourIntactGemma}{\textbf{[TBD]}}
\providecommand{\DiscRFiveIntactGemma}{\textbf{[TBD]}}
\providecommand{\DiffNetIntactGemma}{\textbf{[TBD]}}
\providecommand{\CINetIntactGemma}{\textbf{[TBD]}}
\providecommand{\PHolmNetIntactGemma}{\textbf{[TBD]}}
\providecommand{\DiscRFourDropTwentyFiveGemma}{\textbf{[TBD]}}
\providecommand{\DiscRFiveDropTwentyFiveGemma}{\textbf{[TBD]}}
\providecommand{\DiffNetDropTwentyFiveGemma}{\textbf{[TBD]}}

\providecommand{\DiscRFourDropFiftyGemma}{\textbf{[TBD]}}
\providecommand{\DiscRFiveDropFiftyGemma}{\textbf{[TBD]}}
\providecommand{\DiffNetDropFiftyGemma}{\textbf{[TBD]}}

\providecommand{\PHolmNetDropFiftyGemma}{\textbf{[TBD]}}
\providecommand{\DiscRFourSpuriousTenGemma}{\textbf{[TBD]}}
\providecommand{\DiscRFiveSpuriousTenGemma}{\textbf{[TBD]}}
\providecommand{\DiffNetSpuriousTenGemma}{\textbf{[TBD]}}
\providecommand{\CINetSpuriousTenGemma}{\textbf{[TBD]}}
\providecommand{\PHolmNetSpuriousTenGemma}{\textbf{[TBD]}}
\providecommand{\CatchDroppedGemma}{\textbf{[TBD]}}
\providecommand{\OwnCatchDroppedGemma}{\textbf{[TBD]}}
\providecommand{\DiffDeferGemma}{\textbf{[TBD]}}

\providecommand{\FASpuriousMarkedGemma}{\textbf{[TBD]}}
\providecommand{\OwnFASpuriousMarkedGemma}{\textbf{[TBD]}}

\providecommand{\ScAnyCatchGemmaRZero}{\textbf{[TBD]}}
\providecommand{\ScClauseCatchGemmaRZero}{\textbf{[TBD]}}
\providecommand{\ScAnyFAGemmaRZero}{\textbf{[TBD]}}
\providecommand{\ScReasonFAGemmaRZero}{\textbf{[TBD]}}
\providecommand{\ScAnyCatchGemmaRTwo}{\textbf{[TBD]}}
\providecommand{\ScClauseCatchGemmaRTwo}{\textbf{[TBD]}}
\providecommand{\ScAnyFAGemmaRTwo}{\textbf{[TBD]}}
\providecommand{\ScReasonFAGemmaRTwo}{\textbf{[TBD]}}
\providecommand{\ScAnyCatchGemmaRFour}{\textbf{[TBD]}}
\providecommand{\ScClauseCatchGemmaRFour}{\textbf{[TBD]}}
\providecommand{\ScAnyFAGemmaRFour}{\textbf{[TBD]}}
\providecommand{\ScReasonFAGemmaRFour}{\textbf{[TBD]}}

\providecommand{\NDamageSeedsCapable}{\textbf{[TBD]}}
\providecommand{\NCandidateCases}{\textbf{[TBD]}}
\providecommand{\NDroppedDecision}{\textbf{[TBD]}}
\providecommand{\NClauseDisputed}{\textbf{[TBD]}}
\providecommand{\NStrictKept}{\textbf{[TBD]}}
\providecommand{\NControlsChecked}{\textbf{[TBD]}}

\providecommand{\PctControlsCorrect}{\textbf{[TBD]}}
\providecommand{\NLinksRegistry}{\textbf{[TBD]}} %
\providecommand{\NLinksSample}{\textbf{[TBD]}} %
\providecommand{\NLinksSampleConfirmed}{\textbf{[TBD]}} %
\providecommand{\NLinksSampleRelChanged}{\textbf{[TBD]}} %
\providecommand{\NLinksSampleNotALink}{\textbf{[TBD]}} %
\providecommand{\CINotALinkSample}{\textbf{[TBD]}} %
\providecommand{\NLinksTestPairNotALink}{\textbf{[TBD]}} %
\providecommand{\NLinksTestPairRelChanged}{\textbf{[TBD]}} %
\providecommand{\NDroppedDecisionOnly}{\textbf{[TBD]}} %
\providecommand{\NDroppedBlockingStage}{\textbf{[TBD]}} %
\providecommand{\NControlsCorrect}{\textbf{[TBD]}} %
\providecommand{\ControlsGatePct}{\textbf{[TBD]}} %
\providecommand{\ScorerCheckAgreeN}{\textbf{[TBD]}} %
\providecommand{\ScorerCheckPctObserved}{\textbf{[TBD]}} %
\providecommand{\NScorerDisagree}{\textbf{[TBD]}} %
\providecommand{\NScorerClauseReason}{\textbf{[TBD]}} %
\providecommand{\NScorerCatchReason}{\textbf{[TBD]}} %
\providecommand{\NPrimaryTestsVFive}{\textbf{[TBD]}} %
\providecommand{\DiscRTwoQwenFourteen}{\textbf{[TBD]}} %
\providecommand{\DiscRTwoGemma}{\textbf{[TBD]}} %
\providecommand{\DiffNoRecQwenFourteen}{\textbf{[TBD]}} %
\providecommand{\DiffNoRecGemma}{\textbf{[TBD]}} %
\providecommand{\FAMQwenFourteenRFourN}{\textbf{[TBD]}} %
\providecommand{\FAMGemmaRFourN}{\textbf{[TBD]}} %
\providecommand{\FAMQwenFourteenRFourNDropFifty}{\textbf{[TBD]}} %
\providecommand{\FAMGemmaRFourNDropFifty}{\textbf{[TBD]}} %
\providecommand{\NRepeatItems}{\textbf{[TBD]}}
\providecommand{\NRepeatSameFindings}{\textbf{[TBD]}}

\newcommand{\RfcSample}{48}

\newcommand{\RfcTitleDifferentBothTextsGemmaFourTwoSixB}{1/6}

\newcommand{\RfcTitleDifferentBothTextsClaudeOpusFourFive}{3/6}

\newcommand{\RfcDeferOwnStatusDamagedClaudeOpusFourFive}{11}
\newcommand{\RfcDeferWithStatusDamagedClaudeOpusFourFive}{10}

\newcommand{\RfcCatchBothMaskedClaudeOpusFourFive}{20/24}

\newcommand{\RfcTitleDifferentBothTextsGeminiThreeOnePro}{5/6}

\newcommand{\RfcCatchBothMaskedGeminiThreeOnePro}{20/24}

\newcommand{\RfcDiscDateBothTextsGeminiThreeOnePro}{+92}

\newcommand{\RfcCatchCiteOnlyGeminiThreeOnePro}{11/24}

\newcommand{\RfcTitleDifferentBothTextsGPTFiveFive}{6/6}

\newcommand{\RfcCatchBothMaskedGPTFiveFive}{22/24}

\newcommand{\RfcTitleDifferentBothTextsGPTOSSOneTwoZeroB}{2/6}

\newcommand{\RfcCatchBothMaskedGPTOSSOneTwoZeroB}{13/24}

\newcommand{\RfcDiscDateBothTextsQwenThreeTwoB}{$-$2}

\newcommand{\RfcCatchBothTextsQwenThreeTwoB}{42/48}

\newcommand{\RfcFABothTextsQwenThreeTwoB}{23/48}

\newcommand{\RfcDateFABothTextsQwenThreeTwoB}{43/48}

\newcommand{\RfcDateFAFewShotQwenThreeTwoB}{5/48}

\newcommand{\EComplete}{1}
\newcommand{\ENLinks}{48}
\newcommand{\EBalArea}{37/48}
\newcommand{\EBalTitleRatioSMD}{$-$0.22}
\newcommand{\EBalOldYearSMD}{0.04}
\newcommand{\EBalNewYearSMD}{0.02}
\newcommand{\EBalGapYearsSMD}{$-$0.02}

\newcommand{\EBalNewWordsSMD}{$-$0.55}
\newcommand{\EGemmaCatchTexts}{31/48}
\newcommand{\EGemmaCatchAnyTexts}{31/48}
\newcommand{\EGemmaCtrlFATexts}{0/48}
\newcommand{\EGemmaCtrlAnyTexts}{19/48}
\newcommand{\EGemmaDateFATexts}{0/48}
\newcommand{\EGemmaDecoyFATexts}{6/48}
\newcommand{\EGemmaDecoyAnyTexts}{6/48}
\newcommand{\EGemmaGapTexts}{13/24}

\newcommand{\EGemmaCatchRecord}{47/48}
\newcommand{\EGemmaCatchAnyRecord}{47/48}
\newcommand{\EGemmaCtrlFARecord}{0/48}
\newcommand{\EGemmaCtrlAnyRecord}{20/48}
\newcommand{\EGemmaDateFARecord}{0/48}
\newcommand{\EGemmaDecoyFARecord}{0/48}
\newcommand{\EGemmaDecoyAnyRecord}{0/48}

\newcommand{\EGemmaCatchDamaged}{23/48}
\newcommand{\EGemmaCatchAnyDamaged}{23/48}
\newcommand{\EGemmaCtrlFADamaged}{0/48}
\newcommand{\EGemmaCtrlAnyDamaged}{20/48}
\newcommand{\EGemmaDateFADamaged}{0/48}
\newcommand{\EGemmaDecoyFADamaged}{0/48}
\newcommand{\EGemmaDecoyAnyDamaged}{0/48}
\newcommand{\EGemmaGapDamaged}{0/24}
\newcommand{\EGemmaBareDamaged}{24}
\newcommand{\EGemmaCatchNorecord}{23/48}
\newcommand{\EGemmaCatchAnyNorecord}{23/48}
\newcommand{\EGemmaCtrlFANorecord}{0/48}
\newcommand{\EGemmaCtrlAnyNorecord}{20/48}
\newcommand{\EGemmaDateFANorecord}{0/48}
\newcommand{\EGemmaDecoyFANorecord}{0/48}
\newcommand{\EGemmaDecoyAnyNorecord}{0/48}
\newcommand{\EGemmaGapNorecord}{0/24}

\newcommand{\EGemmaCatchLowauth}{26/48}
\newcommand{\EGemmaCatchAnyLowauth}{26/48}
\newcommand{\EGemmaCtrlFALowauth}{0/48}
\newcommand{\EGemmaCtrlAnyLowauth}{18/48}
\newcommand{\EGemmaDateFALowauth}{0/48}
\newcommand{\EGemmaDecoyFALowauth}{1/48}
\newcommand{\EGemmaDecoyAnyLowauth}{1/48}
\newcommand{\EGemmaGapLowauth}{2/24}

\newcommand{\EGemmaCatchCaveat}{29/48}
\newcommand{\EGemmaCatchAnyCaveat}{29/48}
\newcommand{\EGemmaCtrlFACaveat}{0/48}
\newcommand{\EGemmaCtrlAnyCaveat}{18/48}
\newcommand{\EGemmaDateFACaveat}{0/48}
\newcommand{\EGemmaDecoyFACaveat}{1/48}
\newcommand{\EGemmaDecoyAnyCaveat}{1/48}
\newcommand{\EGemmaGapCaveat}{5/24}

\newcommand{\EGemmaCatchConflict}{23/48}
\newcommand{\EGemmaCatchAnyConflict}{26/48}
\newcommand{\EGemmaCtrlFAConflict}{0/48}
\newcommand{\EGemmaCtrlAnyConflict}{20/48}
\newcommand{\EGemmaDateFAConflict}{0/48}
\newcommand{\EGemmaDecoyFAConflict}{0/48}
\newcommand{\EGemmaDecoyAnyConflict}{1/48}
\newcommand{\EGemmaGapConflict}{0/24}
\newcommand{\EGemmaBareConflict}{21}
\newcommand{\EGemmaHOne}{+52}
\newcommand{\EGemmaHOneCI}{[+35, +69]}
\newcommand{\EGemmaSameCatch}{29/33}
\newcommand{\EGemmaSameDecoy}{6/33}
\newcommand{\EGemmaDifferentCatch}{2/15}
\newcommand{\EGemmaDifferentDecoy}{0/15}
\newcommand{\EGemmaHTwoaLost}{13}
\newcommand{\EGemmaHTwoaGained}{0}

\newcommand{\EGemmaHTwoaHolm}{<.001}
\newcommand{\EGemmaReaskChanged}{0}

\newcommand{\EGemmaConflictCodeGap}{3}

\newcommand{\EGemmaArmNorecordHolm}{1.000}

\newcommand{\EGemmaArmLowauthHolm}{1.000}
\newcommand{\EGemmaArmCaveatGained}{5}

\newcommand{\EGemmaArmCaveatHolm}{.375}
\newcommand{\EGPTCatchTexts}{46/48}
\newcommand{\EGPTCatchAnyTexts}{46/48}
\newcommand{\EGPTCtrlFATexts}{1/48}
\newcommand{\EGPTCtrlAnyTexts}{10/48}
\newcommand{\EGPTDateFATexts}{0/48}
\newcommand{\EGPTDecoyFATexts}{8/48}
\newcommand{\EGPTDecoyAnyTexts}{8/48}
\newcommand{\EGPTGapTexts}{23/24}

\newcommand{\EGPTCatchRecord}{48/48}
\newcommand{\EGPTCatchAnyRecord}{48/48}
\newcommand{\EGPTCtrlFARecord}{0/48}
\newcommand{\EGPTCtrlAnyRecord}{12/48}
\newcommand{\EGPTDateFARecord}{0/48}
\newcommand{\EGPTDecoyFARecord}{0/48}
\newcommand{\EGPTDecoyAnyRecord}{0/48}

\newcommand{\EGPTCatchDamaged}{33/48}
\newcommand{\EGPTCatchAnyDamaged}{33/48}
\newcommand{\EGPTCtrlFADamaged}{0/48}
\newcommand{\EGPTCtrlAnyDamaged}{12/48}
\newcommand{\EGPTDateFADamaged}{0/48}
\newcommand{\EGPTDecoyFADamaged}{0/48}
\newcommand{\EGPTDecoyAnyDamaged}{0/48}
\newcommand{\EGPTGapDamaged}{9/24}
\newcommand{\EGPTBareDamaged}{15}
\newcommand{\EGPTCatchNorecord}{34/48}
\newcommand{\EGPTCatchAnyNorecord}{34/48}
\newcommand{\EGPTCtrlFANorecord}{0/48}
\newcommand{\EGPTCtrlAnyNorecord}{10/48}
\newcommand{\EGPTDateFANorecord}{0/48}
\newcommand{\EGPTDecoyFANorecord}{0/48}
\newcommand{\EGPTDecoyAnyNorecord}{0/48}
\newcommand{\EGPTGapNorecord}{10/24}

\newcommand{\EGPTCatchLowauth}{48/48}
\newcommand{\EGPTCatchAnyLowauth}{48/48}
\newcommand{\EGPTCtrlFALowauth}{1/48}
\newcommand{\EGPTCtrlAnyLowauth}{11/48}
\newcommand{\EGPTDateFALowauth}{0/48}
\newcommand{\EGPTDecoyFALowauth}{10/48}
\newcommand{\EGPTDecoyAnyLowauth}{10/48}
\newcommand{\EGPTGapLowauth}{24/24}

\newcommand{\EGPTCatchCaveat}{47/48}
\newcommand{\EGPTCatchAnyCaveat}{48/48}
\newcommand{\EGPTCtrlFACaveat}{1/48}
\newcommand{\EGPTCtrlAnyCaveat}{9/48}
\newcommand{\EGPTDateFACaveat}{0/48}
\newcommand{\EGPTDecoyFACaveat}{8/48}
\newcommand{\EGPTDecoyAnyCaveat}{8/48}
\newcommand{\EGPTGapCaveat}{23/24}

\newcommand{\EGPTCatchConflict}{31/48}
\newcommand{\EGPTCatchAnyConflict}{33/48}
\newcommand{\EGPTCtrlFAConflict}{0/48}
\newcommand{\EGPTCtrlAnyConflict}{10/48}
\newcommand{\EGPTDateFAConflict}{0/48}
\newcommand{\EGPTDecoyFAConflict}{2/48}
\newcommand{\EGPTDecoyAnyConflict}{2/48}
\newcommand{\EGPTGapConflict}{7/24}
\newcommand{\EGPTBareConflict}{15}
\newcommand{\EGPTHOne}{+79}
\newcommand{\EGPTHOneCI}{[+67, +90]}
\newcommand{\EGPTSameCatch}{33/33}
\newcommand{\EGPTSameDecoy}{8/33}
\newcommand{\EGPTDifferentCatch}{13/15}
\newcommand{\EGPTDifferentDecoy}{0/15}
\newcommand{\EGPTHTwoaLost}{14}
\newcommand{\EGPTHTwoaGained}{0}

\newcommand{\EGPTHTwoaHolm}{<.001}
\newcommand{\EGPTReaskChanged}{7}
\newcommand{\EGPTReaskItems}{1,152}
\newcommand{\EGPTConflictCodeGap}{2}

\newcommand{\EGPTArmNorecordGained}{2}
\newcommand{\EGPTArmNorecordLost}{1}

\newcommand{\EGPTArmNorecordHolm}{1.000}
\newcommand{\EGPTArmLowauthGained}{15}

\newcommand{\EGPTArmLowauthHolm}{<.001}
\newcommand{\EGPTArmCaveatGained}{14}

\newcommand{\EGPTArmCaveatHolm}{.001}
\newcommand{\EGeminiCatchTexts}{43/48}
\newcommand{\EGeminiCatchAnyTexts}{43/48}
\newcommand{\EGeminiCtrlFATexts}{0/48}
\newcommand{\EGeminiCtrlAnyTexts}{22/48}
\newcommand{\EGeminiDateFATexts}{0/48}
\newcommand{\EGeminiDecoyFATexts}{6/48}
\newcommand{\EGeminiDecoyAnyTexts}{6/48}
\newcommand{\EGeminiGapTexts}{23/24}

\newcommand{\EGeminiCatchRecord}{48/48}
\newcommand{\EGeminiCatchAnyRecord}{48/48}
\newcommand{\EGeminiCtrlFARecord}{0/48}
\newcommand{\EGeminiCtrlAnyRecord}{19/48}
\newcommand{\EGeminiDateFARecord}{0/48}
\newcommand{\EGeminiDecoyFARecord}{0/48}
\newcommand{\EGeminiDecoyAnyRecord}{0/48}

\newcommand{\EGeminiCatchDamaged}{25/48}
\newcommand{\EGeminiCatchAnyDamaged}{25/48}
\newcommand{\EGeminiCtrlFADamaged}{0/48}
\newcommand{\EGeminiCtrlAnyDamaged}{19/48}
\newcommand{\EGeminiDateFADamaged}{0/48}
\newcommand{\EGeminiDecoyFADamaged}{0/48}
\newcommand{\EGeminiDecoyAnyDamaged}{0/48}
\newcommand{\EGeminiGapDamaged}{1/24}
\newcommand{\EGeminiBareDamaged}{23}
\newcommand{\EGeminiCatchNorecord}{30/48}
\newcommand{\EGeminiCatchAnyNorecord}{30/48}
\newcommand{\EGeminiCtrlFANorecord}{0/48}
\newcommand{\EGeminiCtrlAnyNorecord}{18/48}
\newcommand{\EGeminiDateFANorecord}{0/48}
\newcommand{\EGeminiDecoyFANorecord}{0/48}
\newcommand{\EGeminiDecoyAnyNorecord}{0/48}
\newcommand{\EGeminiGapNorecord}{6/24}

\newcommand{\EGeminiCatchLowauth}{42/48}
\newcommand{\EGeminiCatchAnyLowauth}{42/48}
\newcommand{\EGeminiCtrlFALowauth}{0/48}
\newcommand{\EGeminiCtrlAnyLowauth}{19/48}
\newcommand{\EGeminiDateFALowauth}{0/48}
\newcommand{\EGeminiDecoyFALowauth}{4/48}
\newcommand{\EGeminiDecoyAnyLowauth}{4/48}
\newcommand{\EGeminiGapLowauth}{18/24}

\newcommand{\EGeminiCatchCaveat}{46/48}
\newcommand{\EGeminiCatchAnyCaveat}{46/48}
\newcommand{\EGeminiCtrlFACaveat}{0/48}
\newcommand{\EGeminiCtrlAnyCaveat}{20/48}
\newcommand{\EGeminiDateFACaveat}{0/48}
\newcommand{\EGeminiDecoyFACaveat}{6/48}
\newcommand{\EGeminiDecoyAnyCaveat}{6/48}
\newcommand{\EGeminiGapCaveat}{22/24}

\newcommand{\EGeminiCatchConflict}{24/48}
\newcommand{\EGeminiCatchAnyConflict}{38/48}
\newcommand{\EGeminiCtrlFAConflict}{0/48}
\newcommand{\EGeminiCtrlAnyConflict}{20/48}
\newcommand{\EGeminiDateFAConflict}{0/48}
\newcommand{\EGeminiDecoyFAConflict}{0/48}
\newcommand{\EGeminiDecoyAnyConflict}{1/48}
\newcommand{\EGeminiGapConflict}{0/24}
\newcommand{\EGeminiBareConflict}{10}
\newcommand{\EGeminiHOne}{+77}
\newcommand{\EGeminiHOneCI}{[+65, +88]}
\newcommand{\EGeminiSameCatch}{32/33}
\newcommand{\EGeminiSameDecoy}{6/33}
\newcommand{\EGeminiDifferentCatch}{11/15}
\newcommand{\EGeminiDifferentDecoy}{0/15}
\newcommand{\EGeminiHTwoaLost}{22}
\newcommand{\EGeminiHTwoaGained}{0}

\newcommand{\EGeminiHTwoaHolm}{<.001}
\newcommand{\EGeminiReaskChanged}{8}

\newcommand{\EGeminiConflictCodeGap}{14}

\newcommand{\EGeminiArmNorecordGained}{5}
\newcommand{\EGeminiArmNorecordLost}{0}

\newcommand{\EGeminiArmNorecordHolm}{.375}
\newcommand{\EGeminiArmLowauthGained}{17}

\newcommand{\EGeminiArmLowauthHolm}{<.001}
\newcommand{\EGeminiArmCaveatGained}{21}

\newcommand{\EGeminiArmCaveatHolm}{<.001}
\newcommand{\EOpusCatchTexts}{37/48}
\newcommand{\EOpusCatchAnyTexts}{38/48}
\newcommand{\EOpusCtrlFATexts}{0/48}
\newcommand{\EOpusCtrlAnyTexts}{19/48}
\newcommand{\EOpusDateFATexts}{3/48}
\newcommand{\EOpusDecoyFATexts}{9/48}
\newcommand{\EOpusDecoyAnyTexts}{9/48}
\newcommand{\EOpusGapTexts}{17/24}

\newcommand{\EOpusCatchRecord}{48/48}
\newcommand{\EOpusCatchAnyRecord}{48/48}
\newcommand{\EOpusCtrlFARecord}{0/48}
\newcommand{\EOpusCtrlAnyRecord}{22/48}
\newcommand{\EOpusDateFARecord}{0/48}
\newcommand{\EOpusDecoyFARecord}{4/48}
\newcommand{\EOpusDecoyAnyRecord}{4/48}

\newcommand{\EOpusCatchDamaged}{33/48}
\newcommand{\EOpusCatchAnyDamaged}{35/48}
\newcommand{\EOpusCtrlFADamaged}{0/48}
\newcommand{\EOpusCtrlAnyDamaged}{22/48}
\newcommand{\EOpusDateFADamaged}{0/48}
\newcommand{\EOpusDecoyFADamaged}{4/48}
\newcommand{\EOpusDecoyAnyDamaged}{4/48}
\newcommand{\EOpusGapDamaged}{9/24}
\newcommand{\EOpusBareDamaged}{13}
\newcommand{\EOpusCatchNorecord}{35/48}
\newcommand{\EOpusCatchAnyNorecord}{36/48}
\newcommand{\EOpusCtrlFANorecord}{1/48}
\newcommand{\EOpusCtrlAnyNorecord}{22/48}
\newcommand{\EOpusDateFANorecord}{0/48}
\newcommand{\EOpusDecoyFANorecord}{5/48}
\newcommand{\EOpusDecoyAnyNorecord}{5/48}
\newcommand{\EOpusGapNorecord}{11/24}

\newcommand{\EOpusCatchLowauth}{41/48}
\newcommand{\EOpusCatchAnyLowauth}{42/48}
\newcommand{\EOpusCtrlFALowauth}{0/48}
\newcommand{\EOpusCtrlAnyLowauth}{20/48}
\newcommand{\EOpusDateFALowauth}{1/48}
\newcommand{\EOpusDecoyFALowauth}{9/48}
\newcommand{\EOpusDecoyAnyLowauth}{9/48}
\newcommand{\EOpusGapLowauth}{17/24}

\newcommand{\EOpusCatchCaveat}{41/48}
\newcommand{\EOpusCatchAnyCaveat}{42/48}
\newcommand{\EOpusCtrlFACaveat}{0/48}
\newcommand{\EOpusCtrlAnyCaveat}{20/48}
\newcommand{\EOpusDateFACaveat}{1/48}
\newcommand{\EOpusDecoyFACaveat}{7/48}
\newcommand{\EOpusDecoyAnyCaveat}{7/48}
\newcommand{\EOpusGapCaveat}{17/24}

\newcommand{\EOpusCatchConflict}{37/48}
\newcommand{\EOpusCatchAnyConflict}{38/48}
\newcommand{\EOpusCtrlFAConflict}{1/48}
\newcommand{\EOpusCtrlAnyConflict}{21/48}
\newcommand{\EOpusDateFAConflict}{0/48}
\newcommand{\EOpusDecoyFAConflict}{5/48}
\newcommand{\EOpusDecoyAnyConflict}{5/48}
\newcommand{\EOpusGapConflict}{13/24}
\newcommand{\EOpusBareConflict}{10}
\newcommand{\EOpusHOne}{+58}
\newcommand{\EOpusHOneCI}{[+42, +73]}
\newcommand{\EOpusSameCatch}{30/33}
\newcommand{\EOpusSameDecoy}{9/33}
\newcommand{\EOpusDifferentCatch}{7/15}
\newcommand{\EOpusDifferentDecoy}{0/15}
\newcommand{\EOpusHTwoaLost}{8}
\newcommand{\EOpusHTwoaGained}{0}

\newcommand{\EOpusHTwoaHolm}{.008}
\newcommand{\EOpusReaskChanged}{12}
\newcommand{\EOpusReaskItems}{1,152}

\newcommand{\EOpusArmNorecordGained}{3}
\newcommand{\EOpusArmNorecordLost}{1}

\newcommand{\EOpusArmNorecordHolm}{1.000}
\newcommand{\EOpusArmLowauthGained}{8}

\newcommand{\EOpusArmLowauthHolm}{.062}

\newcommand{\EOpusArmCaveatHolm}{.062}
\newcommand{\EAbsSimLinksSame}{0.80}
\newcommand{\EAbsSimLinksDifferent}{0.15}
\newcommand{\EAbsSimDecoysSame}{0.17}
\newcommand{\EAbsSimDecoysDifferent}{0.10}
\newcommand{\EGemmaDecoyFASimHigh}{2/7}
\newcommand{\EGemmaDecoyFASimLow}{4/26}
\newcommand{\EGemmaDiffCatchSimHigh}{2/6}
\newcommand{\EGemmaDiffCatchSimLow}{0/9}

\newcommand{\EGemmaMissDamaged}{24}

\newcommand{\EGPTDecoyFASimHigh}{3/7}
\newcommand{\EGPTDecoyFASimLow}{5/26}
\newcommand{\EGPTDiffCatchSimHigh}{6/6}
\newcommand{\EGPTDiffCatchSimLow}{7/9}

\newcommand{\EGPTMissDamaged}{15}

\newcommand{\EGeminiDecoyFASimHigh}{2/7}
\newcommand{\EGeminiDecoyFASimLow}{4/26}
\newcommand{\EGeminiDiffCatchSimHigh}{6/6}
\newcommand{\EGeminiDiffCatchSimLow}{5/9}

\newcommand{\EGeminiMissDamaged}{23}

\newcommand{\EOpusDecoyFASimHigh}{3/7}
\newcommand{\EOpusDecoyFASimLow}{6/26}
\newcommand{\EOpusDiffCatchSimHigh}{5/6}
\newcommand{\EOpusDiffCatchSimLow}{2/9}
\newcommand{\EOpusPilotPilotItems}{12}
\newcommand{\EOpusPilotShared}{5}

\newcommand{\EOpusPilotSharedCaught}{2}
\newcommand{\EOpusPilotPilotOnly}{7}

\newcommand{\EOpusMissDamaged}{15}
\newcommand{\EOpusArmLowauthRaw}{.008}

\newcommand{\EHTwobTests}{12}
\newcommand{\EGemmaSOneOwn}{17/24}
\newcommand{\EGemmaSOneWith}{0/24}
\newcommand{\EGemmaSOneLost}{17}
\newcommand{\EGemmaSOneGained}{0}
\newcommand{\EGemmaSOneHolm}{<.001}
\newcommand{\EGemmaSOneShared}{12}
\newcommand{\EGemmaPoolLinks}{36}
\newcommand{\EGemmaPoolMean}{+61}
\newcommand{\EGemmaPoolCI}{[+44, +78]}
\newcommand{\EGemmaPoolLost}{22}
\newcommand{\EGemmaPoolGained}{0}
\newcommand{\EGemmaPoolP}{<.001}
\newcommand{\EGPTSOneOwn}{24/24}
\newcommand{\EGPTSOneWith}{10/24}
\newcommand{\EGPTSOneLost}{14}
\newcommand{\EGPTSOneGained}{0}
\newcommand{\EGPTSOneHolm}{<.001}
\newcommand{\EGPTSOneShared}{12}
\newcommand{\EGPTPoolLinks}{36}
\newcommand{\EGPTPoolMean}{+61}
\newcommand{\EGPTPoolCI}{[+44, +75]}
\newcommand{\EGPTPoolLost}{22}
\newcommand{\EGPTPoolGained}{0}
\newcommand{\EGPTPoolP}{<.001}
\newcommand{\EGeminiSOneOwn}{22/24}
\newcommand{\EGeminiSOneWith}{4/24}
\newcommand{\EGeminiSOneLost}{18}
\newcommand{\EGeminiSOneGained}{0}
\newcommand{\EGeminiSOneHolm}{<.001}
\newcommand{\EGeminiSOneShared}{12}
\newcommand{\EGeminiPoolLinks}{36}
\newcommand{\EGeminiPoolMean}{+81}
\newcommand{\EGeminiPoolCI}{[+67, +92]}
\newcommand{\EGeminiPoolLost}{29}
\newcommand{\EGeminiPoolGained}{0}
\newcommand{\EGeminiPoolP}{<.001}
\newcommand{\EOpusSOneOwn}{19/24}
\newcommand{\EOpusSOneWith}{12/24}
\newcommand{\EOpusSOneLost}{7}
\newcommand{\EOpusSOneGained}{0}
\newcommand{\EOpusSOneHolm}{.016}
\newcommand{\EOpusSOneShared}{12}
\newcommand{\EOpusPoolLinks}{36}
\newcommand{\EOpusPoolMean}{+31}
\newcommand{\EOpusPoolCI}{[+17, +44]}
\newcommand{\EOpusPoolLost}{11}
\newcommand{\EOpusPoolGained}{0}
\newcommand{\EOpusPoolP}{<.001}

\newcommand{\cfg}[1]{\texttt{#1}}
\newcommand{\rung}[1]{\textsf{#1}}
\newcommand{\Rfourp}{\rung{R4$'$}}

\newsavebox{\authorsbox}
\newcommand{\authorsrow}{\sbox{\authorsbox}{\authorstable}%
\ifdim\wd\authorsbox>\dimexpr\hsize-\leftskip-\rightskip\relax\resizebox{\dimexpr\hsize-\leftskip-\rightskip\relax}{!}{\usebox{\authorsbox}}\else\usebox{\authorsbox}\fi}
\newcommand{\authorstable}{{\normalfont\small\setlength{\tabcolsep}{8pt}%
\begin{tabular}{@{}ccc@{}}
\textbf{Unnati Shah}$^{1}$ & \textbf{Samruddhi Kakade}$^{2}$ & \textbf{Aum Patel}$^{3}$
\end{tabular}}}
\newcommand{\emailUS}{unnati.ix@gmail.com}
\newcommand{\emailAP}{aumrp77@gmail.com}
\makeatletter\newcommand{\plainfntext}{\long\def\@makefntext##1{\noindent##1}}\makeatother
\newcommand{\authornotes}{{\renewcommand{\thefootnote}{}\plainfntext\footnotetext{\raggedright
$^{1}$M.S. in Computer Science, University of Southern California. \texttt{\emailUS} (corresponding author)\\
$^{2}$M.S. in Computer Science, Illinois Institute of Technology. \texttt{samruddhikakade2001@gmail.com}\\
$^{3}$M.S. in Computer Science, New York University. \texttt{\emailAP}\\
This research was conducted independently and was not commissioned, funded, supervised or endorsed by any
university or employer.}}}
\ifdefined\anonymous\newcommand{\repourl}{\anonurl}\else\newcommand{\repourl}{\url{https://github.com/unnatibshah/what-do-llm-judges-verify}}\fi
\begin{document}
\ifdefined\anonymous
\title[What Do LLM Judges Actually Verify?]{What Do LLM Judges Actually Verify? Matched Controls and Record Gaps for
Document Currency}
\author{\fnm{Anonymous} \sur{Authors}}
\affil{\orgname{Anonymous for review}}
\else
\title[What Do LLM Judges Actually Verify?]{What Do LLM Judges Actually Verify? Matched Controls and Record Gaps for
Document Currency\par\vspace{1.2em}\authorsrow}
\hypersetup{pdftitle={What Do LLM Judges Actually Verify? Matched Controls and Record Gaps for Document Currency},
pdfauthor={Unnati Shah, Samruddhi Kakade, Aum Patel}}
\fi
\abstract{LLM judges often check, against an amendment record, whether the documents an answer cites are still in force. Such records are rarely complete. From a table of replacements, our test generates matched answers citing a replaced document, its
successor, the replaced document before its replacement, and an older document that a newer one updated without
replacing; it removes links from the record and scores each finding by its stated reason. On \ENLinks{}
replacements from the IETF RFC index, judges given a correct record caught nearly every replaced citation, as a lookup would. On the 24 faults whose link we removed, every judge caught more from the texts alone
than with the record: Gemini~3.1~Pro \EGeminiGapTexts{} against \EGeminiGapDamaged, GPT-5.5
\EGPTGapTexts{} against \EGPTGapDamaged, Claude Opus~4.5 \EOpusGapTexts{} against \EOpusGapDamaged{} and Gemma~4~26B
\EGemmaGapTexts{} against \EGemmaGapDamaged. The drop recurred on a second draw of removed links.
Most misses carried no finding; in an exploratory run with a conflict code, only Gemini used it often. Rewording the gap
as ``no replacement recorded'' did not help; presenting the record as an unverified note, or warning that it was incomplete, restored most lost catches
for Gemini, GPT-5.5 and Claude (significant after correction for the first two) but at most \EGemmaArmCaveatGained{} of \EGemmaHTwoaLost{} for
Gemma; where catches returned, so did false alarms on updates. With both texts, judges mostly told replacements from updates;
every update flagged had a near-identical title. A descriptive case study on Indian rooftop-solar rules is appended; we release the
harness.
}
\maketitle
\ifdefined\anonymous\else\authornotes\fi

\ifnum\EComplete=0\relax
\noindent\fbox{\parbox{\dimexpr\linewidth-2\fboxsep-2\fboxrule}{\small\color{red}
\textbf{INTERIM DRAFT.} Not every judge has finished the final runs; values marked [pending] are not yet run, and
statements comparing judges will be rechecked when they are. Not for circulation outside the authors.}}
\fi

\section{Introduction}
\label{sec:intro}

Contracts are amended by later agreements, regulations by later notifications, and technical standards by later
standards. An answer drawn from such a collection can quote a provision correctly and still be wrong, because the
document it quotes was replaced before the date the question is about. Systems that use language models
for this can add a second model, an LLM judge, to check the answers before they reach a user. The judge is
often given help: an amendment record listing which document replaced which, and from when.

Our starting point is a practical worry about that arrangement. Amendment records are seldom complete. Consolidated
versions lag behind new notifications; pipelines that extract links from documents miss some; registers cover only the
instruments they were built from. Wherever a record is silent about a replacement, it effectively reports a replaced
document as current. If the judge reads the documents in front of it, such gaps cost little. If it relies on the record,
it will pass exactly the answers the record cannot catch, and it adds little beyond what a lookup in the record would do.

Existing evaluations of LLM judges do not tell a builder which of these to expect. Benchmarks give the judge correct
context or none \citep{tan2025judgebench,xu2025contextualjudgebench}, and most report whether the judge flagged an
answer, not why. A judge that flags every older citation would score well on detection, and so would one that flags
any older document that sits next to a newer one on the same subject. Neither is checking currency.

We therefore built a test aimed at one error, citing a replaced document, and at the question of what the judge relies
on when the record and the documents disagree. The test starts from any table of replacements. For each replacement it
writes an answer citing the replaced document (a fault) and three answers that are correct: one citing the successor,
one citing the replaced document at a date before the replacement existed (a date control), and one citing an older
document that a newer one updated without replacing (a decoy; Figure~\ref{fig:design} shows one link). It then removes half the links from the record, presents
the damaged record in several ways that leave its content unchanged, and counts a fault as caught only when the judge
says the cited document was replaced.

We applied the test to \ENLinks{} replacements recorded by the RFC Editor of the Internet Engineering Task Force (IETF), with four judges (Gemini~3.1~Pro,
GPT-5.5, Claude Opus~4.5 and the open-weight Gemma~4~26B). The hypotheses, tests and the way we would read each outcome
were committed to the project's private development repository (the release lists the commit times) after a smaller pilot (mostly on 24 of the links) and before these runs.

With a correct record line, the judges caught every replaced citation or nearly so, which a lookup in the record
also does. On the faults whose link we removed, every judge caught fewer with the record present than from the texts
alone, on each of two random choices of which links to remove, and most of the missed answers were passed without
any finding. The drop also occurred when the line said only that no replacement was recorded, a statement that is
true. Presenting the record as an unverified note, or warning that it was incomplete, returned Gemini, GPT-5.5 and
Claude most of the way to their text-only behaviour, false alarms included; the change is significant after correction
for Gemini and GPT-5.5 but not for Claude, which had fewer catches to recover, and Gemma recovered at most \EGemmaArmCaveatGained{} of the \EGemmaHTwoaLost{}
it had lost (\S\ref{sec:res-gaps}). Offered a reason code for a record that conflicts with the evidence, in an
exploratory context added afterwards, Gemini named the conflict on \EGeminiConflictCodeGap{} of the 24 gap faults and
the other judges rarely or never did, so the silence is partly a matter of reply format for one judge only. Given both texts and no record, the judges caught many more replacements than they
flagged updates, but every update they flagged had a near-identical title, and an unplanned analysis suggests that
they treat the resemblance between two documents as evidence of replacement.
In the pilot, date controls exposed a 32B open model that flagged almost every old citation whatever its date; the
judges here passed nearly all of them.

The paper's contribution is the test, released as a small harness, together with this evidence about four judges on
one corpus. We do not claim that LLM judges in general defer to records, and we cannot say why the judges we tested
did. A second setting, Indian rooftop-solar regulation with an amendment record extracted by an AI pipeline, is
reported as a descriptive case study in the appendix: its answer key did not pass our own quality checks, and nothing
in the main results rests on it.

\section{Related work}
\label{sec:related}

\paragraph{How LLM judges are evaluated.}
Work on LLM judges has documented position, verbosity and self-preference biases \citep{zheng2023judging} and built
benchmarks of harder judgements. JudgeBench \citep{tan2025judgebench} pairs responses that differ in correctness, and
ContextualJudgeBench \citep{xu2025contextualjudgebench} asks judges to assess responses against supplied context,
which it treats as the reference. Other studies probe what judges are sensitive to: they can be stable under edits
that should not matter and insensitive to edits that should \citep{chen2026judgeknow}, and citation judges accept
claims that their passage does not warrant \citep{qian2026relevant}. What counts as an error, or how it is
scored, changes rankings: of citation verifiers \citep{kim2026hallmark} and of the models a benchmark evaluates
\citep{pall2026scoring}. Our test belongs to this line, with two
differences in emphasis. The supplied context is a record that is deliberately incomplete, and findings are scored by
the reason the judge gives, because in this task a flag for the wrong reason is easy to obtain.

\paragraph{Following a signal over the evidence.}
That models follow supplied signals is well established. Judges shown prior scores as metadata move toward them, and
an instruction to disregard the metadata does not remove the effect \citep{kapetanovic2026anchoring}. Code auditors
given false assurances from static analysis lose detection, and prompt instructions do not reliably restore it
\citep{liu2026trusting}. Agents often adopt corrupted tool outputs (over a third of the time for every
tool tested) and rarely warn users of the conflict \citep{yang2026overreliance}. In the legal domain, models over-trust authoritative but false legal sources and struggle to judge
when a retrieved citation is useful or misleading, on a benchmark that encodes temporal validity, and larger models appear more susceptible \citep{molfetta2026sycophants}. The effect is not
universal: judges given a wrong reference answer often ignore it \citep{lee2026reference}, so how far a model follows
a signal depends on the task and on the form of the signal.

Our results agree with this literature in direction. The setting differs in how the misleading signal arises. In the
studies above it is constructed to be false: an injected score, a fabricated assurance, a corrupted tool output. Our
record errs only through missing links, a common way for real amendment records to fail, though not the only one
(our case study also found wrong links). In its authoritative form the gap
then reads ``in force'', which is false, but in one condition the line says only that no replacement is recorded,
which is true, and the judges' catches fell there too. The
judge also has both documents in front of it, and a deterministic lookup in the same record gives a reference for what
the record alone achieves. Our presentation conditions vary the record's wording, its apparent authority and a
warning about its completeness while keeping the record itself fixed; \S\ref{sec:discussion} compares the warning's
effect with the anchoring result of \citet{kapetanovic2026anchoring}. Deference to an authoritative label is related
to sycophancy towards a user's stated view \citep{sharma2023sycophancy}, but the source of the pressure is different.

\paragraph{Outdated evidence and temporal knowledge.}
A separate line asks whether models that write answers can cope with change. Stale-Document Poisoning
\citep{staledoc2026} shows that outdated retrieved documents overturn answers a model had right, and that an explicit
statement of when the old evidence stops applying helps larger models. EvolveBench \citep{zhu2025evolvebench} and
other benchmarks of time-sensitive questions \citep{chen2021timeqa,zhang2021situatedqa,vu2024freshllms,ouyang2025hoh}
measure whether models track facts that change, and work on knowledge conflicts asks when supplied context overrides
what a model knows \citep{xie2024chameleon,xu2024knowledge,wu2024clasheval}. These studies concern the model that
writes the answer. We study a model that reviews an answer already written, which makes false alarms on correct answers
part of the measurement.

\paragraph{Versioned law and amendment records.}
Legal informatics has long modelled the change of norms explicitly: systems that manage consolidated acts across their
successive versions \citep{palmirani2002norma}, models of the temporal dimensions of norm change
\citep{palmirani2006time}, defeasible-logic accounts of abrogation and
annulment \citep{governatori2010changing}, and temporal metadata (in-force, efficacy and
application times) in LegalRuleML \citep{palmirani2011legalruleml}.
More recent work finds that retrieval over a corpus holding only current texts returns the wrong version of a
provision \citep{cymbler2026temporal},
diagnoses staleness in statutory question answering \citep{prior2026oldfriend}, and shows that LLM consolidation of statutes from amendment
instructions is often not exact (50\% exact matches for single amendments, 21\% for chains)
\citep{prior2025consolidation}. Version-aware retrieval has been built for amended regulations
\citep{nam2026timelyrag} and, outside law, for evolving technical documentation \citep{huwiler2025versionrag}. Our record is a much thinner object than these models, a list of
document-to-document replacements with dates, and our notion of currency (``replaced on or before the date asked
about'') is the simplest of the temporal notions they distinguish. We take such a record as given, imperfections
included, and ask what a judge does with it.

\paragraph{Contrast items and right-for-the-wrong-reason.}
Models can score well by exploiting features that correlate with the label \citep{mccoy2019right}, and minimally
different contrast items are a standard way to expose this \citep{gardner2020contrast}. Our date controls and decoys
play that role for document currency. The date control changes only the date of a fault. The decoy keeps a newer
document on the same topic, often with the same title, but one that leaves the older document in force. As
\S\ref{sec:res-shortcut} shows, our decoys control for title similarity but not for overlap between abstracts, so they
are a partial contrast. Finally, label errors change model rankings \citep{northcutt2021pervasive,gema2025mmlu}, and
a clinical benchmark built with LLM help was found to have at least 27\% of its test labels likely wrong or
incomputable \citep{ye2025stewardship}. This is the main reason our
primary data use the publisher's own record as gold, and the reason the regulatory study, whose AI-assisted key failed
our checks, is reported only descriptively.

\section{The test}
\label{sec:framework}

The test needs two inputs: a record of replacements that is correct wherever it is used as gold, and a short text for
each document. It needs no labelled answers, because the record already says which citations are wrong on which dates.

\begin{figure}[t]
\centering
\resizebox{\linewidth}{!}{%
\begin{tikzpicture}[
  font=\footnotesize, every node/.style={inner sep=4pt},
  box/.style={draw=black!45, rounded corners=2pt, align=left, anchor=north west},
  rec/.style={box, fill=black!4},
  item/.style={box, text width=3.25cm, minimum height=1.75cm},
  arr/.style={-{Stealth[length=4pt]}, black!55}]
\node[rec, text width=10.85cm, align=justify] (r1) at (0,0) {\textbf{Replacement} (from the record): RFC 6312 (2011-07) is replaced by
  RFC 6342 (2011-08).};
\node[rec, text width=3.25cm] (r2) at (11.4,0) {\textbf{Update} (decoy pair): RFC 7124 updates RFC 5066, which stays in force.};
\foreach \x/\n/\t in {0/f/{\textbf{Fault}\\ As of 2026-10-01, cites \textbf{RFC 6312}.\\ \textit{Wrong: replaced.}\\ Expected: flag as replaced},
                      3.8/c/{\textbf{Control}\\ As of 2026-10-01, cites \textbf{RFC 6342}.\\ \textit{Correct: the successor.}\\ Expected: pass},
                      7.6/d/{\textbf{Date control}\\ As of 2011-07-28, cites \textbf{RFC 6312}.\\ \textit{Correct: not yet replaced.}\\ Expected: pass},
                      11.4/y/{\textbf{Decoy}\\ As of 2026-10-01, cites \textbf{RFC 5066}.\\ \textit{Correct: only updated.}\\ Expected: pass}}
  \node[item] (\n) at (\x,-1.45) {\t};
\foreach \n in {f,c,d} \draw[arr] (r1.south -| \n.north) -- (\n.north);
\draw[arr] (r2.south) -- (y.north);
\node[box, text width=14.65cm, align=justify] (k) at (0,-3.65)
  {\textbf{What the judge sees.} The question, the answer, and both documents' titles, dates and abstracts, with any
   sentence that states the replacement or names the other document removed. The context is one of six primary contexts: no record (\texttt{both-texts}); a line
   from the correct record (\texttt{status}); or a line from a damaged record with half its links dropped, where a
   dropped link reads as in force. This link is dropped in the damaged record, so its fault is a \emph{gap fault}. The
   damaged line is shown four ways with the same content: ``in force'', ``no replacement recorded'', as an unverified
   note, or with a warning that the record is incomplete.};
\node[box, text width=14.65cm, align=justify, anchor=north west] (s) at ($(k.south west)+(0,-0.2)$)
  {\textbf{Scoring.} A fault counts as caught only if a finding names the cited RFC with the reason \emph{replaced}. On
   correct answers, a finding that the RFC was replaced or not yet published is a false alarm. A lookup in the record,
   with no language model, is the baseline: it catches exactly the faults whose link the record holds.};
\end{tikzpicture}}
\caption{The test on one link from the RFC index and its matched decoy. Each replacement yields a fault and two
correct answers; a matched update pair yields a decoy. The judge's context is varied while the question and answer
stay fixed (Table~\ref{tab:menu}); the full prompt for this fault is in Figure~\ref{fig:prompt}.}
\label{fig:design}
\end{figure}

\paragraph{Matched items.}
Every item poses the same kind of question, ``As of date $D$, which document specifies $T$?'', where $T$ is the title
of the older document, and supplies an answer that cites one document. For each link in the record (an old document and
the document that replaced it) we write three items, and we pair each link with one decoy (Figure~\ref{fig:design}):
\begin{itemize}
  \item a \emph{fault} cites the old document, with $D$ after the replacement;
  \item a \emph{control} cites the replacing document, with the same $D$;
  \item a \emph{date control} cites the old document, with $D$ before the replacement existed, so the answer is correct;
  \item a \emph{decoy} cites the older document of a different pair, in which a newer document \emph{updates} the
  older one without replacing it. The older document is still in force, so this answer is also correct.
\end{itemize}
The fault and its date control are the same answer with a different date. A judge that objects to old citations
without comparing dates will catch the fault and fail the date control. The decoy targets a different shortcut. In the
fault, a newer document on the same subject sits next to the cited one; in the decoy, so does a newer document, often
with the same title, but it did not replace the older one. A judge that flags any older document with a newer
neighbour will catch the fault and fail the decoy. Catch rates alone cannot tell either judge apart from one that
actually checks.

\paragraph{What the judge sees.}
The context is the part a builder controls, and we vary it while holding the question and the answer fixed
(Table~\ref{tab:menu}). In every primary context the judge gets both documents' titles, dates and abstracts, with any
sentence that states a replacement or names the other document removed, so that the replacement has to be inferred.
Titles are not filtered, and in two links the replacing RFC's title names the old one (\S\ref{sec:limitations}).
Five of the six primary contexts add one line from an amendment record about the cited document. In the damaged record, half
the links are dropped with a fixed seed; for a dropped link the record says nothing about a replacement, and its line
therefore reports the replaced document as in force. We call the faults on these links \emph{gap faults}. The four
primary damaged contexts are built from the same damaged record and differ in how its line is presented:
\begin{itemize}
  \item the authoritative form, an ``amendment record lookup'' reporting the document as \texttt{in\_force};
  \item the same record with the line worded \texttt{no\_replacement\_recorded} (wording);
  \item the same line headed as an ``unverified note from an automated tagger'' (authority);
  \item the authoritative form with a warning that the record is incomplete and may omit replacements (warning).
\end{itemize}
The pilot also used contexts that we did not repeat here: the citation alone (\texttt{cite-only}), the cited document
alone (\texttt{cited-text}), masked document numbers (\texttt{both-masked}), the replacing document's own notice of
replacement (\texttt{both-explicit}), an instruction to check replacement first (\texttt{check-first}), and two worked
examples (\texttt{few-shot}). They are reported in Appendix~\ref{app:pilot}.

\begin{table}[t]
\centering\footnotesize
\begin{tabular}{>{\raggedright\arraybackslash}p{2.6cm}>{\raggedright\arraybackslash}p{4.9cm}}
\toprule
Context & The judge is given, besides both texts \\
\midrule
\texttt{both-texts} & nothing else \\
\texttt{status} & a line from the correct record: the cited document's status on the date \\
\texttt{status-damaged} & the line from a record with half its links dropped, where a dropped link reads ``in force'' (authoritative form) \\
\texttt{norecord-damaged} & as above, the line reading ``no replacement recorded'' (wording) \\
\texttt{lowauth-damaged} & as \texttt{status-damaged}, headed as an unverified note from an automated tagger (authority) \\
\texttt{caveat-damaged} & as \texttt{status-damaged}, with the warning that the record is incomplete and may omit replacements (warning) \\
\midrule
\texttt{conflict-damaged} & as \texttt{status-damaged}, with one more reason a finding may give: the record conflicts with the evidence (exploratory, added later) \\
\bottomrule
\end{tabular}
\caption{The six primary contexts, and one exploratory context added after the main analysis (\S\ref{sec:studies}).
The damaged contexts use the same damaged record; the warning adds one true statement about it.}
\label{tab:menu}
\end{table}

\paragraph{Scoring by reason.}
The judge returns a verdict and a list of findings. Each finding names a document and gives one reason from a closed
list: replaced (coded \texttt{superseded} in the prompt), not yet published, not applicable, content error, unsupported, or other. We count a fault as
\emph{caught} only if some finding names the cited document with the reason that it was replaced. A judge that flags
the fault as, say, a content error has noticed that something is off without identifying the error, and an any-finding
count would credit it with the catch. On the three kinds of correct answer we record two false-alarm counts: a
\emph{validity} false alarm, a finding that a document was replaced or not yet published, and any finding at all. The
first measures mistakes about currency; the second also counts complaints about relevance or content, which the
question format may provoke (\S\ref{sec:res-scoring}). A reply that cannot be parsed counts against the judge: as a miss
on a fault and as a false alarm of both kinds on a correct answer. We call a reply on a fault a \emph{bare pass} when it
contains no finding at all.

\paragraph{Lookup baseline.}
A deterministic lookup, with no language model, flags exactly the documents the record marks as replaced. With the
correct record it catches every fault and raises no false alarm. With the damaged record it catches the faults whose
link survived and none of the gap faults. It tells us what the record alone achieves, and so what a judge given the
record adds.

\paragraph{Scope of the design.}
The test covers one error, citing a whole replaced document, and one decision, the judge's verdict on a fixed answer.
Partial amendments, chains of replacements and cross-references would need other item generators. Following the record
can help as well as hurt. The gap faults measure the cost of following a silent record, and the decoys, whose record
line is correct, give one measure of the benefit.

\section{Data and setup}
\label{sec:studies}
\label{sec:study2}

\paragraph{Replacements.}
IETF Requests for Comments (RFCs) replace one another, and the RFC Editor's index \citep{rfcindex2026} records each
replacement (``Obsoletes'' / ``Obsoleted by'') together with publication dates and abstracts. We kept the simplest
cases: replacements in which the old RFC was replaced by exactly one RFC, the replacing RFC is itself current, was
published in 2005 or later and replaces at most two RFCs, and both documents have abstracts. From these we drew
\ENLinks{} links with a fixed seed. Because the record is the publisher's, the gold needs no annotation. Each document
is represented by its title, publication date and abstract. In 33 of the \ENLinks{} links the two titles are
near-identical (a normalised string-similarity ratio of at least 0.8), which matters for the shortcut analysis below.

\paragraph{Decoys.}
The index also records updates (``Updates'' / ``Updated by''), in which a newer RFC amends part of an older one without
replacing it. A decoy is an update pair in which the older RFC was never replaced and the newer one is current, chosen
under the same filters as the links and sharing no document with them. For each link we took the nearest unused update
pair within the link's same-title group, by a distance that combines IETF area (our proxy for topic), title similarity,
both publication dates and both abstract lengths. The matching is greedy in link order and deterministic. Balance on
the matched features is good for dates (standardised mean differences \EBalOldYearSMD{} for the old date,
\EBalNewYearSMD{} for the new date and \EBalGapYearsSMD{} for the gap) and acceptable for title similarity
(\EBalTitleRatioSMD); \EBalArea{} decoys share their link's IETF area. Two imbalances remain. First, updating RFCs have shorter
abstracts than replacing ones (\EBalNewWordsSMD), a property of updates that we chose not to force away. Second,
abstract overlap was not a matching feature. A replacing RFC usually repeats much of the abstract of the
RFC it replaces (median similarity \EAbsSimLinksSame{} for same-title links, by the same string-similarity ratio
computed on the abstracts as the judges saw them), whereas an updating RFC does not (median
\EAbsSimDecoysSame{} for same-title decoys). We found this only after the runs; its consequences are discussed in
\S\ref{sec:res-shortcut}.

\paragraph{Judges and runs.}
Four judges were run on all six primary contexts with the same links and decoys: Gemini~3.1~Pro, GPT-5.5 and Claude
Opus~4.5 through the Kaggle Benchmarks proxy, and Gemma~4~26B through Google's API at temperature 0. The proxy sets
temperature and seed itself and does not report them, which limits exact reproduction of the hosted judges. Every
request to the hosted judges went through the proxy with an output cap of 4,000 tokens, sending the system and user
text as two messages. No reply is taken from the pilot. A first analysis of these runs had reused the pilot's replies
to identical prompts, most of them obtained with an earlier form of the proxy call; we then requested every such prompt
again under the current call. This changed the score of \EGPTReaskChanged{} of GPT-5.5's \EGPTReaskItems{} items,
\EGeminiReaskChanged{} of Gemini's, \EOpusReaskChanged{} of Claude's and \EGemmaReaskChanged{} of Gemma's, and no
conclusion; all results reported here use the new replies. Within these runs, a prompt that recurs (for example, a
text-only prompt shared by the two draws below) is asked once and its reply reused. For Gemma, the client retries a request once without the API's JSON mode when the
JSON-mode reply is cut off; the run log records that this happened, but not for which items. Every block of replies
records the model version and a hash of the harness code. For this paper we rescored every reply from its raw text, and
every stored score matched.

\paragraph{Hypotheses and analysis.}
The hypotheses below, with their tests and the reading we would give each outcome, were written after the pilot and
committed to the project's private development repository shortly before the runs. This is a timestamped internal record, not a public
registration. Before any result of these runs existed, the plan was amended to add two presentation conditions (wording and
authority, beside the warning) and to match decoys on more features; the amendment is part of the same history.
\begin{description}
  \item[H1] Given both texts, a judge's right-reason catch on faults exceeds its validity false alarms on the matched
  decoys. Tested per judge as the mean paired difference over the \ENLinks{} link--decoy pairs, with a 2,000-resample
  bootstrap interval over pairs.
  \item[H2a] On the 24 gap faults, a judge catches fewer with the damaged record (authoritative form) than with no
  record. Two-sided exact McNemar test on the paired items; Holm adjustment across judges.
  \item[H2b] Each other presentation of the damaged record changes the catch on the gap faults relative to the
  authoritative form. Two-sided exact McNemar tests; Holm adjustment across all \EHTwobTests{} presentation--judge
  pairs.
\end{description}
The unit of analysis is the link, never the model call. With 24 pairs, a change in which every discordant item
moves the same way needs at least six items to reach the 5\% level before adjustment (two-sided exact $p$ = .031), and
more after the Holm adjustment; changes that do not reach it are reported as inconclusive. The plan also listed Wilson
intervals for rates and an exploratory mixed-effects model pooled over judges; we report exact counts and paired tests
instead, and did not fit the model, since no claim depends on it.

\paragraph{A second draw of removed links.} Because the pilot had shown that the choice of removed links matters, we
amended the plan, once results on the first draw (damage seed~0) existed for three judges and before any seed-1 call, to repeat H2a on a second
random choice (damage seed~1) for the same four judges, with the tests fixed in advance: H2a per judge on seed~1 with Holm adjustment across judges, and a per-judge estimate over both seeds. The
two draws drop 24 links each and share 12, so each adds 12 gap faults of its own. A shared link is the same item under
both seeds, with an identical prompt whose cached reply was reused, so the pooled estimate counts it once: it is the
mean drop in catch over the 36 links in either gap set, with a 2,000-resample bootstrap interval over links and an exact
sign test on the links that changed. Text-only prompts do not depend on the record, so seed~1 reuses the text-only
replies; its damaged-record replies on the 12 new gap faults are new requests. The second draw is a robustness check on the choice of removed links, not an independent replication: it shares half
its gap faults and all text-only replies with the first. The hypotheses and thresholds were not changed. Everything else in
\S\ref{sec:results}, including every comparison by title or abstract overlap, is exploratory.

\paragraph{A code for conflicts.} After the results above existed, a reader observed that the reason list gives a judge
no way to report that the record and the texts disagree, so a bare pass might reflect the reply format rather than the
judge. We therefore added one exploratory context, \texttt{conflict-damaged}: the authoritative damaged record, with one
more reason in the list, \texttt{record\_conflict} (``the record and the evidence disagree, or whether the cited document
is still current cannot be confirmed''). A \texttt{record\_conflict} finding is counted separately and is not a catch.
This context was run on the first draw for all four judges; no test was planned for it, and we report counts only.

\paragraph{The pilot.}
Before the final design, six judges ran up to ten contexts on the first 24 links, and Qwen~2.5~32B on all 48
(Appendix~\ref{app:pilot}). The pilot had no decoys, and of the presentation conditions only the ``no replacement
recorded'' wording; its damaged record, built over 24 links rather than 48 (except for Qwen), dropped a different set
of links. We use it for the date-control result and for the contexts not repeated here.

\paragraph{The case study.}
The test was first developed on Indian rooftop-solar regulation, with an amendment record extracted by an AI pipeline
and seven judges. Because its answer key failed the quality checks set for it, it is reported descriptively in
Appendix~\ref{app:case} and the supplementary material, and no result in \S\ref{sec:results} uses it.

\paragraph{Use of AI assistants.}\label{sec:ai-use}
An AI coding assistant, built on models from the same developer as one of the judges (Claude Opus~4.5), drafted the
hypotheses and analysis plan, which the authors reviewed and amended; wrote the harness; sampled the links and
decoys; ran the Gemma judge and prepared the hosted-judge runs, which were executed through the benchmark proxy under
the lead author's account; analysed the replies; and drafted and edited the manuscript. In the case study it drafted
the labels and extracted the amendment record, which two authors checked blind with a third adjudicating. The RFC
gold is the publisher's, and the assistant's own model was not run as a judge. The authors checked judge replies,
statistical results and references against their sources, read and revised the manuscript, and take full
responsibility for its content.

\section{Results}
\label{sec:results}

Table~\ref{tab:e123-main} lists every judge's catches and false alarms in every context. We begin with the correct
record, which sets the baseline, and then turn to the gap faults.

\IfFileExists{generated/tab_e123_main.tex}{\begin{table*}[t]\centering\footnotesize
\caption{All 48 links, per judge and context. Catch: right-reason catches of faults (any: any finding on the fault). False alarms (FA) count validity findings (replaced or not yet published) unless marked any. Controls cite the replacing RFC; date controls cite the replaced RFC as of a date before its replacement; decoys cite an RFC that a newer one updates but does not replace. \texttt{conflict-damaged} is exploratory: a \texttt{record\_conflict} finding counts under any, not as a catch or a validity false alarm. All counts are of 48.}
\label{tab:e123-main}
\resizebox{\textwidth}{!}{%
\begin{tabular}{llccccccc}
\toprule
 & & \multicolumn{2}{c}{Faults} & \multicolumn{2}{c}{Controls} & Date controls & \multicolumn{2}{c}{Decoys} \\
\cmidrule(lr){3-4}\cmidrule(lr){5-6}\cmidrule(lr){7-7}\cmidrule(lr){8-9}
Judge & Context & Catch & any & FA & any & FA & FA & any \\
\midrule
Gemma 4 26B & \texttt{both-texts} & \EGemmaCatchTexts & \EGemmaCatchAnyTexts & \EGemmaCtrlFATexts & \EGemmaCtrlAnyTexts & \EGemmaDateFATexts & \EGemmaDecoyFATexts & \EGemmaDecoyAnyTexts \\
 & \texttt{status} & \EGemmaCatchRecord & \EGemmaCatchAnyRecord & \EGemmaCtrlFARecord & \EGemmaCtrlAnyRecord & \EGemmaDateFARecord & \EGemmaDecoyFARecord & \EGemmaDecoyAnyRecord \\
 & \texttt{status-damaged} & \EGemmaCatchDamaged & \EGemmaCatchAnyDamaged & \EGemmaCtrlFADamaged & \EGemmaCtrlAnyDamaged & \EGemmaDateFADamaged & \EGemmaDecoyFADamaged & \EGemmaDecoyAnyDamaged \\
 & \texttt{norecord-damaged} & \EGemmaCatchNorecord & \EGemmaCatchAnyNorecord & \EGemmaCtrlFANorecord & \EGemmaCtrlAnyNorecord & \EGemmaDateFANorecord & \EGemmaDecoyFANorecord & \EGemmaDecoyAnyNorecord \\
 & \texttt{lowauth-damaged} & \EGemmaCatchLowauth & \EGemmaCatchAnyLowauth & \EGemmaCtrlFALowauth & \EGemmaCtrlAnyLowauth & \EGemmaDateFALowauth & \EGemmaDecoyFALowauth & \EGemmaDecoyAnyLowauth \\
 & \texttt{caveat-damaged} & \EGemmaCatchCaveat & \EGemmaCatchAnyCaveat & \EGemmaCtrlFACaveat & \EGemmaCtrlAnyCaveat & \EGemmaDateFACaveat & \EGemmaDecoyFACaveat & \EGemmaDecoyAnyCaveat \\
 & \texttt{conflict-damaged} & \EGemmaCatchConflict & \EGemmaCatchAnyConflict & \EGemmaCtrlFAConflict & \EGemmaCtrlAnyConflict & \EGemmaDateFAConflict & \EGemmaDecoyFAConflict & \EGemmaDecoyAnyConflict \\
\addlinespace
GPT-5.5 & \texttt{both-texts} & \EGPTCatchTexts & \EGPTCatchAnyTexts & \EGPTCtrlFATexts & \EGPTCtrlAnyTexts & \EGPTDateFATexts & \EGPTDecoyFATexts & \EGPTDecoyAnyTexts \\
 & \texttt{status} & \EGPTCatchRecord & \EGPTCatchAnyRecord & \EGPTCtrlFARecord & \EGPTCtrlAnyRecord & \EGPTDateFARecord & \EGPTDecoyFARecord & \EGPTDecoyAnyRecord \\
 & \texttt{status-damaged} & \EGPTCatchDamaged & \EGPTCatchAnyDamaged & \EGPTCtrlFADamaged & \EGPTCtrlAnyDamaged & \EGPTDateFADamaged & \EGPTDecoyFADamaged & \EGPTDecoyAnyDamaged \\
 & \texttt{norecord-damaged} & \EGPTCatchNorecord & \EGPTCatchAnyNorecord & \EGPTCtrlFANorecord & \EGPTCtrlAnyNorecord & \EGPTDateFANorecord & \EGPTDecoyFANorecord & \EGPTDecoyAnyNorecord \\
 & \texttt{lowauth-damaged} & \EGPTCatchLowauth & \EGPTCatchAnyLowauth & \EGPTCtrlFALowauth & \EGPTCtrlAnyLowauth & \EGPTDateFALowauth & \EGPTDecoyFALowauth & \EGPTDecoyAnyLowauth \\
 & \texttt{caveat-damaged} & \EGPTCatchCaveat & \EGPTCatchAnyCaveat & \EGPTCtrlFACaveat & \EGPTCtrlAnyCaveat & \EGPTDateFACaveat & \EGPTDecoyFACaveat & \EGPTDecoyAnyCaveat \\
 & \texttt{conflict-damaged} & \EGPTCatchConflict & \EGPTCatchAnyConflict & \EGPTCtrlFAConflict & \EGPTCtrlAnyConflict & \EGPTDateFAConflict & \EGPTDecoyFAConflict & \EGPTDecoyAnyConflict \\
\addlinespace
Gemini 3.1 Pro & \texttt{both-texts} & \EGeminiCatchTexts & \EGeminiCatchAnyTexts & \EGeminiCtrlFATexts & \EGeminiCtrlAnyTexts & \EGeminiDateFATexts & \EGeminiDecoyFATexts & \EGeminiDecoyAnyTexts \\
 & \texttt{status} & \EGeminiCatchRecord & \EGeminiCatchAnyRecord & \EGeminiCtrlFARecord & \EGeminiCtrlAnyRecord & \EGeminiDateFARecord & \EGeminiDecoyFARecord & \EGeminiDecoyAnyRecord \\
 & \texttt{status-damaged} & \EGeminiCatchDamaged & \EGeminiCatchAnyDamaged & \EGeminiCtrlFADamaged & \EGeminiCtrlAnyDamaged & \EGeminiDateFADamaged & \EGeminiDecoyFADamaged & \EGeminiDecoyAnyDamaged \\
 & \texttt{norecord-damaged} & \EGeminiCatchNorecord & \EGeminiCatchAnyNorecord & \EGeminiCtrlFANorecord & \EGeminiCtrlAnyNorecord & \EGeminiDateFANorecord & \EGeminiDecoyFANorecord & \EGeminiDecoyAnyNorecord \\
 & \texttt{lowauth-damaged} & \EGeminiCatchLowauth & \EGeminiCatchAnyLowauth & \EGeminiCtrlFALowauth & \EGeminiCtrlAnyLowauth & \EGeminiDateFALowauth & \EGeminiDecoyFALowauth & \EGeminiDecoyAnyLowauth \\
 & \texttt{caveat-damaged} & \EGeminiCatchCaveat & \EGeminiCatchAnyCaveat & \EGeminiCtrlFACaveat & \EGeminiCtrlAnyCaveat & \EGeminiDateFACaveat & \EGeminiDecoyFACaveat & \EGeminiDecoyAnyCaveat \\
 & \texttt{conflict-damaged} & \EGeminiCatchConflict & \EGeminiCatchAnyConflict & \EGeminiCtrlFAConflict & \EGeminiCtrlAnyConflict & \EGeminiDateFAConflict & \EGeminiDecoyFAConflict & \EGeminiDecoyAnyConflict \\
\addlinespace
Claude Opus 4.5 & \texttt{both-texts} & \EOpusCatchTexts & \EOpusCatchAnyTexts & \EOpusCtrlFATexts & \EOpusCtrlAnyTexts & \EOpusDateFATexts & \EOpusDecoyFATexts & \EOpusDecoyAnyTexts \\
 & \texttt{status} & \EOpusCatchRecord & \EOpusCatchAnyRecord & \EOpusCtrlFARecord & \EOpusCtrlAnyRecord & \EOpusDateFARecord & \EOpusDecoyFARecord & \EOpusDecoyAnyRecord \\
 & \texttt{status-damaged} & \EOpusCatchDamaged & \EOpusCatchAnyDamaged & \EOpusCtrlFADamaged & \EOpusCtrlAnyDamaged & \EOpusDateFADamaged & \EOpusDecoyFADamaged & \EOpusDecoyAnyDamaged \\
 & \texttt{norecord-damaged} & \EOpusCatchNorecord & \EOpusCatchAnyNorecord & \EOpusCtrlFANorecord & \EOpusCtrlAnyNorecord & \EOpusDateFANorecord & \EOpusDecoyFANorecord & \EOpusDecoyAnyNorecord \\
 & \texttt{lowauth-damaged} & \EOpusCatchLowauth & \EOpusCatchAnyLowauth & \EOpusCtrlFALowauth & \EOpusCtrlAnyLowauth & \EOpusDateFALowauth & \EOpusDecoyFALowauth & \EOpusDecoyAnyLowauth \\
 & \texttt{caveat-damaged} & \EOpusCatchCaveat & \EOpusCatchAnyCaveat & \EOpusCtrlFACaveat & \EOpusCtrlAnyCaveat & \EOpusDateFACaveat & \EOpusDecoyFACaveat & \EOpusDecoyAnyCaveat \\
 & \texttt{conflict-damaged} & \EOpusCatchConflict & \EOpusCatchAnyConflict & \EOpusCtrlFAConflict & \EOpusCtrlAnyConflict & \EOpusDateFAConflict & \EOpusDecoyFAConflict & \EOpusDecoyAnyConflict \\
\bottomrule
\end{tabular}}
\end{table*}
}{}

\subsection{With a correct record}
\label{sec:res-correct}

Given a line from the correct record, the judges caught all or nearly all faults (Gemma \EGemmaCatchRecord, GPT-5.5
\EGPTCatchRecord, Gemini \EGeminiCatchRecord, Claude \EOpusCatchRecord), as a lookup in the same record does by
construction. The record also helped on correct answers. For decoys, the record line correctly reports the older RFC as
in force, and decoy false alarms fell from \EGPTDecoyFATexts{} to \EGPTDecoyFARecord{} for GPT-5.5, from
\EGeminiDecoyFATexts{} to \EGeminiDecoyFARecord{} for Gemini, from \EGemmaDecoyFATexts{} to \EGemmaDecoyFARecord{} for Gemma and from \EOpusDecoyFATexts{} to \EOpusDecoyFARecord{} for
Claude. Following the record therefore has a benefit as well as the cost measured below, and on this corpus a judge
given a correct record does essentially what the record does in its findings about currency.

With the damaged record the lookup catches the 24 faults whose link survived and none of the 24 gap faults. Over all
\ENLinks{} faults the judges scored Gemma \EGemmaCatchDamaged, GPT-5.5 \EGPTCatchDamaged, Gemini \EGeminiCatchDamaged{}
and Claude \EOpusCatchDamaged. The gap faults themselves are analysed next.

\subsection{Gap faults}
\label{sec:res-gaps}

\begin{figure}[t]
\centering
\includegraphics[width=\linewidth]{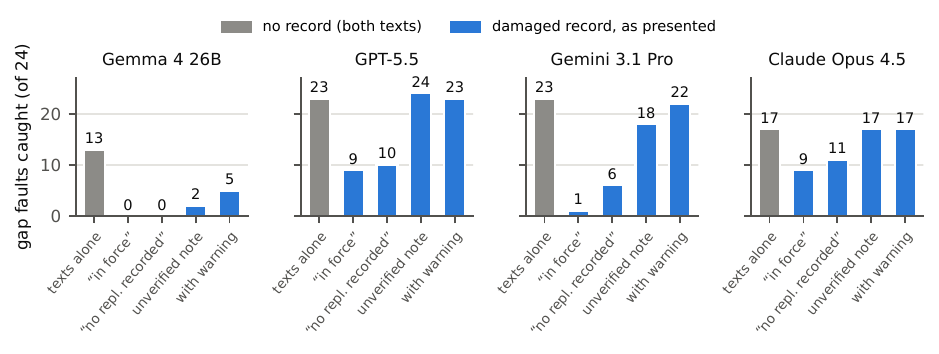}
\caption{Catches on the 24 gap faults per judge (the values of Table~\ref{tab:e123-gap}): from both texts with no record, and with the damaged record in each of its four presentations. A lookup in the damaged record catches none of these faults.}
\label{fig:gap}
\end{figure}

Table~\ref{tab:e123-gap} and Figure~\ref{fig:gap} isolate the 24 gap faults; the items are the same in every column, so each comparison is
within a judge.

\IfFileExists{generated/tab_e123_gap.tex}{\begin{table*}[t]\centering\footnotesize
\caption{The 24 gap faults per judge: faults whose link the damaged record dropped (half the 48 links, seed 0). Same items in every column. The four damaged-record presentations use the same damaged record. Bare: passes with no finding under the authoritative ``in force'' line. $p$: exact McNemar, Holm-adjusted; H2a compares the authoritative line with no record, H2b each other presentation with the authoritative line.}
\label{tab:e123-gap}
\resizebox{\textwidth}{!}{%
\begin{tabular}{lcccccc cccc}
\toprule
 & No record & \multicolumn{5}{c}{Damaged record, presented as} & \multicolumn{4}{c}{Holm $p$} \\
\cmidrule(lr){3-7}\cmidrule(lr){8-11}
Judge & (texts) & record, ``in force'' & (bare) & ``no replacement recorded'' & unverified note & record + warning & H2a & wording & authority & warning \\
\midrule
Gemma 4 26B & \EGemmaGapTexts & \EGemmaGapDamaged & \EGemmaBareDamaged & \EGemmaGapNorecord & \EGemmaGapLowauth & \EGemmaGapCaveat & \EGemmaHTwoaHolm & \EGemmaArmNorecordHolm & \EGemmaArmLowauthHolm & \EGemmaArmCaveatHolm \\
GPT-5.5 & \EGPTGapTexts & \EGPTGapDamaged & \EGPTBareDamaged & \EGPTGapNorecord & \EGPTGapLowauth & \EGPTGapCaveat & \EGPTHTwoaHolm & \EGPTArmNorecordHolm & \EGPTArmLowauthHolm & \EGPTArmCaveatHolm \\
Gemini 3.1 Pro & \EGeminiGapTexts & \EGeminiGapDamaged & \EGeminiBareDamaged & \EGeminiGapNorecord & \EGeminiGapLowauth & \EGeminiGapCaveat & \EGeminiHTwoaHolm & \EGeminiArmNorecordHolm & \EGeminiArmLowauthHolm & \EGeminiArmCaveatHolm \\
Claude Opus 4.5 & \EOpusGapTexts & \EOpusGapDamaged & \EOpusBareDamaged & \EOpusGapNorecord & \EOpusGapLowauth & \EOpusGapCaveat & \EOpusHTwoaHolm & \EOpusArmNorecordHolm & \EOpusArmLowauthHolm & \EOpusArmCaveatHolm \\
\bottomrule
\end{tabular}}
\end{table*}
}{}

From the texts alone, Gemini caught \EGeminiGapTexts{} of the gap faults and, with the damaged record in its
authoritative form, \EGeminiGapDamaged: \EGeminiHTwoaLost{} catches lost and \EGeminiHTwoaGained{} gained (H2a, Holm
$p$ \EGeminiHTwoaHolm). GPT-5.5 went from \EGPTGapTexts{} to \EGPTGapDamaged{} (\EGPTHTwoaLost{} lost,
\EGPTHTwoaGained{} gained; Holm $p$ \EGPTHTwoaHolm) and Claude from \EOpusGapTexts{} to \EOpusGapDamaged{}
(\EOpusHTwoaLost{} lost, \EOpusHTwoaGained{} gained; Holm $p$ \EOpusHTwoaHolm). Gemma caught fewer from the texts,
\EGemmaGapTexts, and none with the record (Holm $p$ \EGemmaHTwoaHolm). H2a therefore holds for all four judges. Under
the damaged record, \EGeminiBareDamaged{} of Gemini's \EGeminiMissDamaged{} misses were replies with no finding at
all; GPT-5.5's were \EGPTBareDamaged{} of \EGPTMissDamaged, Claude's \EOpusBareDamaged{} of \EOpusMissDamaged{} and
Gemma's \EGemmaBareDamaged{} of \EGemmaMissDamaged. Claude's two misses that did carry a finding both mention the newer RFC, but as a content error (a
title wrongly attributed to the cited RFC, or an answer that may be incomplete), not as a replacement; one repeats, as the
record says, that the older RFC remains in force. Under none of the four presentations does a missed gap fault name a
conflict between the record and the texts; the few replies that mention the record are catches.

\IfFileExists{generated/tab_e123_seeds.tex}{\begin{table*}[t]\centering\footnotesize
\caption{H2a on two random draws of dropped links. Each seed drops 24 of the 48 links; the two draws share some links (Shared). Per seed: gap faults caught with no record $\to$ with the damaged record (authoritative form), lost/gained, Holm-adjusted exact McNemar $p$ (across judges, within seed). Pooled: unique links in either gap set, a shared link counted once (its reply is the same under both seeds); mean drop in points with a 95\% bootstrap interval over links, lost/gained, exact sign test. Seed 0 is the planned test; seed 1 and the pooled estimate were added before any seed-1 run, when seed-0 results existed for three judges.}
\label{tab:e123-seeds}
\resizebox{\textwidth}{!}{%
\begin{tabular}{l ccc cccc cccc}
\toprule
 & \multicolumn{3}{c}{Seed 0} & \multicolumn{4}{c}{Seed 1} & \multicolumn{4}{c}{Pooled over unique links} \\
\cmidrule(lr){2-4}\cmidrule(lr){5-8}\cmidrule(lr){9-12}
Judge & Caught & Lost/gained & $p$ & Caught & Lost/gained & $p$ & Shared & Links & Drop & Lost/gained & $p$ \\
\midrule
Gemma 4 26B & \EGemmaGapTexts{} $\to$ \EGemmaGapDamaged & \EGemmaHTwoaLost/\EGemmaHTwoaGained & \EGemmaHTwoaHolm & \EGemmaSOneOwn{} $\to$ \EGemmaSOneWith & \EGemmaSOneLost/\EGemmaSOneGained & \EGemmaSOneHolm & \EGemmaSOneShared & \EGemmaPoolLinks & \EGemmaPoolMean{} {\scriptsize \EGemmaPoolCI} & \EGemmaPoolLost/\EGemmaPoolGained & \EGemmaPoolP \\
GPT-5.5 & \EGPTGapTexts{} $\to$ \EGPTGapDamaged & \EGPTHTwoaLost/\EGPTHTwoaGained & \EGPTHTwoaHolm & \EGPTSOneOwn{} $\to$ \EGPTSOneWith & \EGPTSOneLost/\EGPTSOneGained & \EGPTSOneHolm & \EGPTSOneShared & \EGPTPoolLinks & \EGPTPoolMean{} {\scriptsize \EGPTPoolCI} & \EGPTPoolLost/\EGPTPoolGained & \EGPTPoolP \\
Gemini 3.1 Pro & \EGeminiGapTexts{} $\to$ \EGeminiGapDamaged & \EGeminiHTwoaLost/\EGeminiHTwoaGained & \EGeminiHTwoaHolm & \EGeminiSOneOwn{} $\to$ \EGeminiSOneWith & \EGeminiSOneLost/\EGeminiSOneGained & \EGeminiSOneHolm & \EGeminiSOneShared & \EGeminiPoolLinks & \EGeminiPoolMean{} {\scriptsize \EGeminiPoolCI} & \EGeminiPoolLost/\EGeminiPoolGained & \EGeminiPoolP \\
Claude Opus 4.5 & \EOpusGapTexts{} $\to$ \EOpusGapDamaged & \EOpusHTwoaLost/\EOpusHTwoaGained & \EOpusHTwoaHolm & \EOpusSOneOwn{} $\to$ \EOpusSOneWith & \EOpusSOneLost/\EOpusSOneGained & \EOpusSOneHolm & \EOpusSOneShared & \EOpusPoolLinks & \EOpusPoolMean{} {\scriptsize \EOpusPoolCI} & \EOpusPoolLost/\EOpusPoolGained & \EOpusPoolP \\
\bottomrule
\end{tabular}}
\end{table*}
}{}

\paragraph{A second draw of removed links.} Table~\ref{tab:e123-seeds} repeats H2a on the second draw. Each judge lost
catches again and gained none: Gemini \EGeminiSOneOwn{} to \EGeminiSOneWith, GPT-5.5 \EGPTSOneOwn{} to
\EGPTSOneWith, Claude \EOpusSOneOwn{} to \EOpusSOneWith{} and Gemma \EGemmaSOneOwn{} to \EGemmaSOneWith{} (Holm $p$
\EGeminiSOneHolm, \EGPTSOneHolm, \EOpusSOneHolm{} and \EGemmaSOneHolm). The two draws share \EGPTSOneShared{} gap faults,
whose replies are identical under both, so the seed-1 result is new evidence only through its other 12 items. Counting
each link once, the drop over the \EGPTPoolLinks{} links in either gap set is \EGeminiPoolMean{} points for Gemini
(95\% interval \EGeminiPoolCI), \EGPTPoolMean{} for GPT-5.5 (\EGPTPoolCI), \EGemmaPoolMean{} for Gemma
(\EGemmaPoolCI) and \EOpusPoolMean{} for Claude (\EOpusPoolCI), with no link gained by any judge. On both draws Gemini lost
the most catches and Claude the fewest, but with this many items we do not rank the judges.

\paragraph{Presentation.} Writing ``no replacement recorded'' instead of ``in force'' did not bring any judge back to
its text-only level. Gemini caught \EGeminiGapNorecord{} (\EGeminiArmNorecordGained{} gained, \EGeminiArmNorecordLost{}
lost), GPT-5.5 \EGPTGapNorecord{} (\EGPTArmNorecordGained{} gained, \EGPTArmNorecordLost{} lost), Claude
\EOpusGapNorecord{} (\EOpusArmNorecordGained{} gained, \EOpusArmNorecordLost{} lost) and Gemma \EGemmaGapNorecord; no
change is significant before or after adjustment. Gains this small cannot be resolved with 24 pairs, so the result
is that the rewording did not restore the lost catches, not that it had no effect. In this condition the record line
is true: it reports only that no replacement is recorded. The drop seen here,
for judges that had both texts in front of them, is therefore produced by an omission in the record, without any
false statement. (The comparison with the texts alone was not a planned test; for each judge every item that changed
moved away from a catch.)

The other two presentations moved three of the judges. With the line headed as an unverified note, Gemini caught
\EGeminiGapLowauth{} gap faults and GPT-5.5 \EGPTGapLowauth{}; with a warning that the record omits some replacements,
\EGeminiGapCaveat{} and \EGPTGapCaveat. Every item that changed moved toward a catch (Gemini \EGeminiArmLowauthGained{}
and \EGeminiArmCaveatGained{} items, GPT-5.5 \EGPTArmLowauthGained{} and \EGPTArmCaveatGained), and all four tests are
significant after the Holm adjustment over the twelve presentation tests (Holm $p$ \EGeminiArmLowauthHolm,
\EGeminiArmCaveatHolm, \EGPTArmLowauthHolm{} and \EGPTArmCaveatHolm). Claude moved
the same way, to \EOpusGapLowauth{} and \EOpusGapCaveat{} (\EOpusArmLowauthGained{} items each, none lost; raw
$p$ \EOpusArmLowauthRaw, Holm $p$ \EOpusArmLowauthHolm). In each condition it caught seven of the
\EOpusHTwoaLost{} faults it had lost and one it had missed from the texts (Appendix~\ref{app:items}), equal to its
text-only level of \EOpusGapTexts; with only \EOpusHTwoaLost{} lost catches to recover, even this near-complete
recovery does not survive the Holm adjustment over the twelve tests. Gemma recovered less,
catching \EGemmaGapLowauth{} and \EGemmaGapCaveat{} against \EGemmaGapTexts{} from the texts alone (Holm $p$
\EGemmaArmLowauthHolm{} and \EGemmaArmCaveatHolm).

Where the catches came back, they came back with the judge's text-only false alarms. Under the authoritative record,
Gemini and GPT-5.5 raised no validity false alarm on decoys, because the record correctly lists the decoy's older RFC as
in force. With the unverified note, GPT-5.5 flagged \EGPTDecoyFALowauth{} decoys and Gemini \EGeminiDecoyFALowauth; with
the warning, \EGPTDecoyFACaveat{} and \EGeminiDecoyFACaveat, against \EGPTDecoyFATexts{} and \EGeminiDecoyFATexts{}
from the texts alone. Claude's rose from \EOpusDecoyFADamaged{} to \EOpusDecoyFALowauth{} and \EOpusDecoyFACaveat{}
(texts alone \EOpusDecoyFATexts). A decoy's older RFC was never replaced, so in every presentation its record line is
correct; a judge that weighed record against texts would have no reason to flag more decoys when told the
record might omit replacements. The change therefore looks more like a return to text-only behaviour than like better
use of the record. False alarms on controls and date controls stayed at one of 48 or fewer in every damaged-record
condition.

\paragraph{A code for conflicts.} Given a reason for a record that conflicts with the evidence (exploratory;
\S\ref{sec:studies}), the judges divided. Gemini gave it on \EGeminiConflictCodeGap{} of the 24 gap faults, GPT-5.5
on \EGPTConflictCodeGap, Gemma on \EGemmaConflictCodeGap{} and Claude on none. Right-reason catches on the gap faults
were \EGeminiGapConflict{} for Gemini, \EGPTGapConflict{} for GPT-5.5, \EOpusGapConflict{} for Claude and
\EGemmaGapConflict{} for Gemma, against \EGeminiGapDamaged, \EGPTGapDamaged, \EOpusGapDamaged{} and
\EGemmaGapDamaged{} without the code, and bare passes were \EGeminiBareConflict, \EGPTBareConflict,
\EOpusBareConflict{} and \EGemmaBareConflict{} of 24 (against \EGeminiBareDamaged, \EGPTBareDamaged, \EOpusBareDamaged{} and
\EGemmaBareDamaged{} without the code). The code was almost never used on correct answers (Gemini and
Gemma once each, on a decoy; no judge on a control or date control). For Gemini, then, many of the silent passes under
the authoritative line correspond to a conflict the judge registers when the format lets it say so; the other
three judges rarely or never used the code, although Claude's right-reason catches rose from \EOpusGapDamaged{} to
\EOpusGapConflict{} in this context (untested). With the code, every judge still caught fewer gap faults than from the texts
alone.

\paragraph{Comparison with the pilot.} In the pilot, Claude kept \RfcDeferWithStatusDamagedClaudeOpusFourFive{} of the
\RfcDeferOwnStatusDamagedClaudeOpusFourFive{} gap faults it had caught from the texts. Here it lost
\EOpusHTwoaLost. The two runs used different gap faults: the pilot's damaged record was built over 24 links, and only
\EOpusPilotShared{} of its \EOpusPilotPilotItems{} gap faults are gap faults here, and on those Claude caught
\EOpusPilotSharedCaught{} in this run; the other \EOpusPilotPilotOnly{} pilot items, all of which it caught in the pilot, are not
gap faults in this run. Most of the difference therefore comes from which links were dropped, and on the second draw
Claude again lost catches (\EOpusSOneLost) and gained none. What the comparison does show is that the choice of a dozen dropped links can turn a judge's small loss into a large one, so comparisons between judges on item
sets of this size, including those in our own tables, are provisional.
Appendix~\ref{app:items} lists every gap fault.

\subsection{Replacements, updates and surface similarity}
\label{sec:res-shortcut}

\begin{figure}[t]
\centering
\includegraphics[width=\linewidth]{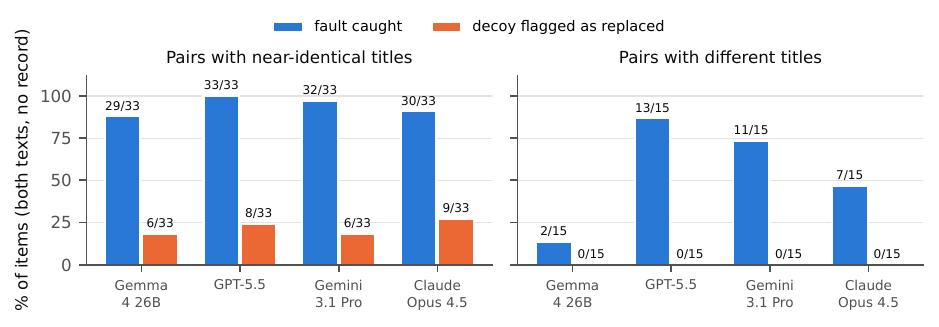}
\caption{Faults caught and decoys flagged as replaced, with both texts and no record, by whether the two RFCs of a pair have near-identical titles (the values of Table~\ref{tab:e123-decoy}). No judge flagged a decoy whose titles differ.}
\label{fig:decoy}
\end{figure}

Table~\ref{tab:e123-decoy} and Figure~\ref{fig:decoy} report the decoy test, with both texts and no record. Every judge caught more
faults than it flagged decoys (paired difference \EGPTHOne{} points for GPT-5.5, 95\% interval \EGPTHOneCI; Gemini
\EGeminiHOne, \EGeminiHOneCI; Claude \EOpusHOne, \EOpusHOneCI; Gemma \EGemmaHOne, \EGemmaHOneCI), so H1 holds. Because each decoy was
drawn from its link's title group, a judge that flagged every pair with near-identical titles would score zero here;
the weakness of H1 lies elsewhere, in abstract overlap, discussed below. The more informative comparison is within
the same-title pairs, where title similarity is held roughly constant. There GPT-5.5 caught \EGPTSameCatch{} faults and flagged \EGPTSameDecoy{}
decoys, Gemini \EGeminiSameCatch{} and \EGeminiSameDecoy, Claude \EOpusSameCatch{} and \EOpusSameDecoy, Gemma \EGemmaSameCatch{} and \EGemmaSameDecoy. Among
different-title pairs, no judge flagged a decoy, and catches differed widely (GPT-5.5 \EGPTDifferentCatch, Gemini \EGeminiDifferentCatch, Claude
\EOpusDifferentCatch, Gemma \EGemmaDifferentCatch).

\IfFileExists{generated/tab_e123_decoy.tex}{\begin{table*}[t]\centering\footnotesize
\caption{Shortcut test with both texts and no record. Faults and their matched decoys, split by whether the two RFCs of the pair share a (near-)identical title. Discrimination: catch minus decoy false alarm, paired by link, in points with a 95\% bootstrap interval over the 48 pairs.}
\label{tab:e123-decoy}
\resizebox{\textwidth}{!}{%
\begin{tabular}{lccccc}
\toprule
 & \multicolumn{2}{c}{Same title} & \multicolumn{2}{c}{Different title} & \\
\cmidrule(lr){2-3}\cmidrule(lr){4-5}
Judge & Fault catch & Decoy FA & Fault catch & Decoy FA & Discrimination \\
\midrule
Gemma 4 26B & \EGemmaSameCatch & \EGemmaSameDecoy & \EGemmaDifferentCatch & \EGemmaDifferentDecoy & \EGemmaHOne{} \EGemmaHOneCI \\
GPT-5.5 & \EGPTSameCatch & \EGPTSameDecoy & \EGPTDifferentCatch & \EGPTDifferentDecoy & \EGPTHOne{} \EGPTHOneCI \\
Gemini 3.1 Pro & \EGeminiSameCatch & \EGeminiSameDecoy & \EGeminiDifferentCatch & \EGeminiDifferentDecoy & \EGeminiHOne{} \EGeminiHOneCI \\
Claude Opus 4.5 & \EOpusSameCatch & \EOpusSameDecoy & \EOpusDifferentCatch & \EOpusDifferentDecoy & \EOpusHOne{} \EOpusHOneCI \\
\bottomrule
\end{tabular}}
\end{table*}
}{}

Our plan read a positive H1 as evidence that text-only catches are not a title shortcut. The data call for a more
qualified reading, and the analysis that supports it was not planned. Title similarity travels with abstract overlap:
in the text the judges saw, a replacing RFC usually repeats much of its predecessor's abstract (median similarity
\EAbsSimLinksSame{} for same-title links, \EAbsSimLinksDifferent{} for different-title links), while the decoys, which
were not matched on overlap, do not (\EAbsSimDecoysSame{} and \EAbsSimDecoysDifferent). Splitting at a similarity of
0.3, chosen after seeing the data, false alarms on same-title decoys were more frequent when the abstracts overlapped
(GPT-5.5 \EGPTDecoyFASimHigh{} against \EGPTDecoyFASimLow; Gemini \EGeminiDecoyFASimHigh{} against
\EGeminiDecoyFASimLow; Claude \EOpusDecoyFASimHigh{} against \EOpusDecoyFASimLow;
Gemma \EGemmaDecoyFASimHigh{} against \EGemmaDecoyFASimLow), and so were catches on different-title replacements (Claude
\EOpusDiffCatchSimHigh{} against \EOpusDiffCatchSimLow; Gemma \EGemmaDiffCatchSimHigh{} against \EGemmaDiffCatchSimLow;
GPT-5.5 \EGPTDiffCatchSimHigh{} against \EGPTDiffCatchSimLow; Gemini \EGeminiDiffCatchSimHigh{} against
\EGeminiDiffCatchSimLow). The high-overlap cells hold six or seven items, and we
report no test. The pattern is consistent with judges using the resemblance between the two documents as evidence of
replacement. In the RFC series that is often good evidence, so we hesitate to call it a shortcut, and the data cannot
separate the title from the abstract.

The decoys also have a labelling caveat. Some of the decoys the judges flagged are updates that introduce a new version of a
protocol, such as RFC~8210 (version~1 of the RPKI-to-Router protocol, updating RFC~6810), flagged by all four judges,
and RFC~9777 (MLDv2, updating RFC~2710), flagged by two. By the publisher's record the older RFC is in force, which is our gold. For a
question asking which document specifies the protocol, however, a reader could reasonably prefer the newer version.
Some of these false alarms may therefore be defensible judgements rather than errors.

\subsection{Scoring by reason}
\label{sec:res-scoring}

For the frontier judges given both texts, counting any finding on a fault instead of the right reason barely changes
the catch (GPT-5.5 \EGPTCatchTexts{} against \EGPTCatchAnyTexts, Gemini \EGeminiCatchTexts{} against \EGeminiCatchAnyTexts,
Claude \EOpusCatchTexts{} against \EOpusCatchAnyTexts). On correct answers the difference is large. With both texts, Claude raised some finding on
\EOpusCtrlAnyTexts{} of the controls that cite the replacing RFC, but a validity finding on \EOpusCtrlFATexts; Gemini
raised some finding on \EGeminiCtrlAnyTexts{} and a validity finding on \EGeminiCtrlFATexts; Gemma raised some finding on \EGemmaCtrlAnyTexts{} and a validity finding on \EGemmaCtrlFATexts. Gemma's findings on these
controls all gave the reason ``content error''. We suspect, without having tested it, that the question format provokes
them: the question names the old document's title, and the answer cites a document with a different number and
sometimes a different title. Under any-finding scoring Claude, Gemini and Gemma would appear to raise false alarms on about two in
five of these controls, none of which concern currency.

The scoring rule also changes the catch when the judge has less evidence. In the pilot, given the citation alone, Gemini raised a finding on
every fault but gave the right reason on \RfcCatchCiteOnlyGeminiThreeOnePro. Reason coding is also what makes the bare
passes of \S\ref{sec:res-gaps} visible as a separate category from misses that carry an irrelevant finding.

\subsection{Date controls}
\label{sec:res-dates}

The judges raised few or no validity false alarms on date controls (with both texts: GPT-5.5 \EGPTDateFATexts,
Gemini \EGeminiDateFATexts, Claude \EOpusDateFATexts{} and Gemma \EGemmaDateFATexts). In this run the controls
did not separate the judges. In the pilot they did. Qwen~2.5~32B caught \RfcCatchBothTextsQwenThreeTwoB{} faults with
both texts, a catch rate comparable to the frontier judges, but also flagged \RfcDateFABothTextsQwenThreeTwoB{} answers
that were correct on their date, for a currency discrimination (right-reason catch minus validity false alarms on date controls) of
\RfcDiscDateBothTextsQwenThreeTwoB{} points over all \RfcSample{} links against \RfcDiscDateBothTextsGeminiThreeOnePro{} for Gemini over the first 24.
Two worked examples in the prompt, one showing a correct answer being passed, reduced Qwen's date false alarms to \RfcDateFAFewShotQwenThreeTwoB. The result rests on one judge, and a catch rate alone would not
have revealed it.

\section{Discussion}
\label{sec:discussion}

\paragraph{The gap result.}
Within each of the four judges, the same gap faults were caught less often when the record line was presented as authoritative, and
most of the lost catches became passes with no finding. Some mundane explanations can be set aside. The items are
identical across conditions. The drop also occurs when the line says only that no replacement is recorded, so it does
not depend on the record stating anything false. And the decoy test shows that most text-only catches were not
reflexive objections to older documents, although some may rest on the documents' resemblance rather than on a
check of replacement (\S\ref{sec:res-shortcut}), so what the record removed is best read as findings of
replacement, whatever their basis. What the design cannot set aside, the small and partly shared item sets, is discussed in
\S\ref{sec:limitations}.

We cannot say why the judges behaved this way, and at least three explanations fit what we observed. The judge may
simply be relying on the record as the task presents it. It is shown an ``amendment record lookup'', nothing suggests
the record is incomplete, and trusting it is defensible; on correct answers, following it removed most decoy false
alarms (\S\ref{sec:res-correct}). On that reading the failure belongs to a record that does not disclose its own gaps,
and what the judge lacks is any sign of registering the conflict with the texts. The line may instead be more salient
than the evidence, being short, explicit and placed just before the task, while the replacement has to be inferred
from two abstracts; we did not vary position or length. Or the judge may treat the status line as part of the question
rather than as evidence to weigh. A bare pass is compatible with all three, and on its own it does not tell us whether the judge
noticed the conflict and resolved it in the record's favour or never noticed it. The exploratory conflict code gives
a partial answer that differs by judge: offered a way to say so, Gemini named the conflict on \EGeminiConflictCodeGap{}
of the 24 gap faults, while Claude never did and GPT-5.5 and Gemma rarely did (\S\ref{sec:res-gaps}). For Gemini, a
silent pass often hid a conflict it had registered; for the others, silence under the authoritative line is not
explained by the lack of a code. A conflict finding is still not a catch: Gemini's right-reason catches on the gap faults
in that context were \EGeminiGapConflict.

\paragraph{Presentation.}
The presentation conditions were meant to separate these explanations, and they partly do. Rewording the gap did not
restore the catches for any judge, which lends no support to the simplest account, that judges are misled by an
explicit claim of currency, although 24 pairs cannot exclude a small effect of wording. For Gemini and GPT-5.5,
lowering the record's standing or warning that it was incomplete restored most or all of the lost catches, and Claude
recovered seven of the eight it had lost, without reaching significance. This fits the first explanation: told that the
record might be wrong, these judges went back to the texts. It fits the salience explanation less well, since the note
and the warning leave the line as short, explicit and close to the task as before. For these three judges, however,
the decoy false alarms came back with the catches, to about their text-only level. This suggests that these
judges moved from relying on the record to relying on the texts, rather than weighing the two item by item. Gemma moved least,
catching at most \EGemmaGapCaveat{} gap faults however the damaged record was presented. Which judges respond in this way,
and why, is a question for more judges and more wordings than our one variant per condition.

Two related studies bear on these results. \citet{kapetanovic2026anchoring} found that a warning did not
reduce the total effect of judges' anchoring on prior scores, though it improved accuracy in one of their experiments.
Our warning differs in stating a specific and true defect of the record, with the contradicting evidence in view, and
either difference could matter. \citet{molfetta2026sycophants} report that larger models defer
more to authoritative but false legal information. Our one open-weight judge responded least to the
presentation of the record, but the hosted judges' sizes are not published, and four judges could not test an effect
of size in any case.

\paragraph{Resemblance between documents.}
The judges did not flag older documents merely
for having a newer neighbour, since they caught far more replacements than they flagged updates. Yet, in exploratory splits, their mistakes in
both directions tracked how alike the two documents looked. Given both texts, updates were flagged only when the titles were
near-identical, and Gemma rarely caught a replacement whose title had changed. Some of this reliance is reasonable,
since RFC replacements often do repeat their predecessor's title and abstract, and some flagged decoys are new
protocol versions that a reader might prefer to cite. The practical concern is transfer. An amended regulation may keep
its title, while a contract amendment can look nothing like the agreement it modifies. A judge evaluated on one kind
of collection may behave differently on the other, and nothing in a single-corpus evaluation, ours included, would show
it.

\paragraph{Scoring.}
Reason codes matter most on correct answers: under any-finding scoring, Claude, Gemini and Gemma would have appeared
to object to roughly two in five controls, all for reasons unrelated to currency. They also make the gap result
measurable. Any-finding scoring would credit a judge that calls a stale citation a ``content error'' with the catch;
reason codes separate that reply from a right-reason catch and from a bare pass, which is how most lost catches
could be seen to be bare passes.

\section{Practical implications}
\label{sec:guide}

These suggestions come from four judges on one corpus. Each is cheap to check with one's own judge, since the harness
runs a judge through the six contexts in about 1,200 calls.

An evaluation in which the judge receives a correct record says little about the judge: with a correct record our
judges' findings about currency nearly matched a lookup's. The more useful test removes links from the record and compares each judge with
itself on the affected items. If its catches there fall to the lookup's level, the judge contributes nothing beyond
the record on those items, and the completeness of the record becomes the real limit of the pipeline. That trade-off
can still be worth accepting, since the same record also spared the judges false alarms on decoys.

How the record is presented can matter, and not in the same way for every judge. For three of our judges a true
warning that the record was incomplete restored most catches on its gaps, but it also brought back the false alarms on
decoys that the record had removed, and the fourth judge recovered few. A builder considering such a warning should
test it on their own judge and items, count false alarms as well as catches, and make sure the warning is true.

Scoring should use reasons. Any-finding scoring inflates false alarms on correct answers and gives a judge that
raises an irrelevant objection on a fault the same catch as one that names the replacement. Where the record lists both replacements and
non-replacing updates, as the RFC index does, date controls and decoys come at no extra labelling cost.

If the replacements in a target collection tend to look unlike the documents they replace, catches on such pairs
should be measured separately, since our judges were weakest there.

\section{Limitations and threats to validity}
\label{sec:limitations}

\paragraph{Two draws, few items.} Each draw gives each judge 24 gap faults; the two draws share 12,
so the gap result rests on 36 distinct links. The drop held on both draws for every judge, but the pilot showed how much
a judge's result can depend on which links are dropped (\S\ref{sec:res-gaps}), and differences between judges of a few
items remain within what such variation could produce. The presentation conditions were run on the first draw only,
and Claude's presentation effects are not significant.

\paragraph{Single runs and decoding.} Hosted judges ran once each per prompt, through a proxy that sets and does not
report temperature and seed. Two checks bound the variation this allows: in the pilot, a repeated block changed 30 of
72 replies but only one catch, and requesting the pilot's prompts again changed the score of at most
\EOpusReaskChanged{} of a judge's \EOpusReaskItems{} items (\S\ref{sec:studies}). The second draw reuses the first
draw's text-only replies, and Gemma's client sometimes retried a cut-off reply without the API's JSON mode.

\paragraph{The conflict code.} The context with a conflict code was added after the main results existed, in one
wording, on one draw, without a planned test. It shows that one judge uses such a code often and the others rarely or
never; it cannot show how other wordings of such a code would fare.

\paragraph{The decoys are a partial control.} They are matched on dates, title similarity and, for most pairs, IETF
area, but not on abstract overlap, which differs sharply between replacements and updates. Some decoys are new
protocol versions, for which flagging the older RFC is a defensible reading of the question even though the
publisher's record says it was not replaced. The similarity analysis is exploratory, post hoc and small.

\paragraph{Construct.} The evidence is titles, dates and abstracts, not full documents. The error tested is always the
citation of a whole replaced document; partial amendments and chains are not covered. Links are dropped at random,
while real records fail in patterns (recent amendments, particular issuers). The question names the old document's
title, which may provoke some of the non-currency findings on controls. Two of the 24 gap faults are weaker items:
RFC~7142 (``Reclassification of RFC~1142 to Historic'') and RFC~6082 (``\ldots RFC~2482 is Historic'') name the
old RFC in their titles, so the replacement is stated rather than inferred, and as reclassification notices they do
not themselves specify the protocol, so the controls citing them are weaker. They account for two of Gemini's lost
catches and one each of GPT-5.5's and Claude's (Table~\ref{tab:e123-items}); without them every judge still lost
catches and gained none (Claude 7 lost, two-sided exact $p$ = .016). Under the damaged record Gemini missed both,
even though the newer RFC's title states the replacement.

\paragraph{Generalisation.} One corpus of technical standards, chosen because its publisher maintains an exact
record, and four judges. Legal collections add problems this test does not address, as the case study illustrates:
partial amendments, records extracted by machine, and answer keys that people disagree about. The title of the paper
asks a general question; the answer we give is limited to these judges, prompts and documents. In this corpus a
replacing RFC usually repeats its predecessor's title and abstract, so text-only catches are comparatively easy;
where amendments look little like the original, as in many contracts and regulations, both the text-only baseline
and what a record suppresses are untested.

\paragraph{Mechanism.} The presentation conditions keep the record fixed and vary single features, one prompt variant
each. They can show that presentation changed what a judge did; they cannot show why.

\paragraph{Provenance.} The harness and the text were produced with an AI assistant from the same developer as one of
the judges, Claude Opus~4.5 (\S\ref{sec:ai-use}). All judges received identical prompts and items, and the gold comes
from the publisher. After the pilot, the plan was drafted with that assistant, then reviewed and amended by the authors; it
was recorded internally rather than registered publicly.

\section{Conclusion}
\label{sec:conclusion}
We asked what LLM judges rely on when they check whether a cited document is still in force. On a corpus with an
exact publisher record, the judges we tested nearly matched a lookup when the record was correct, and on links we removed from
the record they caught markedly fewer replaced citations than they did from the texts alone, usually without comment,
on both of two random choices of removed links.
This happened even when the record line said, truthfully, only that no replacement was recorded, and a reason code
for a conflicting record was taken up by only one judge. Labelling the
record as unverified or incomplete brought three of the four judges most of the way back to the texts, their false alarms
included, and the fourth part of the way. An evaluation with a
correct record and any-finding scoring would have shown none of these behaviours. The test needs only a table of
replacements, and we release it so that others can run it on their own judges and collections.

\section{Data and code availability}
\label{sec:availability}
The repository (\repourl) holds the harness (\texttt{evalkit.py}), the run and analysis scripts, the sampled links and
decoys, every judge reply with its block metadata (model version, context, code hash), and the scripts that regenerate
every number, table and figure of the RFC study from those replies: the analysis script (\texttt{analyse\_final\_e123.py}) rescores every reply
from its raw text, checks the stored scores and regenerates the numbers and tables of \S\ref{sec:results} and Appendix~\ref{app:items}, from which
\texttt{make\_figures\_v6.py} draws Figures~\ref{fig:gap} and~\ref{fig:decoy};
\texttt{python -m suryabench.evalkit} (the harness, in the project's Python package) does the same for the pilot. The RFC Editor index is cited with its retrieval
date and content hash \citep{rfcindex2026} and included unchanged. The hypotheses, tests and readings were committed to
the project's development repository before the runs, and the amendment adding the second draw before any call on it.
That repository is private; the release contains the plan with its amendments and a provenance file listing the commit
and time at which the plan, each amendment and the blocks of results were first recorded. For the case study, the regulatory
documents are released as source URLs with content hashes, and the cases, the fixed answers, the record before and
after adjudication and the annotation sheets are released with the code. Code is under the Apache-2.0 licence; the
judge replies, the analysis outputs and the case study's cases and labels are under CC~BY~4.0.

\clearpage
\begin{appendices}

\renewcommand{\theHtable}{app\thesection.\arabic{table}}
\renewcommand{\theHfigure}{app\thesection.\arabic{figure}}
\section{The pilot}
\suppressfloats[t]
\label{app:pilot}
\label{app:study2}
\setcounter{table}{0}\setcounter{figure}{0}

Six judges (Qwen~2.5~32B, Gemma~4~26B, GPT-OSS~120B, Gemini~3.1~Pro, GPT-5.5 and Claude Opus~4.5) ran up to ten
contexts on the first 24 links before the final design was fixed; Qwen~2.5~32B ran all \RfcSample.
Table~\ref{tab:rfc-deference} gives the pilot's gap result, on the 12 gap faults per judge produced by a damaged
record over 24 links (24 gap faults over all \RfcSample{} links for Qwen~2.5~32B), and Table~\ref{tab:rfc-date} the currency discrimination against date controls in every
context. The pilot tables use the pilot's own word ``reviewer'' for judge. The pilot's gap faults overlap only partly with the final run's (\S\ref{sec:res-gaps}).

\IfFileExists{generated/tab_rfc_deference.tex}{\begin{table*}[t]\centering\small
\caption{Where the record is silent. Left: right-reason catches on all faults with no record (\texttt{both-texts}), a correct record line (\texttt{status}) and a record with half its links dropped (\texttt{status-damaged}, seed 0); a lookup in the damaged record catches exactly half. Right: the faults whose link was dropped, with no record, with the damaged record (the line reads ``in force''), and with the line worded ``no replacement recorded''; same items in every column. Bare: damaged-record replies on these faults that pass the answer with no finding. Qwen~32B ran 48 links, the others 24.}
\label{tab:rfc-deference}
\resizebox{\textwidth}{!}{%
\begin{tabular}{l ccc cccc}
\toprule
 & \multicolumn{3}{c}{All faults} & \multicolumn{4}{c}{Faults whose link was dropped} \\
\cmidrule(lr){2-4}\cmidrule(lr){5-8}
Reviewer & No record & Correct & Damaged & No record & ``In force'' & ``No replacement'' & Bare \\
\midrule
Qwen 32B & 42/48 & 48/48 & 33/48 & 22/24 & 9/24 & 5/24 & 6 \\
Gemma 4 26B & 16/24 & 24/24 & 12/24 & 10/12 & 0/12 & 0/12 & 12 \\
GPT-OSS 120B & 16/24 & 18/24 & 15/24 & 8/12 & 6/12 & -- & 4 \\
Gemini 3.1 Pro & 22/24 & 24/24 & 12/24 & 12/12 & 0/12 & 3/12 & 12 \\
GPT-5.5 & 23/24 & -- & 16/24 & 11/12 & 4/12 & 6/12 & 8 \\
Claude Opus 4.5 & 19/24 & 24/24 & 22/24 & 11/12 & 10/12 & 9/12 & 1 \\
\bottomrule
\end{tabular}}
\end{table*}
}{}
\IfFileExists{generated/tab_rfc_disc_date.tex}{\begin{table*}[t]\centering\scriptsize
\caption{Currency discrimination by context (points): right-reason catch on faults minus validity false alarms on date controls, which cite the same old document as of a date before it was replaced. Mean over links.}
\label{tab:rfc-date}
\setlength{\tabcolsep}{3pt}
\resizebox{\textwidth}{!}{%
\begin{tabular}{lrrrrrr}
\toprule
Context & Qwen 32B & Gemma 4 26B & GPT-OSS 120B & Gemini 3.1 Pro & GPT-5.5 & Claude Opus 4.5 \\
\midrule
\texttt{cite-only} & $-$38 & +12 & -- & +46 & -- & +21 \\
\texttt{cited-text} & $-$69 & -- & -- & -- & -- & +25 \\
\texttt{both-texts} & $-$2 & +67 & +67 & +92 & +96 & +71 \\
\texttt{both-masked} & $-$19 & +62 & +54 & +83 & +92 & +75 \\
\texttt{both-explicit} & $-$2 & -- & +88 & +100 & +100 & +92 \\
\texttt{status} & +77 & +100 & +71 & +100 & -- & +100 \\
\texttt{status-damaged} & +46 & +50 & +62 & +50 & +67 & +92 \\
\texttt{norecord-damaged} & +27 & +50 & -- & +62 & +75 & +83 \\
\texttt{check-first} & +23 & +75 & -- & +100 & -- & +88 \\
\texttt{few-shot} & +79 & -- & -- & -- & -- & +75 \\
\bottomrule
\end{tabular}}
\end{table*}
}{}

\paragraph{Runs.} Hosted models ran once through the Kaggle Benchmarks proxy, which sets temperature and seed itself;
GPT-5.5's two damaged-record blocks ran after a change to the proxy call (an output cap, and the system and user text
sent as two messages), with the prompt text unchanged. An accidental repeat of one block (Gemini~3.1~Pro, citation only;
the first run is the one reported) differed by one catch. Every block records the model version, the context and the
harness's code hash. A reply that cannot be parsed counts as a miss on a fault and a false alarm on a control; GPT-OSS
had two to four such replies per block, the other reported judges at most two. Three further runs stopped before the
two damaged-record contexts and are not analysed; their blocks are in the repository.

\paragraph{Items.} Two of the 48 replacing RFCs name the old RFC in their title (RFC~7142, RFC~6082); titles are not
filtered, so these two links are explicit wherever they appear: for Qwen~2.5~32B in the pilot, and for all four judges
in the final runs, where both are gap faults (Table~\ref{tab:e123-items}). The sentence filter of
\texttt{both-texts} also removes some sentences that mention updates in general.

\paragraph{Links without a shared title.} On the 6 of the first 24 links whose successor has a different title, with
both texts, GPT-5.5 caught \RfcTitleDifferentBothTextsGPTFiveFive, Gemini \RfcTitleDifferentBothTextsGeminiThreeOnePro,
Claude \RfcTitleDifferentBothTextsClaudeOpusFourFive, GPT-OSS \RfcTitleDifferentBothTextsGPTOSSOneTwoZeroB{} and Gemma
\RfcTitleDifferentBothTextsGemmaFourTwoSixB. Masking the document numbers changed little for the frontier judges
(Gemini \RfcCatchBothMaskedGeminiThreeOnePro, Claude \RfcCatchBothMaskedClaudeOpusFourFive, GPT-5.5
\RfcCatchBothMaskedGPTFiveFive) and cost GPT-OSS a few catches (\RfcCatchBothMaskedGPTOSSOneTwoZeroB).

\paragraph{Controls citing the successor.} Table~\ref{tab:rfc-current} gives discrimination against controls that cite
the replacing document. Given the documents, the frontier judges, GPT-OSS and Gemma rarely gave a validity reason on
these controls; Qwen~2.5~32B gave one on \RfcFABothTextsQwenThreeTwoB. Other findings on them (``not applicable'',
``unsupported'') are expected, because the question names the old document's title.

\IfFileExists{generated/tab_rfc_disc_validity.tex}{\begin{table*}[t]\centering\scriptsize
\caption{Discrimination against current-document controls (points): right-reason catch minus validity false alarms on answers citing the replacing document. Mean over links.}
\label{tab:rfc-current}
\setlength{\tabcolsep}{3pt}
\resizebox{\textwidth}{!}{%
\begin{tabular}{lrrrrrr}
\toprule
Context & Qwen 32B & Gemma 4 26B & GPT-OSS 120B & Gemini 3.1 Pro & GPT-5.5 & Claude Opus 4.5 \\
\midrule
\texttt{cite-only} & $-$31 & +12 & -- & +46 & -- & +21 \\
\texttt{cited-text} & $-$69 & -- & -- & -- & -- & +21 \\
\texttt{both-texts} & +40 & +67 & +62 & +92 & +96 & +75 \\
\texttt{both-masked} & +35 & +62 & +50 & +83 & +92 & +75 \\
\texttt{both-explicit} & +46 & -- & +83 & +100 & +100 & +100 \\
\texttt{status} & +100 & +100 & +75 & +100 & -- & +100 \\
\texttt{status-damaged} & +69 & +50 & +62 & +50 & +67 & +92 \\
\texttt{norecord-damaged} & +60 & +50 & -- & +62 & +75 & +88 \\
\texttt{check-first} & +75 & +75 & -- & +100 & -- & +88 \\
\texttt{few-shot} & +88 & -- & -- & -- & -- & +75 \\
\bottomrule
\end{tabular}}
\end{table*}
}{}

\section{Gap faults item by item}
\suppressfloats[t]
\label{app:items}
\setcounter{table}{0}\setcounter{figure}{0}
Table~\ref{tab:e123-items} gives, for each judge, the outcome on every gap fault under no record, the four
presentations of the damaged record and the exploratory context with a conflict code. Every reply, with its raw text,
is in the repository (\texttt{50\_analysis/results/evalkit\_rfc\_v3/}; second draw in \texttt{evalkit\_rfc\_v3\_seed1/}).
The outcome of every link under both draws, with the draws it belongs to, is in
\texttt{evalkit\_rfc\_v3/\allowbreak analysis/\allowbreak seed1\_links.csv}.

\IfFileExists{generated/tab_e123_items.tex}{\begin{table*}[t]\centering\footnotesize
\caption{Per-item outcomes on the gap faults (\ding{51} caught for the right reason, $\cdot$ missed, -- not run). Within each judge, the six marks are, in order T F N U W C: texts only (no record), ``in force'', ``no replacement recorded'', unverified note, record with warning, record with the conflict code (a conflict finding is not a catch).}
\label{tab:e123-items}
\resizebox{\textwidth}{!}{%
\begin{tabular}{lcccc}
\toprule
Link (RFC) & Gemma 4 26B & GPT-5.5 & Gemini 3.1 Pro & Claude Opus 4.5 \\
 & {\scriptsize T F N U W C} & {\scriptsize T F N U W C} & {\scriptsize T F N U W C} & {\scriptsize T F N U W C} \\
\midrule
6312$\to$6342 & $\cdot$ $\cdot$ $\cdot$ $\cdot$ $\cdot$ $\cdot$ & \ding{51} $\cdot$ $\cdot$ \ding{51} \ding{51} $\cdot$ & \ding{51} $\cdot$ $\cdot$ $\cdot$ \ding{51} $\cdot$ & $\cdot$ $\cdot$ $\cdot$ $\cdot$ $\cdot$ $\cdot$ \\
8773$\to$9973 & $\cdot$ $\cdot$ $\cdot$ $\cdot$ $\cdot$ $\cdot$ & \ding{51} $\cdot$ $\cdot$ \ding{51} \ding{51} $\cdot$ & \ding{51} $\cdot$ $\cdot$ $\cdot$ $\cdot$ $\cdot$ & \ding{51} $\cdot$ \ding{51} \ding{51} \ding{51} \ding{51} \\
5046$\to$7145 & \ding{51} $\cdot$ $\cdot$ $\cdot$ $\cdot$ $\cdot$ & \ding{51} $\cdot$ $\cdot$ \ding{51} \ding{51} $\cdot$ & \ding{51} $\cdot$ $\cdot$ \ding{51} \ding{51} $\cdot$ & \ding{51} \ding{51} \ding{51} \ding{51} \ding{51} \ding{51} \\
8407$\to$9907 & \ding{51} $\cdot$ $\cdot$ \ding{51} \ding{51} $\cdot$ & \ding{51} \ding{51} $\cdot$ \ding{51} \ding{51} $\cdot$ & \ding{51} $\cdot$ $\cdot$ \ding{51} \ding{51} $\cdot$ & \ding{51} \ding{51} \ding{51} \ding{51} \ding{51} \ding{51} \\
3205$\to$9205 & $\cdot$ $\cdot$ $\cdot$ $\cdot$ $\cdot$ $\cdot$ & \ding{51} $\cdot$ $\cdot$ \ding{51} \ding{51} $\cdot$ & \ding{51} $\cdot$ $\cdot$ \ding{51} \ding{51} $\cdot$ & $\cdot$ $\cdot$ $\cdot$ $\cdot$ $\cdot$ $\cdot$ \\
2821$\to$5321 & \ding{51} $\cdot$ $\cdot$ $\cdot$ $\cdot$ $\cdot$ & \ding{51} \ding{51} \ding{51} \ding{51} \ding{51} $\cdot$ & \ding{51} $\cdot$ \ding{51} \ding{51} \ding{51} $\cdot$ & \ding{51} $\cdot$ $\cdot$ \ding{51} \ding{51} $\cdot$ \\
4402$\to$7802 & \ding{51} $\cdot$ $\cdot$ \ding{51} $\cdot$ $\cdot$ & \ding{51} \ding{51} \ding{51} \ding{51} \ding{51} \ding{51} & \ding{51} $\cdot$ \ding{51} \ding{51} \ding{51} $\cdot$ & \ding{51} $\cdot$ \ding{51} \ding{51} \ding{51} \ding{51} \\
5157$\to$7707 & $\cdot$ $\cdot$ $\cdot$ $\cdot$ $\cdot$ $\cdot$ & \ding{51} $\cdot$ $\cdot$ \ding{51} \ding{51} $\cdot$ & \ding{51} $\cdot$ $\cdot$ $\cdot$ \ding{51} $\cdot$ & $\cdot$ $\cdot$ $\cdot$ $\cdot$ $\cdot$ $\cdot$ \\
2837$\to$4044 & $\cdot$ $\cdot$ $\cdot$ $\cdot$ $\cdot$ $\cdot$ & \ding{51} $\cdot$ $\cdot$ \ding{51} \ding{51} $\cdot$ & \ding{51} $\cdot$ $\cdot$ $\cdot$ \ding{51} $\cdot$ & $\cdot$ $\cdot$ $\cdot$ $\cdot$ $\cdot$ $\cdot$ \\
2621$\to$4671 & $\cdot$ $\cdot$ $\cdot$ $\cdot$ $\cdot$ $\cdot$ & \ding{51} $\cdot$ $\cdot$ \ding{51} \ding{51} $\cdot$ & \ding{51} $\cdot$ $\cdot$ $\cdot$ \ding{51} $\cdot$ & $\cdot$ $\cdot$ $\cdot$ $\cdot$ $\cdot$ $\cdot$ \\
2858$\to$4760 & \ding{51} $\cdot$ $\cdot$ $\cdot$ \ding{51} $\cdot$ & \ding{51} \ding{51} \ding{51} \ding{51} \ding{51} \ding{51} & \ding{51} $\cdot$ \ding{51} \ding{51} \ding{51} $\cdot$ & \ding{51} \ding{51} \ding{51} \ding{51} \ding{51} \ding{51} \\
2478$\to$4178 & \ding{51} $\cdot$ $\cdot$ $\cdot$ $\cdot$ $\cdot$ & \ding{51} \ding{51} \ding{51} \ding{51} \ding{51} $\cdot$ & \ding{51} $\cdot$ $\cdot$ \ding{51} \ding{51} $\cdot$ & \ding{51} $\cdot$ $\cdot$ \ding{51} \ding{51} $\cdot$ \\
1657$\to$4273 & $\cdot$ $\cdot$ $\cdot$ $\cdot$ $\cdot$ $\cdot$ & \ding{51} $\cdot$ $\cdot$ \ding{51} \ding{51} $\cdot$ & \ding{51} $\cdot$ $\cdot$ \ding{51} \ding{51} $\cdot$ & \ding{51} $\cdot$ $\cdot$ $\cdot$ \ding{51} $\cdot$ \\
1142$\to$7142 & $\cdot$ $\cdot$ $\cdot$ $\cdot$ $\cdot$ $\cdot$ & $\cdot$ $\cdot$ $\cdot$ \ding{51} $\cdot$ $\cdot$ & \ding{51} $\cdot$ $\cdot$ \ding{51} \ding{51} $\cdot$ & \ding{51} $\cdot$ \ding{51} \ding{51} \ding{51} $\cdot$ \\
8941$\to$9651 & \ding{51} $\cdot$ $\cdot$ $\cdot$ \ding{51} $\cdot$ & \ding{51} \ding{51} \ding{51} \ding{51} \ding{51} \ding{51} & \ding{51} $\cdot$ \ding{51} \ding{51} \ding{51} $\cdot$ & \ding{51} \ding{51} $\cdot$ \ding{51} \ding{51} \ding{51} \\
5430$\to$6460 & \ding{51} $\cdot$ $\cdot$ $\cdot$ $\cdot$ $\cdot$ & \ding{51} $\cdot$ $\cdot$ \ding{51} \ding{51} \ding{51} & \ding{51} $\cdot$ \ding{51} \ding{51} \ding{51} $\cdot$ & \ding{51} \ding{51} \ding{51} \ding{51} \ding{51} \ding{51} \\
2482$\to$6082 & $\cdot$ $\cdot$ $\cdot$ $\cdot$ $\cdot$ $\cdot$ & \ding{51} $\cdot$ $\cdot$ \ding{51} \ding{51} $\cdot$ & \ding{51} $\cdot$ $\cdot$ \ding{51} \ding{51} $\cdot$ & \ding{51} \ding{51} \ding{51} \ding{51} \ding{51} \ding{51} \\
3984$\to$6184 & \ding{51} $\cdot$ $\cdot$ $\cdot$ $\cdot$ $\cdot$ & \ding{51} \ding{51} \ding{51} \ding{51} \ding{51} \ding{51} & \ding{51} \ding{51} \ding{51} \ding{51} \ding{51} $\cdot$ & \ding{51} \ding{51} \ding{51} \ding{51} \ding{51} \ding{51} \\
4244$\to$7044 & \ding{51} $\cdot$ $\cdot$ $\cdot$ $\cdot$ $\cdot$ & \ding{51} \ding{51} \ding{51} \ding{51} \ding{51} \ding{51} & \ding{51} $\cdot$ $\cdot$ \ding{51} \ding{51} $\cdot$ & \ding{51} \ding{51} \ding{51} \ding{51} \ding{51} \ding{51} \\
3230$\to$9530 & $\cdot$ $\cdot$ $\cdot$ $\cdot$ $\cdot$ $\cdot$ & \ding{51} $\cdot$ $\cdot$ \ding{51} \ding{51} $\cdot$ & \ding{51} $\cdot$ $\cdot$ \ding{51} \ding{51} $\cdot$ & $\cdot$ $\cdot$ $\cdot$ \ding{51} \ding{51} $\cdot$ \\
7613$\to$8265 & \ding{51} $\cdot$ $\cdot$ $\cdot$ \ding{51} $\cdot$ & \ding{51} $\cdot$ \ding{51} \ding{51} \ding{51} \ding{51} & \ding{51} $\cdot$ $\cdot$ \ding{51} \ding{51} $\cdot$ & \ding{51} \ding{51} \ding{51} \ding{51} \ding{51} \ding{51} \\
2767$\to$6535 & $\cdot$ $\cdot$ $\cdot$ $\cdot$ $\cdot$ $\cdot$ & \ding{51} $\cdot$ $\cdot$ \ding{51} \ding{51} $\cdot$ & $\cdot$ $\cdot$ $\cdot$ $\cdot$ $\cdot$ $\cdot$ & $\cdot$ $\cdot$ $\cdot$ $\cdot$ $\cdot$ $\cdot$ \\
8843$\to$9143 & \ding{51} $\cdot$ $\cdot$ $\cdot$ \ding{51} $\cdot$ & \ding{51} $\cdot$ \ding{51} \ding{51} \ding{51} $\cdot$ & \ding{51} $\cdot$ $\cdot$ \ding{51} \ding{51} $\cdot$ & \ding{51} $\cdot$ $\cdot$ \ding{51} \ding{51} \ding{51} \\
2910$\to$8010 & \ding{51} $\cdot$ $\cdot$ $\cdot$ $\cdot$ $\cdot$ & \ding{51} \ding{51} \ding{51} \ding{51} \ding{51} $\cdot$ & \ding{51} $\cdot$ $\cdot$ \ding{51} \ding{51} $\cdot$ & \ding{51} $\cdot$ $\cdot$ \ding{51} $\cdot$ \ding{51} \\
\bottomrule
\end{tabular}}
\end{table*}
}{}

\section{Prompt}
\label{app:prompt-rfc}
\setcounter{table}{0}\setcounter{figure}{0}
Figure~\ref{fig:prompt} gives the system text and the full prompt for one gap fault under the authoritative line, as sent, and the record
block of each damaged context. The other contexts differ only in the record block, which is absent from
\texttt{both-texts}, except \texttt{conflict-damaged}, whose reason list ends with one more line:
``\texttt{- record\_conflict: the record and the evidence disagree, or whether the cited document is still current cannot
be confirmed}''.

\IfFileExists{generated/prompt_example.tex}{\begin{figure}[p]
{\fontsize{6.5}{7.3}\selectfont\begin{verbatim}
You review answers that cite documents. Be precise and brief. Respond
  in strict JSON.

[QUESTION]
As of 2026-10-01, which document specifies "Mobile Networks
  Considerations for IPv6 Deployment"?
[ANSWER UNDER REVIEW]
The document that specifies this is RFC6312, "Mobile Networks
  Considerations for IPv6 Deployment" (2011-07).
[EVIDENCE]
[RFC6312] Mobile Networks Considerations for IPv6 Deployment
  (published 2011-07)
Mobile Internet access from smartphones and other mobile devices is
  accelerating the exhaustion of IPv4 addresses. IPv6 is widely seen
  as crucial for the continued operation and growth of the Internet,
  and in particular, it is critical in mobile networks. This document
  discusses the issues that arise when deploying IPv6 in mobile
  networks. Hence, this document can be a useful reference for service
  providers and network designers. This document is not an Internet
  Standards Track specification; it is published for informational
  purposes.

[RFC6342] Mobile Networks Considerations for IPv6 Deployment
  (published 2011-08)
Mobile Internet access from smartphones and other mobile devices is
  accelerating the exhaustion of IPv4 addresses. IPv6 is widely seen
  as crucial for the continued operation and growth of the Internet,
  and in particular, it is critical in mobile networks. This document
  discusses the issues that arise when deploying IPv6 in mobile
  networks. Hence, this document can be a useful reference for service
  providers and network designers. This document is not an Internet
  Standards Track specification; it is published for informational
  purposes.
[AMENDMENT RECORD LOOKUP OF THE CITED DOCUMENT]
- RFC6312: status on 2026-10-01: in_force
[TASK]
Decide whether the answer is valid as of the date in the question. For
  each problem give a finding naming the cited document and one reason
  from:
- superseded: the cited document had been replaced or obsoleted by a
  later one on the date asked about
- not_yet_published: the cited document did not yet exist on the date
  asked about
- not_applicable: the cited document does not address the question
- content_error: the answer misstates what the document says
- unsupported: the evidence does not support the answer
- other: any other problem
Reply as JSON: {"verdict": "PASS" or "FLAG", "findings": [{"doc_id":
  "...", "reason": "...", "rationale": "..."}]}
\end{verbatim}}
{\fontsize{6.5}{7.3}\selectfont\begin{verbatim}
[AMENDMENT RECORD LOOKUP OF THE CITED DOCUMENT]
- RFC6312: status on 2026-10-01: in_force

[AMENDMENT RECORD LOOKUP OF THE CITED DOCUMENT]
- RFC6312: status on 2026-10-01: no_replacement_recorded

[UNVERIFIED NOTE FROM AN AUTOMATED TAGGER ABOUT THE CITED DOCUMENT]
- RFC6312: status on 2026-10-01: in_force

[AMENDMENT RECORD LOOKUP OF THE CITED DOCUMENT (this record is
  incomplete: it omits some replacements, so a document it lists as in
  force may have been replaced)]
- RFC6312: status on 2026-10-01: in_force
\end{verbatim}}
\caption{Top: system text and prompt for one gap fault under the authoritative line. Bottom: the record block of each damaged context, in table order.}
\label{fig:prompt}
\end{figure}
}{}

\clearpage
\section{Case study: Indian rooftop-solar regulation}
\label{app:case}
\setcounter{table}{0}\setcounter{figure}{0}

\noindent\fbox{\parbox{\dimexpr\linewidth-2\fboxsep-2\fboxrule}{\small
\textbf{Status: descriptive only.} The test was first developed on this setting, where no curated consolidation
exists and the amendment record had to be extracted from the documents by an AI pipeline. Its quality checks
failed: people reviewing the AI-built answer key reached a different outcome on \NDroppedDecision{} of
\NCandidateCases{} candidate cases, only \PctControlsCorrect{} of the checked ``correct'' answers were judged correct,
and a person applying the right-reason rule agreed with the automatic scorer on \ScorerCheckPctObserved{} of replies on the development set, against a planned threshold of 95\%. None of its numbers is used as evidence for the
hypotheses or findings of \S\ref{sec:results}. It is reported because its failures, annotation disagreements and
negative results show what the test needs from an uncurated amendment record; its planned analyses and readings are
kept as they were fixed, so that nothing is selected after the fact. The full report (reference data and their audit,
conditions, planned analyses and results, readings fixed in advance, prompts, per-clause results, annotation
agreement, deviations from the plan and the data statement) is in the supplementary material.}}
\medskip

Two consequences follow for reading the numbers below. False alarms on correct answers cannot be taken as the judge's
errors. For example, Claude Opus~4.5 raised some finding on \RobFlaggedClaudeOpusRFour{} correct answers when given the
status line, but most of its findings were not about validity (\RobNonValidityClaudeOpus), and only a corrected key
could say whether they were wrong. Likewise, the large differences between any-finding and right-reason catches in this
study are real outputs on this key, but we do not use them to support \S\ref{sec:res-scoring}, which rests on the RFC
data. The text below is adapted from the study's own report, written before the RFC study, and keeps some of
its terms (reviewer for judge, clause for document); its table, like the supplementary material, also says registry
for record. Its reason list differs from the RFC study's, and
a ``validity'' reason there also includes wrong jurisdiction and not applicable.

\subsection{Setting}
\label{sec:study1}
\label{sec:dataset}
\paragraph{Documents and record.}
The corpus holds \NCorpusDocs{} regulatory instruments on rooftop solar from \NCorpusJurisdictions{} Indian
jurisdictions, superseded ones included. No consolidated record exists, so an AI pipeline extracted one from the
documents: \NLinksRegistry{} clause-level links, each from an old clause to the clause that changes it, with a relation
and an effective date. Two authors checked a random sample of the links blind; \NLinksSampleNotALink{} of
\NLinksSample{} were judged to be no change at all, and every link the test relies on was checked and corrected
(supplementary material, \S S1.1). The links the test relies on therefore follow the adjudicated audit, but other links may be wrong, and nothing
measures how many the record misses, which is also the usual situation in practice.

\paragraph{Items.}
A case is a synthetic solar application (a bill, an installer proposal and a utility record) with a date and a
jurisdiction, labelled by rule functions whose parameters are quoted from the clauses. Its answer is an
approve-or-reject decision with the clauses it rests on. A fault cites a replaced clause in place of its successor; its
matched control is the correct decision citing the current clause. The case study has no date control, because each case has
one date. People audited the answer key before the test set was sealed: the AI-built key reached a different outcome
from the adjudicated one on \NDroppedDecision{} of \NCandidateCases{} candidate cases, which were dropped. The sealed
set has \NTestCases{} cases on \NClauses{} distinct replaced clauses from \NStates{} jurisdictions; the record links
\NRegistryFlagged{} of the \NClauses, so the other replaced clauses are gaps the record already has.

\paragraph{What the reviewer sees.}
The main conditions are four contexts of the test (\S\ref{sec:framework}): both versions of the clause (\rung{R2}, \texttt{both-texts}),
both versions plus the record's status line for each cited clause (\rung{R4}, \texttt{status}), and the same line from
a damaged record (\rung{R4}-damage, \texttt{status-damaged}; \rung{R4n}-damage, \texttt{norecord-damaged}). The record
is damaged by dropping \DamageDropTwentyFive\% or \DamageDropFifty\% of its links, or adding \DamageSpuriousTen\%
spurious ones, over \NDamageSeeds{} seeds. A deterministic lookup in the same record (\rung{R5}, no language model)
flags exactly the clauses the record marks as replaced; it is the reference for what the record alone achieves.
Further conditions (the old clause alone, source metadata, the amending instrument's words) are in the supplementary
material (\S S1.2).

\paragraph{Reviewers and scoring.}
Seven reviewers judged the same items: \texttt{llama3.1:8b}, \texttt{qwen2.5:7b}, \texttt{qwen2.5:14b},
\texttt{gemma-4-26b-a4b-it}, \texttt{gemini-3.5-flash-lite} (on \rung{R0}, \rung{R2} and \rung{R4} only),
\texttt{qwen2.5:32b} and Claude Opus~4.5, with greedy decoding except for Claude, which ran through a proxy that sets
temperature and seed itself. A fault is caught when a finding names the replaced
clause with the reason ``not in force''. The unit of inference is the distinct replaced clause: rates are averaged within
a clause, then over clauses. The questions, analyses and readings were fixed in a dated, version-controlled plan before
any model saw a test item; the planned tests, all reported, are in the supplementary material (\S S1.3 and S1.4).

\subsection{Descriptive results}
\label{sec:res-robust}
\begin{figure}[t]
\centering
\includegraphics[width=\linewidth]{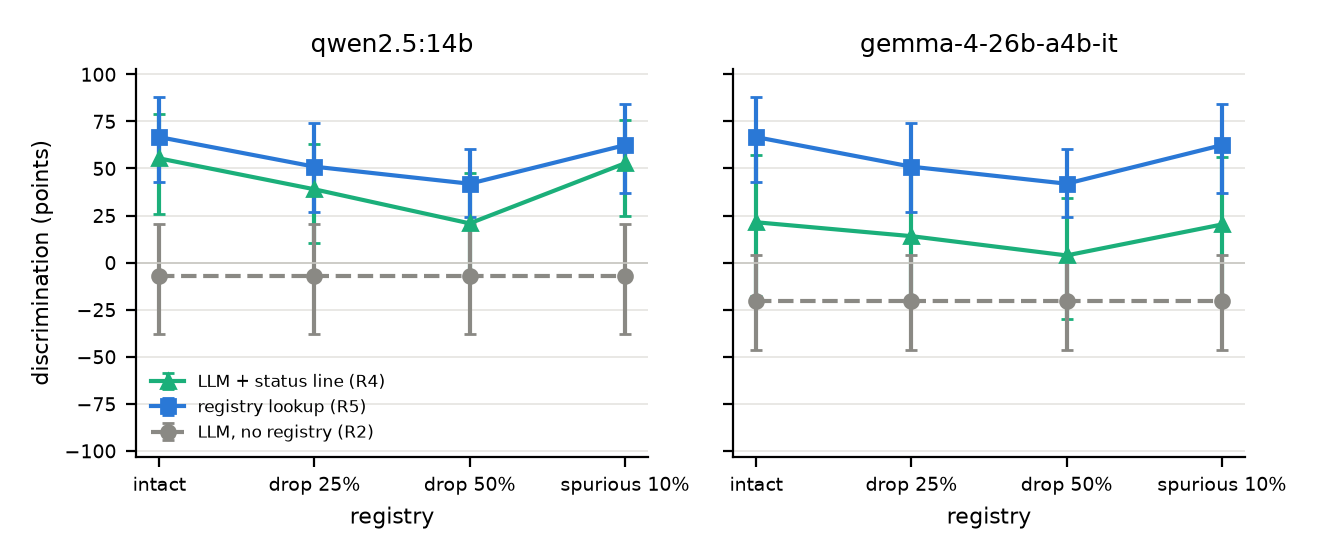}
\caption{Case study: discrimination (right-reason catch minus false alarms on matched correct answers) as the record
degrades, for the judge given the status line (\rung{R4}), a lookup in the same record (\rung{R5}), and the judge with
no record (\rung{R2}). Clause-averaged, damage seeds averaged within clause, 95\% bootstrap intervals over
\NClauses{} clauses.}
\label{fig:damage}
\end{figure}

Table~\ref{tab:robust} summarises the findings across the seven judges on the same items. Figure~\ref{fig:damage}
compares the judge given the record's status line with a lookup in the same record as the record degrades.

\IfFileExists{generated/tab_robust.tex}{\begin{table*}[t]\centering\scriptsize
\caption{The findings across all models, on the same items (pooled k/n over S-both faults, where both versions of the clause are shown, or matched correct answers; 50\%-dropped arms pooled over each model's damage seeds). The five planned models come first; the last two columns were added after the analysis plan was fixed. Gains and losses compare each fault and seed with the damaged lookup. The dropped-link catch is clause-averaged over the faults the intact registry catches and the damaged one drops.}
\label{tab:robust}
\resizebox{\textwidth}{!}{%
\begin{tabular}{lccccccc}
\toprule
Finding & Llama 8B & Qwen 7B & Qwen 14B & Gemma 4 26B & Flash-Lite & Qwen 32B & Claude Opus 4.5 \\
\midrule
Right-reason catch, both versions (R2) & 0/42 & 5/42 & 13/42 & 9/42 & 19/42 & 4/42 & 34/42 \\
Any-finding catch, R2 & 24/42 & 14/42 & 26/42 & 32/42 & 35/42 & 35/42 & 42/42 \\
Scoring gap at R2, any minus right reason (points) & +57 & +21 & +31 & +55 & +38 & +74 & +19 \\
Right-reason catch, status line (R4) & 3/42 & 2/42 & 32/42 & 25/42 & 33/42 & 29/42 & 35/42 \\
Correct answers flagged, R4 & 19/42 & 2/42 & 10/42 & 20/42 & 19/42 & 23/42 & 37/42 \\
R4 findings on correct answers not about validity & 13/25 & 0/2 & 4/11 & 28/30 & 19/22 & 21/24 & 44/55 \\
Catches gained / lost vs lookup, 50\% dropped & +0 / $-$15 & +2 / $-$15 & +13 / $-$13 & +25 / $-$24 & -- & +4 / $-$8 & +14 / $-$0 \\
Catch on dropped links, R2 $\to$ R4-damage, \% (clauses) & 0 $\to$ 0 (6) & 33 $\to$ 17 (6) & 41 $\to$ 7 (9) & 32 $\to$ 31 (9) & -- & 13 $\to$ 0 (9) & 92 $\to$ 67 (6) \\
\bottomrule
\end{tabular}}
\end{table*}
}{}

\clearpage

\end{appendices}

\bibliography{references}

\begin{thebibliography}{37}
\providecommand{\natexlab}[1]{#1}
\providecommand{\url}[1]{{#1}}
\providecommand{\urlprefix}{URL }
\providecommand{\doi}[1]{\url{https://doi.org/#1}}
\providecommand{\eprint}[2][]{\url{#2}}
 \bibcommenthead

\bibitem[{Ahmed et~al.(2026)Ahmed, Galke~Poech, and R{\"o}ttger}]{staledoc2026}
Ahmed MS, Galke~Poech L, R{\"o}ttger R (2026) Stale-document poisoning: When
  outdated retrieval overrides correct model answers. arXiv preprint
  {arXiv}:2609.31342

\bibitem[{Chen et~al.(2026)Chen, Chen, Lin, and Vong}]{chen2026judgeknow}
Chen J, Chen W, Lin Z, et~al (2026) A judge should know what changed: Construct
  validity for {LLM}-as-a-judge evaluation. arXiv preprint {arXiv}:2608.24419

\bibitem[{Chen et~al.(2021)Chen, Wang, and Wang}]{chen2021timeqa}
Chen W, Wang X, Wang WY (2021) A dataset for answering time-sensitive
  questions. In: Proceedings of the Neural Information Processing Systems Track
  on Datasets and Benchmarks, \urlprefix\url{https://arxiv.org/abs/2108.06314}

\bibitem[{Cymbler et~al.(2026)Cymbler, Guez, and Fabre}]{cymbler2026temporal}
Cymbler R, Guez D, Fabre L (2026) Temporal misgrounding in legal {RAG}: A
  versioned-corpus benchmark for {French} tax law. In: ICML 2026 Workshop on AI
  for Law (AI4Law), {arXiv}:2608.09393

\bibitem[{Gardner et~al.(2020)Gardner, Artzi, Basmov
  et~al.}]{gardner2020contrast}
Gardner M, Artzi Y, Basmov V, et~al (2020) Evaluating models' local decision
  boundaries via contrast sets. In: Findings of the Association for
  Computational Linguistics: EMNLP 2020, pp 1307--1323,
  \doi{10.18653/v1/2020.findings-emnlp.117}

\bibitem[{Gema et~al.(2025)Gema, Leang, Hong, Devoto, Mancino, Saxena, He,
  Zhao, Du, Ghasemi~Madani, Barale, McHardy, Harris, Kaddour, Van~Krieken, and
  Minervini}]{gema2025mmlu}
Gema AP, Leang JOJ, Hong G, et~al (2025) Are we done with {MMLU}? In:
  Proceedings of the 2025 Conference of the Nations of the Americas Chapter of
  the Association for Computational Linguistics: Human Language Technologies
  (Volume 1: Long Papers), pp 5069--5096, \doi{10.18653/v1/2025.naacl-long.262}

\bibitem[{Governatori and Rotolo(2010)}]{governatori2010changing}
Governatori G, Rotolo A (2010) Changing legal systems: legal abrogations and
  annulments in defeasible logic. Logic Journal of the IGPL 18(1):157--194.
  \doi{10.1093/jigpal/jzp075}

\bibitem[{Huwiler et~al.(2025)Huwiler, Stockinger, and
  F{\"u}rst}]{huwiler2025versionrag}
Huwiler D, Stockinger K, F{\"u}rst J (2025) {VersionRAG}: Version-aware
  retrieval-augmented generation for evolving documents. arXiv preprint
  {arXiv}:2510.08109

\bibitem[{Kapetanovic et~al.(2026)Kapetanovic, Altwlkany, Mercep, Duricic, and
  Lacic}]{kapetanovic2026anchoring}
Kapetanovic A, Altwlkany K, Mercep A, et~al (2026) Anchoring bias in
  {LLM}-as-a-judge systems: Prior scores compromise evaluation independence.
  In: Proceedings of the 35th ACM International Conference on Information and
  Knowledge Management (CIKM), to appear; arXiv:2608.25869

\bibitem[{Kim et~al.(2026)Kim, Brendel, and Reizinger}]{kim2026hallmark}
Kim K, Brendel W, Reizinger P (2026) {HALLMARK}: How to diagnose {LLM} citation
  verifiers and when to deploy them. arXiv preprint {arXiv}:2607.18360

\bibitem[{Lee et~al.(2026)Lee, Hwang, Kang, Lee, Chae, and
  Jung}]{lee2026reference}
Lee D, Hwang Y, Kang T, et~al (2026) Judging against the reference: Uncovering
  knowledge-driven failures in {LLM}-judges on {QA} evaluation. arXiv preprint
  {arXiv}:2601.07506

\bibitem[{Liu and Gao(2026)}]{liu2026trusting}
Liu Z, Gao S (2026) Trusting the tool: Asymmetric deference of {LLM}
  vulnerability auditors to static-analysis evidence. arXiv preprint
  {arXiv}:2610.32876. Submitted to IEEE Access

\bibitem[{McCoy et~al.(2019)McCoy, Pavlick, and Linzen}]{mccoy2019right}
McCoy RT, Pavlick E, Linzen T (2019) Right for the wrong reasons: Diagnosing
  syntactic heuristics in natural language inference. In: Proceedings of the
  57th Annual Meeting of the Association for Computational Linguistics, pp
  3428--3448, \doi{10.18653/v1/P19-1334}

\bibitem[{Molfetta et~al.(2026)Molfetta, Cocchieri, Ragazzi, Bartolini,
  Patella, and Moro}]{molfetta2026sycophants}
Molfetta L, Cocchieri A, Ragazzi L, et~al (2026) Sycophants in the courtroom:
  Are {LLMs} fragile to juridical authority and evolving legal standards? In:
  Proceedings of the 64th Annual Meeting of the Association for Computational
  Linguistics (Volume 1: Long Papers), pp 10865--10886,
  \doi{10.18653/v1/2026.acl-long.497},
  \urlprefix\url{https://aclanthology.org/2026.acl-long.497/}

\bibitem[{Nam et~al.(2026)Nam, Kim, Song, Yoon, Lee, and
  Lee}]{nam2026timelyrag}
Nam Y, Kim J, Song H, et~al (2026) {TimelyRAG}: Semantic-temporal hybrid
  retrieval for time-critical question answering in overlapping-evolving
  documents. In: Findings of the Association for Computational Linguistics:
  EMNLP 2026, pp 8414--8430, \doi{10.18653/2026.findings-emnlp.398},
  \urlprefix\url{https://aclanthology.org/2026.findings-emnlp.398/}

\bibitem[{Northcutt et~al.(2021)Northcutt, Athalye, and
  Mueller}]{northcutt2021pervasive}
Northcutt CG, Athalye A, Mueller J (2021) Pervasive label errors in test sets
  destabilize machine learning benchmarks. In: Proceedings of the Neural
  Information Processing Systems Track on Datasets and Benchmarks,
  {arXiv}:2103.14749

\bibitem[{Ouyang et~al.(2025)Ouyang, Pan, Cheng, Yan, Luo, Lin, and
  Liu}]{ouyang2025hoh}
Ouyang J, Pan T, Cheng M, et~al (2025) {HoH}: A dynamic benchmark for
  evaluating the impact of outdated information on retrieval-augmented
  generation. In: Proceedings of the 63rd Annual Meeting of the Association for
  Computational Linguistics (Volume 1: Long Papers), pp 6036--6063,
  \doi{10.18653/v1/2025.acl-long.301},
  \urlprefix\url{https://aclanthology.org/2025.acl-long.301/}

\bibitem[{Pall(2026)}]{pall2026scoring}
Pall RS (2026) Scoring-rule sensitivity in {LLM} evaluation: Evidence from
  {Indian} financial regulatory text. arXiv preprint {arXiv}:2604.19298

\bibitem[{Palmirani and Brighi(2002)}]{palmirani2002norma}
Palmirani M, Brighi R (2002) Norma-system: A legal document system for managing
  consolidated acts. In: Database and Expert Systems Applications (DEXA 2002),
  Lecture Notes in Computer Science, vol 2453. Springer, pp 310--320,
  \doi{10.1007/3-540-46146-9_31}

\bibitem[{Palmirani and Brighi(2006)}]{palmirani2006time}
Palmirani M, Brighi R (2006) Time model for managing the dynamic of normative
  system. In: Electronic Government (EGOV 2006), Lecture Notes in Computer
  Science, vol 4084. Springer, pp 207--218, \doi{10.1007/11823100_19}

\bibitem[{Palmirani et~al.(2011)Palmirani, Governatori, Rotolo, Tabet, Boley,
  and Paschke}]{palmirani2011legalruleml}
Palmirani M, Governatori G, Rotolo A, et~al (2011) {LegalRuleML}: {XML}-based
  rules and norms. In: Rule-Based Modeling and Computing on the Semantic Web
  (RuleML 2011 America), Lecture Notes in Computer Science, vol 7018. Springer,
  pp 298--312, \doi{10.1007/978-3-642-24908-2_30}

\bibitem[{Prior et~al.(2025)Prior, Hof, Wais, and
  Grabmair}]{prior2025consolidation}
Prior M, Hof A, Wais N, et~al (2025) Risks and limits of automatic
  consolidation of statutes. In: Proceedings of the Natural Legal Language
  Processing Workshop 2025, pp 396--407, \doi{10.18653/v1/2025.nllp-1.29}

\bibitem[{Prior et~al.(2026)Prior, Schultz, and Grabmair}]{prior2026oldfriend}
Prior M, Schultz A, Grabmair M (2026) Asking for an old friend: Diagnosing and
  mitigating temporal failure modes in {LLM}-based statutory question
  answering. In: Proceedings of the International Conference on Artificial
  Intelligence and Law (ICAIL), {arXiv}:2605.23497

\bibitem[{Qian et~al.(2026)Qian, Wang, Chen, Xu, Wang, Yu, Lin, Zhang, You, and
  Wei}]{qian2026relevant}
Qian P, Wang S, Chen Y, et~al (2026) Relevant is not warranted: Evidence-force
  calibration for cited {RAG}. In: Findings of the Association for
  Computational Linguistics: EMNLP 2026, pp 33590--33603,
  \doi{10.18653/2026.findings-emnlp.1577},
  \urlprefix\url{https://aclanthology.org/2026.findings-emnlp.1577/}

\bibitem[{{RFC Editor}(2026)}]{rfcindex2026}
{RFC Editor} (2026) {RFC} index ({XML}).
  \url{https://www.rfc-editor.org/rfc-index.xml}, retrieved 29 October 2026;
  SHA-256 recorded in the repository

\bibitem[{Sharma et~al.(2024)Sharma, Tong, Korbak, Duvenaud, Askell, Bowman,
  Cheng, Durmus, Hatfield-Dodds, Johnston, Kravec, Maxwell, McCandlish,
  Ndousse, Rausch, Schiefer, Yan, Zhang, and Perez}]{sharma2023sycophancy}
Sharma M, Tong M, Korbak T, et~al (2024) Towards understanding sycophancy in
  language models. In: The Twelfth International Conference on Learning
  Representations (ICLR 2024)

\bibitem[{Tan et~al.(2025)Tan, Zhuang, Montgomery, Tang, Cuadron, Wang, Popa,
  and Stoica}]{tan2025judgebench}
Tan S, Zhuang S, Montgomery K, et~al (2025) {JudgeBench}: A benchmark for
  evaluating {LLM}-based judges. In: International Conference on Learning
  Representations (ICLR), \urlprefix\url{https://arxiv.org/abs/2410.12784}

\bibitem[{Vu et~al.(2024)Vu, Iyyer, Wang, Constant, Wei, Wei, Tar, Sung, Zhou,
  Le, and Luong}]{vu2024freshllms}
Vu T, Iyyer M, Wang X, et~al (2024) {FreshLLMs}: Refreshing large language
  models with search engine augmentation. In: Findings of the Association for
  Computational Linguistics: ACL 2024, pp 13697--13720,
  \doi{10.18653/v1/2024.findings-acl.813},
  \urlprefix\url{https://aclanthology.org/2024.findings-acl.813/}

\bibitem[{Wu et~al.(2024)Wu, Wu, and Zou}]{wu2024clasheval}
Wu K, Wu E, Zou J (2024) {ClashEval}: Quantifying the tug-of-war between an
  {LLM}'s internal prior and external evidence. In: Advances in Neural
  Information Processing Systems 37 (NeurIPS 2024), pp 33402--33422,
  \doi{10.52202/079017-1053}, \urlprefix\url{https://arxiv.org/abs/2404.10198}

\bibitem[{Xie et~al.(2024)Xie, Zhang, Chen, Lou, and Su}]{xie2024chameleon}
Xie J, Zhang K, Chen J, et~al (2024) Adaptive chameleon or stubborn sloth:
  Revealing the behavior of large language models in knowledge conflicts. In:
  International Conference on Learning Representations (ICLR),
  {arXiv}:2305.13300

\bibitem[{Xu et~al.(2025)Xu, Bansal, Ming, Yavuz, and
  Joty}]{xu2025contextualjudgebench}
Xu A, Bansal S, Ming Y, et~al (2025) Does context matter?
  {ContextualJudgeBench} for evaluating {LLM}-based judges in contextual
  settings. In: Proceedings of the 63rd Annual Meeting of the Association for
  Computational Linguistics (Volume 1: Long Papers), pp 9541--9564,
  \doi{10.18653/v1/2025.acl-long.470},
  \urlprefix\url{https://aclanthology.org/2025.acl-long.470/}

\bibitem[{Xu et~al.(2024)Xu, Qi, Guo, Wang, Wang, Zhang, and
  Xu}]{xu2024knowledge}
Xu R, Qi Z, Guo Z, et~al (2024) Knowledge conflicts for {LLMs}: A survey. In:
  Proceedings of the 2024 Conference on Empirical Methods in Natural Language
  Processing, pp 8541--8565, \doi{10.18653/v1/2024.emnlp-main.486},
  \urlprefix\url{https://aclanthology.org/2024.emnlp-main.486/}

\bibitem[{Yang et~al.(2026)Yang, Song, Kim, Song, Park, and
  Jo}]{yang2026overreliance}
Yang H, Song W, Kim T, et~al (2026) Agents' overreliance on unreliable tools.
  arXiv preprint {arXiv}:2609.05587

\bibitem[{Ye et~al.(2025)Ye, Tawfik, Goodell, Kotha, Buyyounouski, and
  Bayati}]{ye2025stewardship}
Ye J, Tawfik D, Goodell AJ, et~al (2025) Scalable stewardship of an
  {LLM}-assisted clinical benchmark with physician oversight. arXiv preprint
  {arXiv}:2512.19691

\bibitem[{Zhang and Choi(2021)}]{zhang2021situatedqa}
Zhang MJQ, Choi E (2021) {SituatedQA}: Incorporating extra-linguistic contexts
  into {QA}. In: Proceedings of the 2021 Conference on Empirical Methods in
  Natural Language Processing, pp 7371--7387,
  \doi{10.18653/v1/2021.emnlp-main.586},
  \urlprefix\url{https://aclanthology.org/2021.emnlp-main.586/}

\bibitem[{Zheng et~al.(2023)Zheng, Chiang, Sheng, Zhuang, Wu, Zhuang, Lin, Li,
  Li, Xing, Zhang, Gonzalez, and Stoica}]{zheng2023judging}
Zheng L, Chiang WL, Sheng Y, et~al (2023) Judging {LLM}-as-a-judge with
  {MT-Bench} and {Chatbot Arena}. In: Advances in Neural Information Processing
  Systems (Datasets and Benchmarks Track), \doi{10.52202/075280-2020},
  {arXiv}:2306.05685

\bibitem[{Zhu et~al.(2025)Zhu, Liao, Chen, Wang, Guan, Wang, and
  Wang}]{zhu2025evolvebench}
Zhu Z, Liao Y, Chen Z, et~al (2025) {EvolveBench}: A comprehensive benchmark
  for assessing temporal awareness in {LLMs} on evolving knowledge. In:
  Proceedings of the 63rd Annual Meeting of the Association for Computational
  Linguistics (Volume 1: Long Papers), pp 16173--16188,
  \doi{10.18653/v1/2025.acl-long.788}

\end{thebibliography}

\ifdefined\withsupplement\clearpage\setcounter{section}{0}\setcounter{subsection}{0}\setcounter{table}{0}\setcounter{figure}{0}
\renewcommand{\thesection}{S\arabic{section}}
\renewcommand{\thetable}{S\arabic{table}}
\renewcommand{\thefigure}{S\arabic{figure}}
\renewcommand{\theHsection}{S\arabic{section}}
\renewcommand{\theHtable}{S\arabic{table}}
\renewcommand{\theHfigure}{S\arabic{figure}}
\section*{Supplementary material: the case study in full}
\noindent This supplement holds the full report of the descriptive case study on Indian rooftop-solar regulation that
the paper summarises in its final appendix: the reference data and their human audit, the conditions, damage and
scoring, the planned analyses and their results, the readings fixed in advance, the prompts, the per-clause results,
the annotation agreement, the deviations from the analysis plan and the data statement. The case study's answer key
failed the quality checks planned for it, so nothing here supports a finding of the main paper; it is reported for
transparency, with its planned analyses and readings kept as they were fixed. The sections are adapted, with cross-references to the main paper added, from the
study's own report, written before the RFC study, and use its terms (reviewer for judge, registry for record, clause
for document). The study's conditions are the rungs \rung{R0}--\rung{R5} of a reference ladder, defined in \S S1.2.2 and
Table~\ref{tab:conditions}; a ``validity'' reason there includes wrong jurisdiction and not applicable. Q1 and Q2 name
the two primary questions (\S\ref{sec:res-q1} and \S\ref{sec:res-q2}), and E1 to E5 the secondary analyses listed
under ``Primary family'' in \S\ref{sec:setup}.

\section{Case study: reference data, conditions and planned analyses}
\label{app:study1}

\subsection{Reference data and its audit}
\label{app:study1-data}
\paragraph{Corpus.}
The corpus holds \NCorpusDocs{} authentic regulatory instruments on rooftop solar from \NCorpusJurisdictions{}
jurisdictions in India: central schemes and the regulations and orders of state commissions and distribution
companies. Superseded instruments stay in the corpus, because their continued availability is the hazard studied.
Every file is hashed and listed with its issuer, dates and source URL.

\paragraph{Amendment registry.}
The reviewer's amendment record is a registry of \NLinksRegistry{} clause-level links, each from an old clause to the
clause that changes it, with a relation (supersedes, partially supersedes, amends), an effective date and a quote from
the amending instrument. It was extracted from the documents by an AI assistant, without access to the gold graph (the
adjudicated replacement links used for scoring). It is the source of the \rung{R3} quotes, the \rung{R4} and \rung{R4n} status lines and the \rung{R5}
lookup. The registry as corrected by adjudication (below) is the one every run uses.

\paragraph{Cases and reference answers.}
A case is a synthetic application: an electricity bill, an installer proposal and a distribution-company portal record,
with a case date and a jurisdiction. Cases were generated backward from rule functions whose parameters are quoted from
the clause text, so no label was typed by hand. Each case records the rule function that labelled it; most cases list a
single applicable rule. The applications contain no personal data. The candidate test set had \NCandidateCases{} cases
on the replaced clauses with enough fixed answers per condition; the dev split shares no replaced clause with it and was
used only for engineering and the pilot.

\subsubsection{Human audit of the reference data}
\label{sec:annotation}
The registry, the answer key and the case labels were produced with AI help, so two authors checked them blind before
the test set was sealed, splitting the work and labelling a seeded shared subset (\NSharedCases{} cases, \NSharedLinks{}
links) in common. They labelled each case's decision, blocking stage and deciding clauses from the case file and the
source documents; each checked link's relation, effective date and wording; and a sample of \NControlsAnnotated{} matched
correct answers. A third author, who led the pipeline that produced the first-pass labels, adjudicated every
disagreement from the annotators' sheets and the source documents, without the answer keys. The planned procedure
had the annotators meet first; to save time the meeting was dropped and all disagreements went to the adjudicator
(Appendix~\ref{app:deviations}). The annotators are not legal professionals.

On the shared cases the annotators agreed on the decision in \AgreeCasesDecision{} of cases (kappa
\KappaCasesDecision{}) and on the blocking stage in \AgreeCasesBlockingStage{} (kappa \KappaCasesBlockingStage{}), but on
the exact set of deciding clauses in only \AgreeCasesDecidingClauses{} (kappa \KappaCasesDecidingClauses{}). On shared
links they agreed on the relation in \AgreeLinksRelation{} (kappa \KappaLinksRelation{}). Part of the clause-level
disagreement is notation (\cfg{cl-4.2} against \cfg{cl-4(2)}); part is that several clause sets support the same outcome.

\paragraph{The answer key.}
Against the adjudicated labels, the AI-built key reached a different outcome on \NDroppedDecision{} of the
\NCandidateCases{} cases: \NDroppedDecisionOnly{} on the decision and \NDroppedBlockingStage{} on the blocking stage of a
rejection. In the disagreements we inspected, the readers based their decision on a second clause that the case's rule
function does not evaluate, often in a newer instrument. In one such case, the key approves an application because it passes the
transformer-capacity rule its rule function checks; the readers reject it under an eligibility clause of a later
regulation. These cases are wrong at the base, and they are dropped; a case is never relabelled, because its faults
were planted on the key's answer. On \NClauseDisputed{} of the remaining cases, the outcome agrees and only the deciding
clauses differ. Those cases are kept and listed, because a planted fault and its catch are defined on the replaced
clause alone. Applied as originally planned, which dropped clause-only disputes too, the rules would have kept
\NStrictKept{} cases.

\paragraph{The correct answers.}
Of the \NControlsChecked{} matched correct answers checked, \NControlsCorrect{} (\PctControlsCorrect) were judged correct.
Most of the rest were judged incomplete or citing a wrong clause, the same single-rule cause: they cite the clause the
rule function evaluated and omit another that also decides the case. A reviewer that flags such an answer may be right,
so these items bias false alarms upwards. The planned gate (\ControlsGatePct\% correct) would have stopped the seal; it
was replaced by reporting, and by sensitivity set S1, which removes every case whose checked correct answer was judged
anything but correct (\S\ref{sec:setup}).

\paragraph{The registry.}
The \NLinksAnnotated{} checked links are the \NLinksUsed{} that test items rely on and a seeded \LinkSamplePct\% random
sample of \NLinksSample{} of the other links. In the random sample, \NLinksSampleConfirmed{} were confirmed,
\NLinksSampleRelChanged{} had a different relation, and \NLinksSampleNotALink{} were judged to be no change at all
(\CINotALinkSample{}, Wilson 95\% interval). Among the test-item links, \NLinksTestPairNotALink{} was no change and
\NLinksTestPairRelChanged{} had a different relation. Rejected links were removed and corrected dates applied before
the run. These figures describe links the registry asserts. Nothing in the audit looked for links the registry misses,
so we do not estimate that rate. The full table, with both annotators' raw answers beside the adjudicated one, is
released (\cfg{40\_data/link\_audit\_v1}); links outside the check keep the extractor's provisional relation and are not
used as gold.

\paragraph{The scorer.}
One annotator labelled \ScorerCheckN{} dev replies by hand under the written scoring rule. The automatic scorer agreed on
\ScorerCheckAgreeN{} (\ScorerCheckPctObserved), below the planned \ScorerCheckPct\% gate. In each of the
\NScorerDisagree{} disagreements, the reply's own clause ids and reason codes give the scorer's label under the written
rule, so the scorer was not changed. Most disagreements lie on the boundary between a wrong clause and a wrong reason
(\NScorerClauseReason{}), or between a right catch and a wrong reason (\NScorerCatchReason{}). We read this as evidence
that the right-reason rule is hard to apply by hand, not as a defect of the scorer, and report every result under the
two looser rules as well (E5).

\paragraph{The sealed test set.}
The sealed set has \NTestCases{} cases on \NClauses{} distinct replaced clauses from \NStates{} jurisdictions
(Table~\ref{tab:items}). They come from \InstrNIntactGemma{} amending instruments (\NSharedInstrument{} clauses in one
state from a single one), so clauses are not independent. The registry links \NRegistryFlagged{} of the \NClauses{} replaced clauses; on the others the intact
registry itself reads \cfg{in\_force}, so they are naturally missing links.

\begin{table}[t]
\centering
\small
\begin{tabular}{lrrr}
\toprule
Condition & Faults & Matched correct answers & Clauses \\
\midrule
S-both (old and current clause shown) & \NFaultsSBoth & \NControlsSBoth & \NClausesSBoth \\
S-swap (old clause in place of current) & \NFaultsSSwap & \NControlsSSwap & \NClausesSSwap \\
C (clean correct answers) & -- & \NCleanC & -- \\
\bottomrule
\end{tabular}
\caption{Fixed answers per condition in the sealed test set. Clauses: distinct replaced clauses. Jurisdictions of the
replaced clauses: DL \NClausesStateDL, GJ \NClausesStateGJ, central \NClausesStateIN, KA \NClausesStateKA,
TN \NClausesStateTN, UP \NClausesStateUP.}
\label{tab:items}
\end{table}

\subsection{Conditions, damage and scoring}
\label{app:study1-conditions}
\label{sec:method}

\subsubsection{Fixed answers}
\label{sec:probe}
We measure review on fixed answers, so the result does not depend on what a generator happens to write. For each case
and condition we build a matched pair. The \emph{fault} cites the replaced clause in place of its successor. When the old
clause changes the approve-or-reject outcome, the fault also carries the decision that clause implies; otherwise it has
the right decision on an invalid basis. The \emph{matched correct answer} (control) has the key's decision and cites only
clauses its own evidence shows. Both come from one neutral template: the decision, the cited clause identifiers, one
sentence that quotes the basis clause, and the blocking stage if any. A pair therefore differs only in its basis and,
where the outcome changes, in the decision. Condition C adds one clean correct answer per case. The reviewer sees the
case, the evidence and one answer, and is never told which answers are faults.

The evidence differs by condition. In \emph{S-both} the old clause and its successor are both shown; in \emph{S-swap}
the old clause replaces the current one, so the reviewer sees only outdated text.

\subsubsection{Conditions}
\label{sec:conditions}
Every language-model condition uses the same validity prompt, which asks whether each cited clause is in force on the
application date, issued for the case's jurisdiction and applicable to its facts. Conditions differ only in what the
prompt contains (Table~\ref{tab:conditions}; prompts in Appendix~\ref{app:prompts}). \rung{R3}, \rung{R4} and \rung{R4n}
are built on \rung{R2}: the \rung{R3} block holds the amending instrument's verbatim substitution text for each cited
clause the registry links, with no status, date or target; the \rung{R4} block holds one line per cited clause giving
the registry's entries, jurisdiction and status at the case date. \Rfourp{} gives the same line at \rung{R1}, without
the current clause; it shows what the status line does when the text cannot confirm it. \rung{R5} uses no language
model: it raises a ``not in force'' finding on a cited clause exactly when the registry marks that clause superseded or
partially superseded at the case date (an ``amends'' link alone does not), and it raises no other kind of finding. Its
catches are therefore the registry's own links, and its false alarms are the registry's own wrong links.

\paragraph{The no-record arm.}
In the registry, a clause with no outgoing link has status \cfg{in\_force}. The registry cannot distinguish a clause
that was never amended from one whose amendment it failed to record, so this value states more than the record holds.
\rung{R4n} replaces the status value \cfg{in\_force} by \cfg{no\_\allowbreak amendment\_\allowbreak recorded} and changes nothing else: the
prompt sent at \rung{R4n} equals the \rung{R4} prompt with that one substitution, which a unit test checks on an
intact and a damaged registry. Superseded, partially superseded and not-yet-in-force values are unchanged. The substitution applies to every clause
the registry does not link, including current clauses on correct answers, so \rung{R4n} can change false alarms as well as
catches.

\begin{table}[t]
\centering
\scriptsize
\setlength{\tabcolsep}{3pt}
\begin{tabular}{p{1.45cm}p{2.85cm}p{2.2cm}p{2.95cm}p{1.5cm}}
\toprule
Condition & Reviewer sees & Reviewer does not see & Compared with; confound & Role \\
\midrule
\rung{R0} & old clause (S-swap), source metadata removed & current clause, dates, issuer & -- & ladder \\
\rung{R1} & old clause with issuer, dates, title & current clause & \rung{R0}: metadata only & ladder \\
\rung{R2} & old and current clause (S-both) & any amendment record & \rung{R1}: adds the current clause & ladder; Q2 baseline \\
\rung{R3} & \rung{R2} + amending instrument's words & status, dates, target & \rung{R2} & ladder \\
\rung{R4} & \rung{R2} + registry status line & -- & intact: told the answer for linked clauses & reference \\
\Rfourp{} & \rung{R1} + registry status line & current clause & \rung{R1}: status line without the text & ladder \\
\rung{R4}-damage & \rung{R2} + status line of a damaged registry & which links are wrong & \rung{R5}-damage (Q1); \rung{R2} (Q2) & primary \\
\rung{R4n}, \rung{R4n}-damage & as \rung{R4}, ``no amendment recorded'' for unlinked clauses & -- & \rung{R4} on the same items and seeds & primary \\
\rung{R5}, \rung{R5}-damage & no LLM: ``not in force'' on every cited clause the registry marks superseded & -- & catches exactly the linked clauses, by construction & reference \\
\bottomrule
\end{tabular}
\caption{Conditions. ``Reference'' rows are not findings: \rung{R5} recovers the registry's own links, and \rung{R4} at the
intact registry is told, for every linked clause, that it is not in force. Faults were planted on replaced clauses of the
gold graph; the registry links most but not all of them (\S\ref{app:study1-data}). The E2 variants of \rung{R0} to
\rung{R2} add a ``can't confirm'' reason.}
\label{tab:conditions}
\end{table}

\paragraph{Can't confirm.}
For E2, variants of \rung{R0}, \rung{R1} and \rung{R2} add the reason \cfg{currency\_unconfirmed}: the evidence shown does
not establish whether the clause is in force at the case date. Such a finding is neither a catch nor a false alarm.

\subsubsection{Damaged registries}
\label{sec:damage}
To vary the record's reliability in a controlled way we damage the registry in three ways: \DamageDropTwentyFive\% or
\DamageDropFifty\% of its links dropped, or \DamageSpuriousTen\% spurious links added. Each damage has \NDamageSeeds{}
seeds. A drop permutes all of the registry's links with the seed and removes the first quarter or half, so at one seed
the 50\% set contains the 25\% set. A spurious link is added from a seeded choice of dated clauses that have no
outgoing link, to another clause of the same jurisdiction, as a ``superseded by'' link effective from the clause's own
date; the number added is 10\% of the registry's links. A damaged registry is used in each way the intact one is: as the lookup
(\rung{R5}-damage) and as the status line (\rung{R4}-damage, \rung{R4n}-damage). A dropped link makes the status line
report a replaced clause as \cfg{in\_force} (\rung{R4}) or as having no recorded amendment (\rung{R4n}), while the evidence
still shows both versions; a spurious link makes it report a current clause as replaced. In both cases the text in view
contradicts the line. The damage is random and synthetic; it is not an estimate of how real registries fail.

\subsubsection{Scoring}
\label{sec:scoring}
The reviewer returns a verdict and findings; each finding names a clause and gives one reason from a closed list (not in
force, wrong jurisdiction, not applicable, missing rule, content or arithmetic, other). The list is the same in every
prompt and names no clause.

\paragraph{Catch, false alarm, confusion counts.}
A fault is caught when a finding names the replaced clause with the reason ``not in force''; an identifier of a
retrieval chunk counts as the clauses it contains. On a correct answer, any finding is a false alarm. Each condition
therefore yields four counts: caught and missed faults, flagged and passed correct answers. A finding on the right clause
for another reason is not a catch; we count it as a \emph{misnamed} alarm (E4).

\paragraph{Discrimination.}
For one replaced clause, discrimination is its catch rate on faults minus the false-alarm rate on their matched correct
answers; it is averaged over clauses, and for a damaged registry first over seeds within a clause. It is the headline
measure, because a reviewer that flags every answer catches every fault and is of no use.

\paragraph{Deference and NoRec.}
Both use the same items: every (S-both fault, drop rate, seed) where the intact registry catches the fault and that
damaged registry does not. Deference is, item by item, the reviewer's catch with no registry (\rung{R2})
minus its catch with the damaged status line (\rung{R4}-damage). NoRec is the item's catch at \rung{R4n}-damage minus its
catch at \rung{R4}-damage on the same seed. Both are averaged within clause, then over clauses.

\paragraph{Three scoring rules (E5).}
Under \emph{any finding}, a fault is caught if the reply has any finding; under \emph{clause named}, if a finding names
the replaced clause for any reason; the right-reason rule above is the strictest. On correct answers we report any finding
(our false-alarm rule) and a finding with a validity reason only.

\paragraph{Post hoc measures.} These were added after the results were seen and were not pre-specified.
Because the lookup can only raise ``not in force'' while the reviewer can raise any reason, we also report Q1 with false
alarms counted only when the finding gives a validity reason (not in force, wrong jurisdiction, not applicable), the
reason codes behind the false alarms, the catches the reviewer gains and loses against the lookup on the same item and
seed, and Q1 with the amending instrument rather than the clause as the unit (the \NClauses{} clauses come from fewer
amending instruments). These are descriptive and outside the family.

\paragraph{Invalid replies.}
A reply that is not valid JSON, is cut off at the output cap after one retry, or has no verdict is invalid. It counts as a
miss on a fault and as a false alarm on a correct answer, so an invalid reply never improves a condition.

\subsection{Setup and planned analyses}
\label{app:study1-planned}
\label{sec:setup}

\paragraph{Models.}
The panel has \NPanelModels{} open models: \texttt{llama3.1:8b} (weak), \texttt{qwen2.5:7b}, and the \NCapable{} capable
models \texttt{qwen2.5:14b} and \texttt{gemma-4-26b-a4b-it}. The first three run through Ollama with 4-bit quantised
weights fixed in advance by digest (\texttt{46e0c10c039e}, \texttt{845dbda0ea48} and \texttt{7cdf5a0187d5}; full digests
in the analysis plan), checked on every machine: \texttt{qwen2.5:14b} on a cloud notebook with two NVIDIA T4 GPUs, the
other two on an Apple-silicon laptop. Gemma, open-weight, runs through a hosted API, with the version recorded per call.
A closed hosted model, \texttt{gemini-3.5-flash-lite}, is a secondary model on \rung{R0}, \rung{R2} and \rung{R4} only.
Each model ran on one machine. Models added after the plan was fixed are exploratory and reported separately
(\S\ref{sec:res-explore}).

\paragraph{Settings.}
Decoding is greedy (temperature \Temperature), with run seed \RunSeed{} and an output cap of \OutputCap{} tokens. Every
reply is cached, and every block records the model digest, the prompt hash per item, the code hash and the hash of the
sealed test file. Every model sees the same items.

\paragraph{Run order and cutoff.}
Blocks run in a fixed priority order: for the capable models \rung{R2}, \rung{R4}, \rung{R4n}, \rung{R5} and
\rung{R5}-damage, then \rung{R4}-damage and \rung{R4n}-damage seed by seed, so that a cut run still leaves the two arms
paired on the same seeds; then the secondary conditions; then the other models. The analysis cutoff was fixed as a
wall-clock time before any test result was read. A block not finished by then is reported as not run, and a contrast whose blocks are
incomplete is reported as not run rather than estimated from part of a block; every planned block finished
(Appendix~\ref{app:deviations}).

\paragraph{Unit of inference and tests.}
The unit of inference is the distinct replaced clause. Rates are averaged within each clause and then over clauses, with
pooled counts beside each rate; false alarms on condition C are averaged by rule group, because C plants nothing. A
contrast is the mean per-clause difference, paired by clause within one model, tested with a two-sided sign-flip test over
clauses (exact for few clauses, otherwise \NPermutations{} random sign vectors with a fixed seed). Intervals are
\CILevel\% bootstrap intervals over clauses, printed only with at least \MinClauses{} clauses. A cell with fewer than
\SparseItems{} items or \SparseClauses{} clauses is printed as counts.

\paragraph{Primary family.}
\NPrimaryTestsVFive{} tests, Holm-corrected at $\alpha=\Alpha$: for each capable model, \rung{R4} minus \rung{R5}
discrimination at the intact, \DamageDropFifty\%-drop and \DamageSpuriousTen\%-spurious registries (Q1), deference (Q2) and
NoRec. A contrast on fewer than \MinClauses{} clauses enters the family with $p=1$. The \DamageDropTwentyFive\% drop is
reported beside Q1, uncorrected. Everything else is secondary and uncorrected or descriptive: the ladder contrasts (E1),
can't confirm (E2), spurious-link false alarms (E3$'$), misnamed alarms (E4), the scoring rules (E5), the false-alarm cost
of \rung{R4n}, and descriptive splits by amendment wording (explicit or implicit), by whether the fault changes the
outcome, and by cost per call (tokens and seconds).

\paragraph{Sensitivity sets.}
Every primary result is also computed on two subsets, with the same tests and outside the family. S1 removes
every case whose checked correct answer was judged anything but correct. S2 removes every case the originally
planned rule would have dropped, that is, every case whose deciding clauses differ from the key; it has fewer than
\MinClauses{} clauses and is descriptive. Only a sample of correct answers was checked, so S1 removes the correct
answers known to be flawed, not every flawed one; a result that keeps its sign under S1 does not depend on those.

\paragraph{Damage seeds.}
The capable models run \rung{R4}-damage and \rung{R4n}-damage on all \NDamageSeedsCapable{} seeds; the other models run
seed 0. \rung{R5}-damage is deterministic and runs every seed.

\paragraph{Analysis plan.}
The design, outcomes, analysis and the sentence reporting each outcome were written into a dated, version-controlled
analysis plan before any model saw a test item, and the test set was sealed with a fingerprint recorded in the
repository. The plan was not filed with a public registry; the commit times of the plan and its amendments are listed in the release's provenance file. Two dated
amendments were made before the seal. The first, written after the dev pilot, made the
damaged-registry comparisons primary. The second, written after the human audit and before any model saw a test item,
changed which cases annotation keeps, replaced the correct-answer and scorer gates by reporting, and added \rung{R4n},
NoRec, S1 and S2 (Appendix~\ref{app:deviations}).

\subsection{Planned results}
\label{app:study1-results}

Seven reviewers ran on the same items: the five planned models, and two added after the analysis plan was fixed,
\texttt{qwen2.5:32b} (damage seeds 0--2) and Claude Opus~4.5 through a hosted proxy (damage seed 0; temperature and
seed set by the proxy). Every planned block finished for the planned models.
Table~\ref{tab:main} gives the confusion counts of every main arm on the same \NFaultsSBoth{} S-both faults and
\NControlsSBoth{} matched correct answers. A repeat of \texttt{qwen2.5:14b}'s \rung{R2}, \rung{R4} and \rung{R4n}
blocks on the same machine with an empty cache returned identical findings on \NRepeatSameFindings{} of
\NRepeatItems{} items; we did not test whether replies are identical across machines or for the hosted models. Invalid replies were rare (Gemma: \InvalidKNGemmaRFourDropFifty{} at
\rung{R4} with \DamageDropFifty\% of links dropped; \texttt{qwen2.5:14b}: none).

\IfFileExists{generated/tab_main.tex}{\begin{table*}[t]\centering\scriptsize
\caption{Main results on S-both faults and their matched correct answers. TP: right-reason catches of the planted superseded clause; FN: faults not caught; FP: matched correct answers with any finding; TN: matched correct answers passed. Recall, specificity, precision and balanced accuracy (\%) pool items, which share clauses, so they are descriptive. FA$_C$: clean correct answers flagged. Disc.: discrimination (catch minus matched false alarm), clause-averaged, with a 95\% bootstrap interval over clauses (points). R5 recovers the registry's own links by construction and R4 (intact) is given the answer: reference rows.}
\label{tab:main}
\resizebox{\textwidth}{!}{%
\begin{tabular}{lllrrrrrrrrrll}
\toprule
Model & Arm & Reviewer sees & TP & FN & FP & TN & Rec. & Spec. & Prec. & Bal. & FA$_C$ & Disc. [95\% CI] & Cl. \\
\midrule
Qwen 14B & R2 & both versions & 13 & 29 & 18 & 24 & 31 & 57 & 42 & 44 & 12/42 & $-$7 [$-$38, +20] & 14 \\
Qwen 14B & R4 & status line & 32 & 10 & 10 & 32 & 76 & 76 & 76 & 76 & 11/42 & +55 [+26, +79] & 14 \\
Qwen 14B & R4n & no-record wording & 31 & 11 & 13 & 29 & 74 & 69 & 70 & 71 & 12/42 & +47 [+17, +73] & 14 \\
Lookup & R5 & lookup (ref.) & 31 & 11 & 5 & 37 & 74 & 88 & 86 & 81 & 5/42 & +67 [+43, +88] & 14 \\
Qwen 14B & R4-drop50 & status, drop 50\% & 88 & 122 & 54 & 156 & 42 & 74 & 62 & 58 & 55/210 & +21 [$-$6, +48] & 14 \\
Qwen 14B & R4n-drop50 & no-record, drop 50\% & 95 & 115 & 58 & 152 & 45 & 72 & 62 & 59 & 60/210 & +23 [$-$6, +54] & 14 \\
Lookup & R5-drop50 & lookup, drop 50\% & 88 & 122 & 8 & 202 & 42 & 96 & 92 & 69 & 8/210 & +42 [+24, +60] & 14 \\
Qwen 14B & R4-spurious10 & status, spur. 10\% & 160 & 50 & 55 & 155 & 76 & 74 & 74 & 75 & 61/210 & +53 [+25, +76] & 14 \\
Lookup & R5-spurious10 & lookup, spur. 10\% & 155 & 55 & 34 & 176 & 74 & 84 & 82 & 79 & 34/210 & +62 [+37, +84] & 14 \\
Gemma 4 26B & R2 & both versions & 9 & 33 & 20 & 22 & 21 & 52 & 31 & 37 & 17/42 & $-$20 [$-$46, +4] & 14 \\
Gemma 4 26B & R4 & status line & 25 & 17 & 20 & 22 & 60 & 52 & 56 & 56 & 18/42 & +21 [$-$18, +57] & 14 \\
Gemma 4 26B & R4n & no-record wording & 26 & 16 & 21 & 21 & 62 & 50 & 55 & 56 & 15/42 & +18 [$-$20, +54] & 14 \\
Gemma 4 26B & R4-drop50 & status, drop 50\% & 89 & 121 & 100 & 110 & 42 & 52 & 47 & 47 & 90/210 & +4 [$-$30, +35] & 14 \\
Gemma 4 26B & R4n-drop50 & no-record, drop 50\% & 96 & 114 & 105 & 105 & 46 & 50 & 48 & 48 & 72/210 & +2 [$-$31, +31] & 14 \\
Gemma 4 26B & R4-spurious10 & status, spur. 10\% & 125 & 85 & 103 & 107 & 60 & 51 & 55 & 55 & 94/210 & +20 [$-$20, +56] & 14 \\
\bottomrule
\end{tabular}}
\end{table*}
}{}

\subsubsection{Q1: does the reviewer add anything over the record?}
\label{sec:res-q1}

\begin{table*}[t]
\centering
\footnotesize
\setlength{\tabcolsep}{3pt}
\begin{tabular}{l ccc ccc}
\toprule
& \multicolumn{3}{c}{\texttt{qwen2.5:14b}} & \multicolumn{3}{c}{\texttt{gemma-4-26b}} \\
\cmidrule(lr){2-4}\cmidrule(lr){5-7}
Registry & \rung{R4} & \rung{R5} & \rung{R4}$-$\rung{R5} & \rung{R4} & \rung{R5} & \rung{R4}$-$\rung{R5} \\
\midrule
Intact & \DiscRFourIntactQwenFourteen & \DiscRFiveIntactQwenFourteen & \DiffNetIntactQwenFourteen & \DiscRFourIntactGemma & \DiscRFiveIntactGemma & \DiffNetIntactGemma \\
Drop \DamageDropTwentyFive\% & \DiscRFourDropTwentyFiveQwenFourteen & \DiscRFiveDropTwentyFiveQwenFourteen & \DiffNetDropTwentyFiveQwenFourteen & \DiscRFourDropTwentyFiveGemma & \DiscRFiveDropTwentyFiveGemma & \DiffNetDropTwentyFiveGemma \\
Drop \DamageDropFifty\% & \DiscRFourDropFiftyQwenFourteen & \DiscRFiveDropFiftyQwenFourteen & \DiffNetDropFiftyQwenFourteen & \DiscRFourDropFiftyGemma & \DiscRFiveDropFiftyGemma & \DiffNetDropFiftyGemma \\
Spurious \DamageSpuriousTen\% & \DiscRFourSpuriousTenQwenFourteen & \DiscRFiveSpuriousTenQwenFourteen & \DiffNetSpuriousTenQwenFourteen & \DiscRFourSpuriousTenGemma & \DiscRFiveSpuriousTenGemma & \DiffNetSpuriousTenGemma \\
\bottomrule
\end{tabular}
\caption{Discrimination in points of the LLM reviewer with the status line (\rung{R4}) and of the registry lookup
alone (\rung{R5}), by registry state, over \NClauses{} clauses. The difference is the Q1 contrast; intervals and
Holm $p$ are in Table~\ref{tab:provenance}. The \DamageDropTwentyFive\% row is outside the primary family.}
\label{tab:q1}
\end{table*}

\IfFileExists{generated/tab_provenance.tex}{\begin{table*}[t]\centering\small
\caption{Primary contrasts (capable models; Holm over the family of ten). Per-clause differences in points, mean over clauses, 95\% bootstrap interval over clauses, two-sided sign-flip test. S1: without the cases whose checked correct answer was judged flawed; S2: under the original strict annotation rule (descriptive when it leaves fewer than ten clauses).}
\label{tab:provenance}
\resizebox{\textwidth}{!}{%
\begin{tabular}{llrrlrrr}
\toprule
Model & Contrast & Clauses & Diff. & 95\% CI & $p$ (Holm) & S1 & S2 \\
\midrule
Gemma 4 26B & Q1, intact: R4 $-$ R5 discrimination & 14 & $-$45 & [$-$69, $-$24] & 0.043 & $-$47 & $-$50 \\
Gemma 4 26B & Q1, 50\% dropped & 14 & $-$38 & [$-$65, $-$11] & 0.178 & $-$38 & $-$50 \\
Gemma 4 26B & Q1, 10\% spurious & 14 & $-$42 & [$-$66, $-$22] & 0.043 & $-$45 & $-$45 \\
Gemma 4 26B & Q2: R2 $-$ R4-damage catch, dropped links & 9 & +1 & -- & 1.000 & +4 & -- \\
Gemma 4 26B & R4n $-$ R4 catch, dropped links & 9 & +9 & -- & 1.000 & +20 & -- \\
Qwen 14B & Q1, intact: R4 $-$ R5 discrimination & 14 & $-$11 & [$-$33, +9] & 1.000 & $-$12 & +17 \\
Qwen 14B & Q1, 50\% dropped & 14 & $-$21 & [$-$38, $-$6] & 0.206 & $-$17 & +2 \\
Qwen 14B & Q1, 10\% spurious & 14 & $-$10 & [$-$30, +10] & 1.000 & $-$11 & +17 \\
Qwen 14B & Q2: R2 $-$ R4-damage catch, dropped links & 9 & +34 & -- & 1.000 & +33 & -- \\
Qwen 14B & R4n $-$ R4 catch, dropped links & 9 & +9 & -- & 1.000 & +9 & -- \\
\bottomrule
\end{tabular}}
\end{table*}
}{}

Under the planned rule, the reviewer given the record had a lower point estimate of discrimination than a lookup
in the same record at every registry state, for both capable models (Table~\ref{tab:q1}). For Gemma the difference
was \DiffNetIntactGemma{} points with the registry intact (CI \CINetIntactGemma, Holm $p=$ \PHolmNetIntactGemma) and
\DiffNetSpuriousTenGemma{} with spurious links (CI \CINetSpuriousTenGemma, Holm $p=$ \PHolmNetSpuriousTenGemma),
both significant after Holm, and \DiffNetDropFiftyGemma{} with half the links dropped (Holm $p=$
\PHolmNetDropFiftyGemma). For \texttt{qwen2.5:14b} it was \DiffNetIntactQwenFourteen{} (Holm $p=$
\PHolmNetIntactQwenFourteen), \DiffNetDropFiftyQwenFourteen{} (CI \CINetDropFiftyQwenFourteen, Holm $p=$
\PHolmNetDropFiftyQwenFourteen) and \DiffNetSpuriousTenQwenFourteen{} (Holm $p=$ \PHolmNetSpuriousTenQwenFourteen),
none significant. The reading fixed in advance for both models is that the reviewer did not discriminate better than a
lookup in the registry it was given, even when the registry was damaged; the reading fixed in advance for Gemma's intact
result is that adding the reviewer to a correct registry made review worse than the registry alone.

That last reading should be taken narrowly, for three reasons we found after the run (post hoc, \S\ref{sec:scoring}).
First, the gap is made of findings the lookup cannot raise. The planned false-alarm rule counts any finding on a
correct answer, but the lookup only ever says ``not in force''. Most of Gemma's findings on matched correct answers gave
``content or arithmetic'' or ``missing rule'' as the reason, not a validity reason. Counting only validity-reason
findings as false alarms, the difference is \DiffNetVIntactGemma{} points for Gemma with the registry intact (CI
\CINetVIntactGemma) and \DiffNetVDropFiftyGemma{} with half the links dropped; for \texttt{qwen2.5:14b} it is
\DiffNetVIntactQwenFourteen{} and \DiffNetVDropFiftyQwenFourteen{} (CI \CINetVDropFiftyQwenFourteen). On the
question the lookup answers, whether a cited clause is in force, the reviewer did about as well as the lookup, except perhaps \texttt{qwen2.5:14b} with half the links dropped, whose
bootstrap interval ends just below zero (the printed 0 is rounded) although its sign-flip test does not reject. The
reviewers' other findings may be right: the audit found that many correct answers omit a clause that also decides the case
(\S\ref{sec:annotation}), which this design cannot score. Second, the \NClauses{} clauses come from
\InstrNIntactGemma{} amending instruments; with the instrument as the unit, Gemma's intact and spurious differences
have uncorrected sign-flip $p=$ \InstrPIntactGemma{} and \InstrPSpuriousTenGemma, on fewer units than the planned
minimum of \MinClauses. Third, both lose significance under S1 (\S\ref{sec:res-sens}). We therefore report Gemma's
Q1 result as planned, but do not read it as evidence that a reviewer degrades a correct record.

The expectation fixed in advance for Q1, that the gap would narrow or reverse as links were dropped because the reviewer
can still read both versions, was not met. Table~\ref{tab:main} shows why. With \DamageDropFifty\% of links
dropped, the reviewer's right-reason catches on S-both faults nearly equal the lookup's for both models. Item by item
and seed by seed, Gemma caught \RecGainedDropFiftyGemma{} faults the damaged lookup missed and missed
\RecLostDropFiftyGemma{} it caught; \texttt{qwen2.5:14b} gained \RecGainedDropFiftyQwenFourteen{} and lost
\RecLostDropFiftyQwenFourteen. The reviewer recovered some dropped links from the text, and lost about as many
catches elsewhere, so on net it did not repair the record. Its false alarms on matched correct answers stayed well
above the lookup's.

Two properties of the design limit this result. The lookup's catches are the registry's own links: it misses
the faults on clauses the registry does not link and flags whatever a wrong link marks, so its discrimination is a
property of the record, not a benchmark of reasoning (\S\ref{sec:method}). And the result holds for these models,
these \NClauses{} clauses and fixed answers with a single planted fault.

\subsubsection{Q2: does the reviewer follow a wrong ``in force'' line?}
\label{sec:res-q2}
The deference contrast could be computed on \NDroppedClausesQwenFourteen{} clauses: those whose planted link the
intact registry catches and some drop seed removes. That is below our planned minimum of \MinClauses{}, so by
rule it entered the family with $p=1$ and supports no confirmatory claim; we report it descriptively and, as
planned, as no evidence of deference rather than evidence of its absence. On those faults
\texttt{qwen2.5:14b} caught \OwnCatchDroppedQwenFourteen{} with no registry (\rung{R2}) and
\CatchDroppedQwenFourteen{} when the damaged status line said the replaced clause was in force (deference
\DiffDeferQwenFourteen{} points). Gemma caught \OwnCatchDroppedGemma{} and \CatchDroppedGemma{} (deference
\DiffDeferGemma). The direction for \texttt{qwen2.5:14b} is the one the deference account predicts and matches the
effect of false assurances that \citet{liu2026trusting} report for code auditors; Gemma's catches on these items
hardly changed.

\paragraph{The no-record wording (R4n).} On the same faults, stating that no amendment is recorded instead of that
the clause is in force raised catches from \CatchDroppedRFourQwenFourteen{} to \CatchDroppedRFournQwenFourteen{}
for \texttt{qwen2.5:14b} and from \CatchDroppedRFourGemma{} to \CatchDroppedRFournGemma{} for Gemma (NoRec
\DiffNoRecQwenFourteen{} and \DiffNoRecGemma{} points, \NNoRecClausesQwenFourteen{} clauses, below the minimum,
no confirmatory claim). False alarms on matched correct answers rose a little: with half the links dropped, from
\FAMQwenFourteenRFourDropFifty{} to \FAMQwenFourteenRFourNDropFifty{} for \texttt{qwen2.5:14b} and from
\FAMGemmaRFourDropFifty{} to \FAMGemmaRFourNDropFifty{} for Gemma; with the registry intact, from
\FAMQwenFourteenRFour{} to \FAMQwenFourteenRFourN{} and from \FAMGemmaRFour{} to \FAMGemmaRFourN. As planned, we
report no evidence that the wording matters, not evidence that it does not; the point estimates are small and
positive (for \texttt{qwen2.5:14b} a fraction of what the dropped links cost; for Gemma, which lost almost nothing to
them, more than that), and come with a small rise in false alarms. The set is
small by construction: deference and NoRec need a fault the intact registry catches, and the registry links only
\NRegistryFlagged{} of the \NClauses{} clauses.

\paragraph{Spurious links (E3$'$).} On correct answers citing a clause that a spurious link marks as replaced
(\KNSpuriousMarkedQwenFourteen{} items on \NSpuriousClausesQwenFourteen{} clauses), \texttt{qwen2.5:14b} raised a false alarm on
\FASpuriousMarkedQwenFourteen{} with the status line against \OwnFASpuriousMarkedQwenFourteen{} with no registry;
Gemma on \FASpuriousMarkedGemma{} against \OwnFASpuriousMarkedGemma{}. A wrong link marking a current clause as
replaced was followed into a false alarm more often than not. The counts are small and no test was planned. Unlike the code auditors of \citet{liu2026trusting}, whose accuracy was never significantly lowered
when a tool raised false alarms, these reviewers followed a wrong record in both directions; on three clauses this is a
hypothesis for confirmatory work, not a finding.

\subsubsection{How much the scoring rule matters}
\label{app:res-e5}
\begin{table*}[t]
\centering
\footnotesize
\setlength{\tabcolsep}{3pt}
\begin{tabular}{ll ccc ccc}
\toprule
& & \multicolumn{3}{c}{\texttt{qwen2.5:14b}} & \multicolumn{3}{c}{\texttt{gemma-4-26b}} \\
\cmidrule(lr){3-5}\cmidrule(lr){6-8}
& Rule & \rung{R0} & \rung{R2} & \rung{R4} & \rung{R0} & \rung{R2} & \rung{R4} \\
\midrule
Catch & any finding & \ScAnyCatchQwenFourteenRZero & \ScAnyCatchQwenFourteenRTwo & \ScAnyCatchQwenFourteenRFour & \ScAnyCatchGemmaRZero & \ScAnyCatchGemmaRTwo & \ScAnyCatchGemmaRFour \\
& clause named & \ScClauseCatchQwenFourteenRZero & \ScClauseCatchQwenFourteenRTwo & \ScClauseCatchQwenFourteenRFour & \ScClauseCatchGemmaRZero & \ScClauseCatchGemmaRTwo & \ScClauseCatchGemmaRFour \\
& right reason & \CatchQwenFourteenRZero & \CatchQwenFourteenRTwo & \CatchQwenFourteenRFour & \CatchGemmaRZero & \CatchGemmaRTwo & \CatchGemmaRFour \\
\midrule
False alarm & any finding & \ScAnyFAQwenFourteenRZero & \ScAnyFAQwenFourteenRTwo & \ScAnyFAQwenFourteenRFour & \ScAnyFAGemmaRZero & \ScAnyFAGemmaRTwo & \ScAnyFAGemmaRFour \\
& validity reason & \ScReasonFAQwenFourteenRZero & \ScReasonFAQwenFourteenRTwo & \ScReasonFAQwenFourteenRFour & \ScReasonFAGemmaRZero & \ScReasonFAGemmaRTwo & \ScReasonFAGemmaRFour \\
\bottomrule
\end{tabular}
\caption{The same replies under three scoring rules, on faults (catch) and matched correct answers (false alarm).
Clause-averaged. The right-reason row is the catch reported everywhere else. At \rung{R0} the reviewer sees only
the old clause, so a right-reason catch is not expected there.}
\label{tab:scoring}
\end{table*}
The scoring rule changed the apparent skill of the same replies by tens of points (Table~\ref{tab:scoring}).
With both versions in view (\rung{R2}), any-finding scoring credited \texttt{qwen2.5:14b} with
\ScAnyCatchQwenFourteenRTwo{} catches and Gemma with \ScAnyCatchGemmaRTwo, against \CatchQwenFourteenRTwo{} and
\CatchGemmaRTwo{} when the finding had to name the replaced clause for the reason that it is no longer in force.
Naming the clause hardly mattered: almost every finding on a fault named it, and the gap is the reason. Under the
any-finding rule the reviewers also flagged a large share of matched correct answers, so a catch rate reported alone
would overstate them. The reading fixed in advance applies: scoring by flags alone would have credited these reviewers
with many catches made for the wrong reason.

The right-reason rule is itself hard to apply. In the human check of the scorer on \ScorerCheckN{} dev replies the
annotator and the code agreed on \ScorerCheckAgreeN{} (\ScorerCheckPctObserved), below the pre-specified
\ScorerCheckPct\% gate; \NScorerClauseReason{} of the \NScorerDisagree{} disagreements lie on the boundary between
a wrong clause and a wrong reason. Amendment 2 kept the scorer and reports the failed gate
(\S\ref{sec:annotation}); the any-finding and clause-named rows of Table~\ref{tab:scoring} show every result under
the looser rules, and the post hoc Q1 analysis in \S\ref{sec:res-q1} applies the validity-reason rule to false alarms.

\subsubsection{The reference ladder (E1) and the other models}
\label{sec:res-e1}
\begin{figure}[t]
\centering
\includegraphics[width=0.7\linewidth]{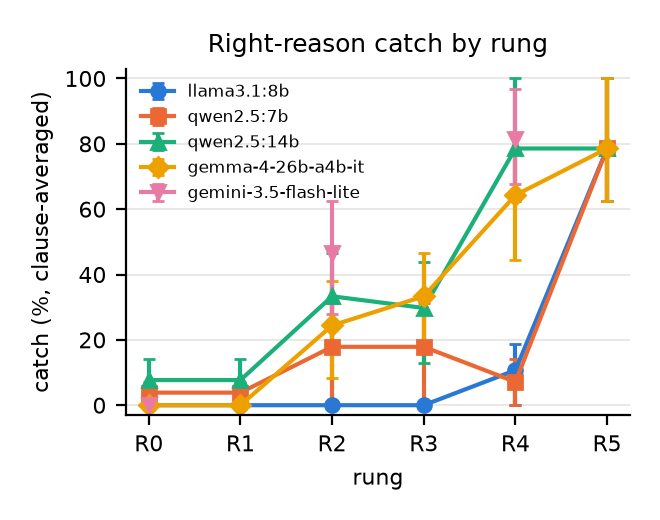}
\caption{Right-reason catch by rung, one line per model, clause-averaged, with \CILevel\% bootstrap intervals
over clauses. \rung{R5} is the registry lookup and is the same for every model.}
\label{fig:ladder}
\end{figure}

\begin{table*}[t]
\centering
\footnotesize
\setlength{\tabcolsep}{2pt}
\begin{tabular}{l ccc ccc ccc ccc}
\toprule
& \multicolumn{3}{c}{\texttt{llama3.1:8b}} & \multicolumn{3}{c}{\texttt{qwen2.5:7b}}
& \multicolumn{3}{c}{\texttt{qwen2.5:14b}} & \multicolumn{3}{c}{\texttt{gemma-4-26b}} \\
\cmidrule(lr){2-4}\cmidrule(lr){5-7}\cmidrule(lr){8-10}\cmidrule(lr){11-13}
Rung & Catch & FA-m & FA-C & Catch & FA-m & FA-C & Catch & FA-m & FA-C & Catch & FA-m & FA-C \\
\midrule
\rung{R0} & \CatchLlamaRZero & \FAMLlamaRZero & \FACLlamaRZero & \CatchQwenSevenRZero & \FAMQwenSevenRZero & \FACQwenSevenRZero & \CatchQwenFourteenRZero & \FAMQwenFourteenRZero & \FACQwenFourteenRZero & \CatchGemmaRZero & \FAMGemmaRZero & \FACGemmaRZero \\
\rung{R1} & \CatchLlamaROne & \FAMLlamaROne & \FACLlamaROne & \CatchQwenSevenROne & \FAMQwenSevenROne & \FACQwenSevenROne & \CatchQwenFourteenROne & \FAMQwenFourteenROne & \FACQwenFourteenROne & \CatchGemmaROne & \FAMGemmaROne & \FACGemmaROne \\
\rung{R2} & \CatchLlamaRTwo & \FAMLlamaRTwo & \FACLlamaRTwo & \CatchQwenSevenRTwo & \FAMQwenSevenRTwo & \FACQwenSevenRTwo & \CatchQwenFourteenRTwo & \FAMQwenFourteenRTwo & \FACQwenFourteenRTwo & \CatchGemmaRTwo & \FAMGemmaRTwo & \FACGemmaRTwo \\
\rung{R3} & \CatchLlamaRThree & \FAMLlamaRThree & \FACLlamaRThree & \CatchQwenSevenRThree & \FAMQwenSevenRThree & \FACQwenSevenRThree & \CatchQwenFourteenRThree & \FAMQwenFourteenRThree & \FACQwenFourteenRThree & \CatchGemmaRThree & \FAMGemmaRThree & \FACGemmaRThree \\
\rung{R4} & \CatchLlamaRFour & \FAMLlamaRFour & \FACLlamaRFour & \CatchQwenSevenRFour & \FAMQwenSevenRFour & \FACQwenSevenRFour & \CatchQwenFourteenRFour & \FAMQwenFourteenRFour & \FACQwenFourteenRFour & \CatchGemmaRFour & \FAMGemmaRFour & \FACGemmaRFour \\
\Rfourp & \CatchLlamaRFourP & \FAMLlamaRFourP & \FACLlamaRFourP & \CatchQwenSevenRFourP & \FAMQwenSevenRFourP & \FACQwenSevenRFourP & \CatchQwenFourteenRFourP & \FAMQwenFourteenRFourP & \FACQwenFourteenRFourP & \CatchGemmaRFourP & \FAMGemmaRFourP & \FACGemmaRFourP \\
\rung{R5} & \CatchLlamaRFive & \FAMLlamaRFive & \FACLlamaRFive & \CatchQwenSevenRFive & \FAMQwenSevenRFive & \FACQwenSevenRFive & \CatchQwenFourteenRFive & \FAMQwenFourteenRFive & \FACQwenFourteenRFive & \CatchGemmaRFive & \FAMGemmaRFive & \FACGemmaRFive \\
\bottomrule
\end{tabular}
\caption{Right-reason catch and false alarms by rung (clause-averaged; C false alarms averaged by rule group).
FA-m: matched correct answers of the same blocks (S-swap pairs for \rung{R0}, \rung{R1} and \Rfourp, S-both
otherwise; \rung{R5} pools both). FA-C: clean correct answers. Table~\ref{tab:main} gives S-both counts only, so its
false alarms can differ from FA-m here.}
\label{tab:ladder}
\end{table*}

The ladder is the reference that places Q1 and Q2. For both capable models the status line lifted right-reason
catches over seeing both versions: \rung{R4}$-$\rung{R2} was \DiffRFourMinusRTwoQwenFourteen{} points for
\texttt{qwen2.5:14b} (CI \CIRFourMinusRTwoQwenFourteen) and \DiffRFourMinusRTwoGemma{} for Gemma (CI
\CIRFourMinusRTwoGemma), while \rung{R2}$-$\rung{R0} was \DiffRTwoMinusRZeroQwenFourteen{} and
\DiffRTwoMinusRZeroGemma{} (on the \ClausesRTwoMinusRZeroGemma{} clauses where both rungs ran, so they differ from the
clause averages of Table~\ref{tab:ladder}), not significant by the planned sign-flip test (Gemma's bootstrap interval,
\CIRTwoMinusRZeroGemma, excludes zero; \rung{R0} runs on S-swap items, where a right-reason catch is near zero by
design). In the words fixed in advance, seeing both versions is not enough, and
\rung{R4} lifts these capable models. The amending instrument's quote (\rung{R3}) added little over \rung{R2}
(\DiffRThreeMinusRTwoQwenFourteen{} and \DiffRThreeMinusRTwoGemma{} points). Without a record the reviewers did
not discriminate at all: \rung{R2} discrimination was \DiscRTwoQwenFourteen{} points for \texttt{qwen2.5:14b} and
\DiscRTwoGemma{} for Gemma, because they flagged matched correct answers as often as, or more often than, they caught faults for
the right reason.

The weak model gained little: \texttt{llama3.1:8b} caught \CatchLlamaRTwo{} at \rung{R2} and \CatchLlamaRFour{} at
\rung{R4} (\DiffRFourMinusRTwoLlama{} points, interval \CIRFourMinusRTwoLlama). That is just outside the
\InterpPoints-point margin of the reading fixed in advance for this case (reading 12), so that reading does not apply;
we report the small gain as a limit of the claim. \texttt{gemini-3.5-flash-lite}, run at \rung{R0}, \rung{R2} and \rung{R4} only, caught
\CatchFlashRTwo{} at \rung{R2} and \CatchFlashRFour{} at \rung{R4}, with false alarms on matched correct answers
of \FAMFlashRFour{} at \rung{R4}; seeing both versions did not close the gap for it either. Given the ``can't confirm'' option (E2), neither capable model used it on a fault at \rung{R2}
(\CantConfirmKNQwenFourteenRTwoCC{} and \CantConfirmKNGemmaRTwoCC): the option was available and unused. Of the
\rung{R2} findings that flagged the replaced clause, \MisnamedKNQwenFourteenRTwo{} (\texttt{qwen2.5:14b}) and
\MisnamedKNGemmaRTwo{} (Gemma) never gave ``not in force'' as the reason (E4).

\subsubsection{The added models}
\label{sec:res-explore}
\IfFileExists{generated/tab_exploratory.tex}{\begin{table}[t]\centering\small
\caption{The two models added after the plan: a size comparison within one family, and Claude Opus 4.5. Rows marked ``no'' were added after the analysis plan was fixed and are exploratory; the planned sizes are shown beside them. The 32B model ran damage seeds 0--2 (the planned capable models ran 0--4). S-both right-reason catches, false alarms on matched correct answers (any finding) and any-finding catches, pooled k/n; discrimination clause-averaged (points), and the Q1 difference with a 95\% bootstrap interval over clauses. No test.}
\label{tab:exploratory}
\footnotesize\setlength{\tabcolsep}{3pt}
\begin{tabular}{lllrrrl}
\toprule
Model & Planned & Arm & Catch & FA & Any & Disc. / diff. \\
\midrule
Qwen 7B & yes & R2 & 5/42 & 7/42 & 14/42 & $-$1 \\
Qwen 7B & yes & R4 & 2/42 & 2/42 & 3/42 & +0 \\
Qwen 7B & yes & R4-drop50 & 2/42 & 2/42 & 3/42 & +0 \\
Qwen 7B & yes & R4$-$R5, intact & \multicolumn{3}{c}{} & $-$67 [$-$88, $-$43] \\
Qwen 7B & yes & R4$-$R5, drop50 & \multicolumn{3}{c}{} & $-$42 [$-$60, $-$24] \\
\addlinespace
Qwen 14B & yes & R2 & 13/42 & 18/42 & 26/42 & $-$7 \\
Qwen 14B & yes & R4 & 32/42 & 10/42 & 32/42 & +55 \\
Qwen 14B & yes & R4n & 31/42 & 13/42 & 32/42 & +47 \\
Qwen 14B & yes & R4-drop50 & 88/210 & 54/210 & 126/210 & +21 \\
Qwen 14B & yes & R4$-$R5, intact & \multicolumn{3}{c}{} & $-$11 [$-$33, +9] \\
Qwen 14B & yes & R4$-$R5, drop50 & \multicolumn{3}{c}{} & $-$21 [$-$38, $-$6] \\
\addlinespace
Qwen 32B & no & R2 & 4/42 & 25/42 & 35/42 & $-$48 \\
Qwen 32B & no & R4 & 29/42 & 23/42 & 38/42 & +18 \\
Qwen 32B & no & R4n & 26/42 & 24/42 & 38/42 & +11 \\
Qwen 32B & no & R4-drop50 & 46/126 & 69/126 & 104/126 & $-$15 \\
Qwen 32B & no & R4$-$R5, intact & \multicolumn{3}{c}{} & $-$48 [$-$68, $-$24] \\
Qwen 32B & no & R4$-$R5, drop50 & \multicolumn{3}{c}{} & $-$57 [$-$76, $-$37] \\
\addlinespace
Claude Opus 4.5 & no & R2 & 34/42 & 40/42 & 42/42 & $-$11 \\
Claude Opus 4.5 & no & R4 & 35/42 & 37/42 & 42/42 & $-$3 \\
Claude Opus 4.5 & no & R4n & 35/42 & 42/42 & 42/42 & $-$14 \\
Claude Opus 4.5 & no & R4-drop50 & 29/42 & 37/42 & 42/42 & $-$21 \\
Claude Opus 4.5 & no & R4$-$R5, intact & \multicolumn{3}{c}{} & $-$70 [$-$89, $-$48] \\
Claude Opus 4.5 & no & R4$-$R5, drop50 & \multicolumn{3}{c}{} & $-$63 [$-$85, $-$43] \\
\bottomrule
\end{tabular}
\end{table}
}{}
The two models added after the analysis plan was fixed are outside every planned test; Table~\ref{tab:exploratory}
and the descriptive-results table of the paper's case-study appendix report them on the main arms. Claude Opus~4.5
behaves differently from the open models: it is the only reviewer whose catches survive the absence of a record, and
the one with the clearest net gain against the damaged lookup (gaining many and losing none). Its catches on dropped links still
fell under the damaged status line, as in the main study, and its false alarms on this key's correct answers are
the highest; against a flawed key they cannot be interpreted. A model from the same developer also helped build this
benchmark (see the paper's statement on the use of AI assistants); its items were not tuned on any model's replies,
but readers should weigh that relation. It ran one damage seed, through a
proxy that sets temperature and seed itself, so its numbers carry more uncertainty than the planned models'. In the one family we ran at three sizes, the largest model did not review better than the 14B one: \texttt{qwen2.5:32b}, run with
the pre-specified prompts on the same items, flagged matched correct answers more often than
\texttt{qwen2.5:14b} at every arm, and about twice as often once given the record, so its discrimination was lower, and its gap to the lookup wider, with the record
intact and with half the links dropped. Without a record it named the replaced clause on most faults but rarely for
the reason that it is no longer in force. On the dropped-link faults its catches fell to none under the damaged
status line, and the no-record wording did not restore them. One model, one family and the same \NClauses{} clauses
cannot separate size from other differences between checkpoints; we report it because a reader will ask whether a
larger model removes the pattern, and here it did not.

\subsubsection{Sensitivity to the annotation audit}
\label{sec:res-sens}
Removing the cases whose checked correct answer annotators judged flawed (S1; only a sample was checked) left the sign
of every primary contrast unchanged (Table~\ref{tab:provenance}). Gemma's two significant Q1 differences became
\SOneDiffNetIntactGemma{} and \SOneDiffNetSpuriousTenGemma{} points with Holm $p=$ \SOnePHolmNetIntactGemma{} and
\SOnePHolmNetSpuriousTenGemma, no longer below \Alpha{} on the smaller set. Under the reading fixed in advance, the
direction of the Q1 result does not depend on the correct answers known to be flawed; its significance for Gemma does
depend on keeping them. Under the original strict rule (S2) only \NStrictKept{} cases remain, on fewer than \MinClauses{} clauses; that
is descriptive only, and there \texttt{qwen2.5:14b}'s Q1 difference turns positive (\STwoDiffNetIntactQwenFourteen{}
points intact), Gemma's stays negative (\STwoDiffNetIntactGemma).

\section{Case study: readings fixed in advance}
\label{app:interpretations}
Each rule was planned with the sentence we report when it applies; where a reading names a place of report (``the
abstract''), it refers to the case study's own report. Readings 1--5 are from amendment 1, readings
6--8 from amendment 2, and the rest from the original analysis plan, where they now apply to the reference ladder.
\begin{enumerate}
  \item \rung{R4}$-$\rung{R5} not positive after Holm at any registry state, for a capable model: ``For this
  model, the reviewer did not discriminate better than a lookup in the registry it was given, even when the
  registry was damaged.''
  \item \rung{R4}$-$\rung{R5} positive after Holm at a damaged state but not with the registry intact: ``The
  reviewer added value over the registry only once the registry was damaged, from [the first such state].''
  \item \rung{R4}$-$\rung{R5} negative after Holm with the registry intact: ``Adding the reviewer to a correct
  registry made review worse than the registry alone.''
  \item Deference positive after Holm: ``A wrong `in force' line removed catches this reviewer makes without
  any registry: it followed the registry over the text.'' Not significant: ``We found no evidence of
  deference''; never ``the reviewer does not defer''.
  \item Any-finding catch above right-reason catch by more than \InterpPoints{} points for a capable model at
  \rung{R0} to \rung{R2}: ``Scoring by flags alone would have credited this reviewer with [the gap] points of
  catches it made for the wrong reason.''
  \item NoRec positive after Holm and the false-alarm cost small: ``Telling the reviewer that no amendment is
  recorded, instead of that the clause is in force, recovers catches the damaged record loses.'' Not significant:
  ``We found no evidence that the wording matters''; never that it does not.
  \item NoRec positive but the false-alarm cost larger than the gain: ``The wording trades catches for false alarms''
  (both numbers reported).
  \item A primary result that keeps its sign under S1: ``It does not depend on the flawed correct answers.'' One that
  changes sign: ``It depends on them.''
  \item Fewer than \MinClauses{} distinct replaced clauses in the sealed test: no primary claim. ``With
  this many distinct clauses the test cannot support a population claim; we report per-clause results
  descriptively.''
  \item \rung{R2} at least as high as \rung{R4} for the capable models: ``The bottleneck is retrieval,
  not representation: given both versions, reviewers catch the replaced rule as well as when told its
  status.''
  \item \rung{R4}$-$\rung{R2} not significant (uncorrected, secondary since amendment 1): no ladder claim. ``Telling the reviewer the
  registry status did not reliably raise catches for these models.''
  \item The weak model at \rung{R4} within \InterpPoints{} points of its \rung{R2}: ``Below a capability
  threshold, the information does not help.'' Reported as a limit of the claim.
  \item The secondary model's \rung{R2} within \InterpPoints{} points of its \rung{R4}: ``For a stronger
  model, seeing both versions is enough; the ladder result is limited to the local panel.'' Reported in
  the abstract.
  \item ``Can't confirm'' never used by a model: ``The option was available and unused.''
\end{enumerate}
``Seeing both versions is not enough'' means that \rung{R2}$-$\rung{R0} is not significant and that
\rung{R2}'s catch is below half of \rung{R4}'s.

\section{Case study: prompts}
\label{app:prompts}
The exact prompt of every rung, rendered for one dev item by \texttt{render\_prereg\_prompts.py}, is kept with
the analysis plan (\texttt{PREREGISTRATION\_v3\_PROMPTS.md}; the plan's working file names say ``preregistration'',
but it was not publicly registered) and released with the code; we do not retype it here.
\rung{R4n} is \rung{R4}'s prompt with one change: in the registry lookup block, a status value
\texttt{in\_force} reads \texttt{no\_amendment\_recorded}. A unit test checks that the two prompts differ in that word
only, with the registry intact and damaged.

\section{Case study: per-clause results}
\label{app:perclause}
\IfFileExists{generated/tab_perclause.tex}{\begin{table*}[t]\centering\scriptsize
\caption{Right-reason catches per replaced clause (k/n faults; R5 also counts S-swap faults), capable models. The clause is the unit of every interval and test in the paper.}
\label{tab:perclause}
\resizebox{\textwidth}{!}{%
\begin{tabular}{lcccccccc}
\toprule
Replaced clause & \multicolumn{4}{c}{Qwen 14B} & \multicolumn{4}{c}{Gemma 4 26B} \\
 & R2 & R4 & R4n & R5 & R2 & R4 & R4n & R5 \\
\midrule
\texttt{DL-C3-004\#cl-3.8} & 2/4 & 4/4 & 4/4 & 8/8 & 0/4 & 4/4 & 4/4 & 8/8 \\
\texttt{DL-C3-004\#cl-9.2b} & 2/2 & 2/2 & 2/2 & 4/4 & 2/2 & 2/2 & 2/2 & 4/4 \\
\texttt{DL-C3-006\#cl-3.2} & 2/3 & 3/3 & 3/3 & 6/6 & 3/3 & 3/3 & 3/3 & 6/6 \\
\texttt{DL-C3-009\#cl-3.6} & 2/4 & 4/4 & 4/4 & 8/8 & 1/4 & 4/4 & 4/4 & 8/8 \\
\texttt{GJ-C3-001\#cl-13.5-check} & 0/4 & 4/4 & 4/4 & 8/8 & 0/4 & 2/4 & 3/4 & 8/8 \\
\texttt{GJ-C3-001\#cl-13.5-class} & 3/3 & 3/3 & 3/3 & 6/6 & 2/3 & 3/3 & 3/3 & 6/6 \\
\texttt{GJ-C3-001\#cl-5-proviso} & 0/3 & 0/3 & 0/3 & 6/6 & 0/3 & 0/3 & 0/3 & 6/6 \\
\texttt{GJ-C3-003\#cl-6.2} & 0/2 & 2/2 & 2/2 & 4/4 & 0/2 & 2/2 & 2/2 & 4/4 \\
\texttt{IN-C1-011\#cl-annex-phase2-claims} & 0/2 & 2/2 & 2/2 & 4/4 & 0/2 & 2/2 & 2/2 & 4/4 \\
\texttt{KA-C5-001\#cl-15} & 0/4 & 4/4 & 3/4 & 0/8 & 0/4 & 0/4 & 0/4 & 0/8 \\
\texttt{KA-C5-001\#cl-26} & 0/4 & 0/4 & 0/4 & 0/4 & 0/4 & 0/4 & 0/4 & 0/4 \\
\texttt{TN-C5-001\#cl-2} & 0/3 & 0/3 & 0/3 & 0/6 & 0/3 & 0/3 & 0/3 & 0/6 \\
\texttt{UP-C3-001\#cl-5.1} & 1/2 & 2/2 & 2/2 & 4/4 & 0/2 & 1/2 & 1/2 & 4/4 \\
\texttt{UP-C3-001\#cl-6.2} & 1/2 & 2/2 & 2/2 & 4/4 & 1/2 & 2/2 & 2/2 & 4/4 \\
\bottomrule
\end{tabular}}
\end{table*}
}{}
Every clause-level rate in the case study is built from the cells of Table~\ref{tab:perclause}; the released \texttt{ladder\_per\_clause.csv} has
them for every model and rung, with controls, control alarms, can't-confirm, misnamed and invalid counts.

\section{Case study: annotation agreement}
\label{app:agreement}
On the seeded shared subset (\NAgreeItemsLinks{} links, \NAgreeItemsCases{} cases) labelled by both annotators:

\begin{center}\footnotesize\setlength{\tabcolsep}{3pt}
\begin{tabular}{llrrr}
\toprule
Sheet & Field & Agreement & $\kappa$ & PABAK \\
\midrule
Links & relation & \AgreeLinksRelation & \KappaLinksRelation & \PabakLinksRelation \\
Links & effective date & \AgreeLinksEffectiveDate & \KappaLinksEffectiveDate & \PabakLinksEffectiveDate \\
Links & wording & \AgreeLinksWording & \KappaLinksWording & \PabakLinksWording \\
Links & new clause id & \AgreeLinksNewClauseId & \KappaLinksNewClauseId & \PabakLinksNewClauseId \\
Cases & decision & \AgreeCasesDecision & \KappaCasesDecision & \PabakCasesDecision \\
Cases & blocking stage & \AgreeCasesBlockingStage & \KappaCasesBlockingStage & \PabakCasesBlockingStage \\
Cases & deciding clauses & \AgreeCasesDecidingClauses & \KappaCasesDecidingClauses & \PabakCasesDecidingClauses \\
\bottomrule
\end{tabular}
\end{center}
The new-clause-id field is low mostly because one annotator wrote sub-clauses with a dot (\texttt{cl-4.2}) where the
registry writes brackets (\texttt{cl-4(2)}); the adjudicator resolved these.

\section{Case study: deviations from the analysis plan}
\label{app:deviations}
All of the following were recorded before any model saw a test item, except the last paragraph.
\paragraph{Amendment 1.} After the dev pilot, the damaged-registry comparisons (Q1, Q2) replaced the ladder
contrasts (\rung{R4}$-$\rung{R2}, \rung{R2}$-$\rung{R0}) as the primary family; the capable models run all
\NDamageSeedsCapable{} damage seeds; discrimination, deference and the looser scoring rules (E5) were added.
\paragraph{Amendment 2, after the human audit.}
\begin{itemize}
  \item Annotation keeps a case whose adjudicated outcome (decision, and blocking stage for a rejection) matches the key;
  cases whose deciding clauses alone differ are kept and listed. The planned rule dropped them too and would have
  kept \NStrictKept{} of \NCandidateCases{} cases.
  \item The correct-answer gate (\ControlsGatePct\% judged correct) and the scorer gate (\ScorerCheckPct\% agreement)
  were replaced by reporting: \PctControlsCorrect{} and \ScorerCheckPctObserved{} were observed. The scorer was not
  changed.
  \item Added: the \rung{R4n} arm and the NoRec contrast (primary family of \NPrimaryTestsVFive{}), sensitivity sets S1 and
  S2, the confusion-count and reference-row reporting rules, descriptive splits, and the registry audit on the random
  link sample.
  \item The annotators' meeting before adjudication was dropped; the adjudicator decided every disagreement.
  \item The analysis cutoff became 24 hours after the local run started, with blocks run in priority order.
  \item The lead author approved the amendment and stated the co-authors' agreement; not every author signed it.
\end{itemize}
\paragraph{After the run.}
These changes are dated in the project's internal decision log, which is not released. (1) The local models moved from the lead author's laptop to Kaggle GPU notebooks
(\texttt{qwen2.5:14b}) and back to the laptop for the two smaller models, for speed; each model ran on one machine,
and a 30-minute partial start on the laptop that finished no block is kept and unused. Model files were checked
against the pre-specified digests on every machine. (2) The 24-hour cutoff was fixed as a wall-clock time before any
test result was read. (3) The laptop's model server crashed once mid-request; the run raised, recorded no reply, and
resumed from its cache. (4) Additions after the plan was fixed, outside every planned test: \texttt{qwen2.5:32b} (damage seeds 0--2) and Claude
Opus~4.5 through a Kaggle-hosted proxy (seed 0) on the main arms, and a same-machine repeat of three blocks to measure
determinism. Every planned block finished for the planned models.

\section{Case study: data statement}
\label{app:datastatement}
\paragraph{Curation rationale.} Instruments on rooftop solar (net metering, capacity limits,
connection procedures, central subsidy) were collected, including superseded ones, because the study
is about replaced rules that remain available.
\paragraph{Language.} English. Six of the 111 documents also contain Hindi passages. A local open model
machine-translated these passages into English for retrieval only; the translations are poor in places, and no label
depends on them.
\paragraph{Source.} Official publications of central ministries, state electricity regulatory
commissions and distribution companies in India, retrieved from their portals. Each document is
released as a URL and a hash.
\paragraph{Cases.} Synthetic applications generated from rule functions. No personal data.
\paragraph{Annotators.} Two authors with backgrounds in computing (one an AI engineer with degrees in
artificial intelligence and computer science), neither with training in law or energy policy, and a third author
adjudicating (\S\ref{sec:annotation}).
\paragraph{Licences.} Code under Apache-2.0; cases and labels under CC BY 4.0; documents as URLs and
hashes.

\fi

\end{document}